\documentclass[nologo]{physcapbench_report}
\usepackage{bm}
\usepackage{array,tabularx}
\usepackage{xltabular}
\usepackage{float}
\usepackage{fontawesome5}
\usepackage[section]{placeins}
\usepackage[numbers,square,sort&compress]{natbib}
\setcitestyle{numbers,square,comma}
\hypersetup{
    colorlinks=true,
    allcolors=bluelink,
    pdftitle={PhysCapBench: A Cross-Domain Benchmark for Physically Grounded Video Captioning}
}
\newcolumntype{Y}{>{\raggedright\arraybackslash}X}

\renewcommand\Authfont{\normalfont\fontsize{11}{13}\selectfont\centering}
\renewcommand\Affilfont{\normalfont\fontsize{11}{13}\selectfont\centering}

\title{\centering PhysCapBench: A Cross-Domain Benchmark for Physically Grounded Video Captioning}

\author[1,2]{Haofeng Wang}
\author[2]{Xuanyi Liu}
\author[1]{Pengcheng Wen}
\author[2]{Ruiqi Li}
\author[1]{Yuhao Zhang}
\author[2]{Tiange Zhang}
\author[2]{Boren Zheng}

\author[4]{Siyu Tao}
\author[3]{Danni Yu}
\author[2]{Gan Lin}
\author[2]{Haocheng Tang}
\author[2]{Tianzizhe Guo}
\author[1]{Mingyuan Su}
\author[1,2,5]{Boyuan Chen}
\author[1]{Zhongyuan Wang}
\affil[1]{Beijing Academy of Artificial Intelligence}
\affil[2]{Peking University}
\affil[3]{Tsinghua University}
\affil[4]{Beihang University}
\affil[5]{Physis}

\newcommand{\projectwebsiteurl}{https://hfwang2001.github.io/physcapbench-site/}
\newcommand{\projectcodeurl}{https://github.com/hfwang2001/PhysCapBench}
\newcommand{\projecthuggingfaceurl}{https://huggingface.co/datasets/hfwang01/PhysCapBench}

\newcommand{\projectlink}[3]{%
    \if\relax\detokenize\expandafter{#1}\relax
        {\textcolor{black!65}{#2\hspace{0.4em}#3}}%
    \else
        \href{#1}{#2\hspace{0.4em}#3}%
    \fi
}
\newcommand{\linktable}{%
    {\centering\normalfont\fontsize{10}{12}\selectfont
    \projectlink{\projectwebsiteurl}{\faGlobe}{Website}%
    \hspace{1.8em}%
    \projectlink{\projectcodeurl}{\faGithub}{Code}%
    \hspace{1.8em}%
    \projectlink{\projecthuggingfaceurl}{%
        \raisebox{-2.5pt}{\includegraphics[height=12pt]{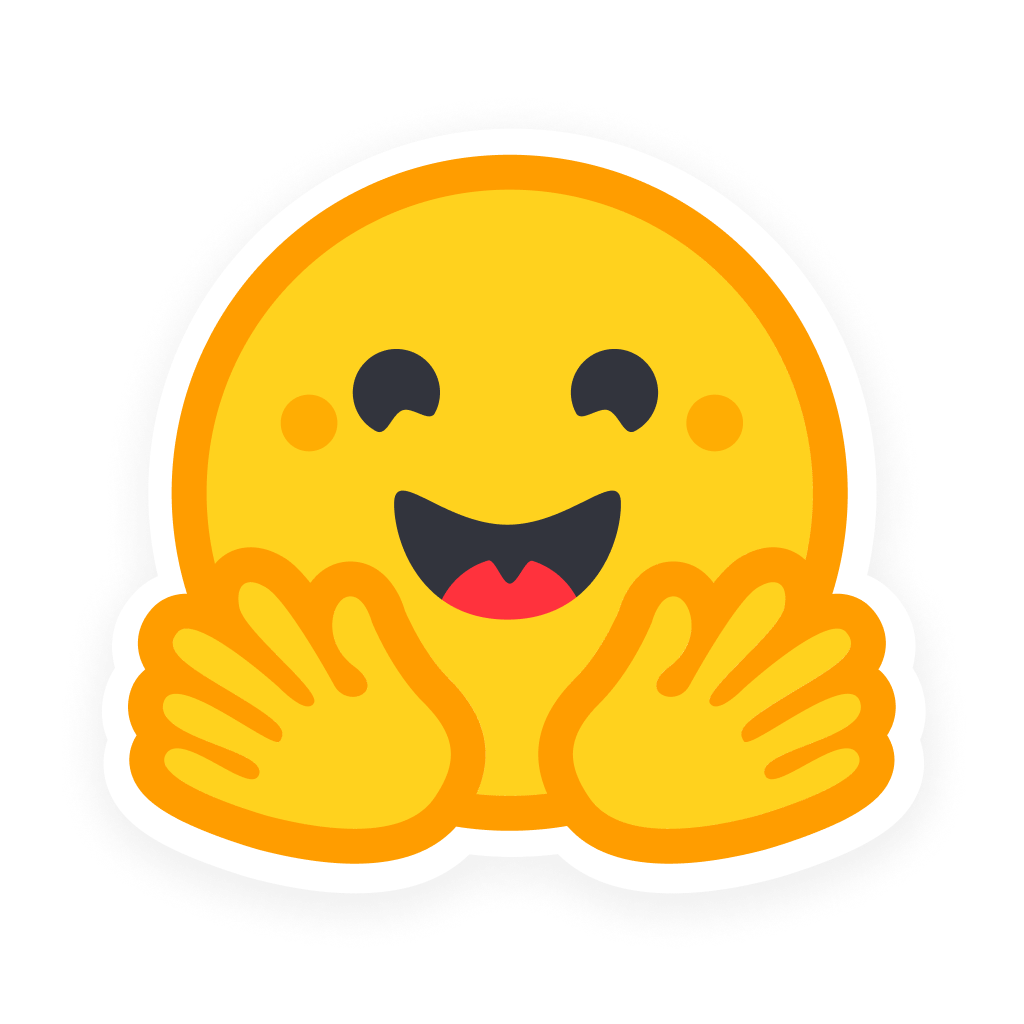}}%
    }{Hugging Face}%
    \par}%
}

\newcommand{\physcapbenchlogopath}{figures/physcapbench_logo_20261110}
\newcommand{\physcapbenchlogo}[1][140pt]{%
    \IfFileExists{\physcapbenchlogopath.pdf}{%
        \includegraphics[width=#1,height=36pt,keepaspectratio]{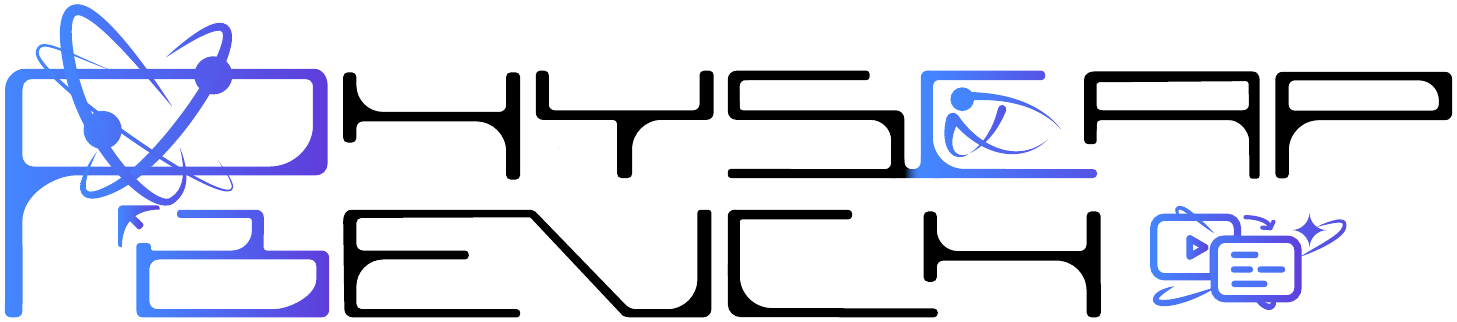}%
    }{%
        \IfFileExists{\physcapbenchlogopath.png}{%
            \includegraphics[width=#1,height=36pt,keepaspectratio]{\physcapbenchlogopath.png}%
        }{%
            \resizebox{#1}{!}{{\color{BAAIHeadingBlue}\sffamily\bfseries\fontsize{18}{22}\selectfont PhysCapBench}}%
        }%
    }%
}

\newcommand{\baaicoverlogo}[1][24pt]{%
    \includegraphics[height=#1]{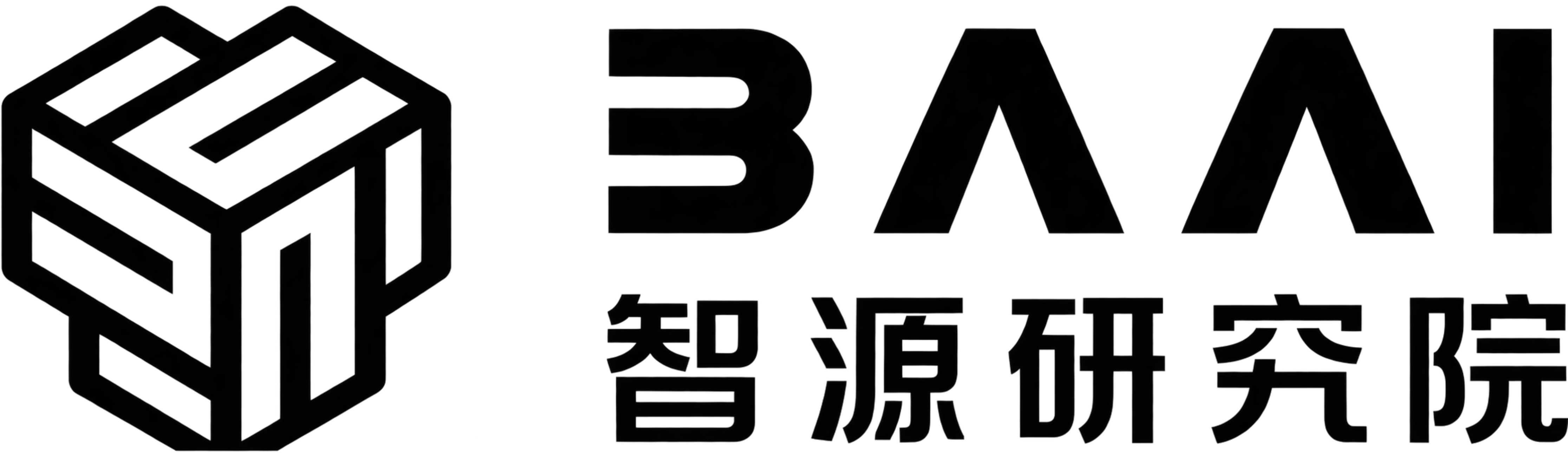}%
}
\newcommand{\physiscoverlogo}[1][24pt]{%
    \includegraphics[height=#1]{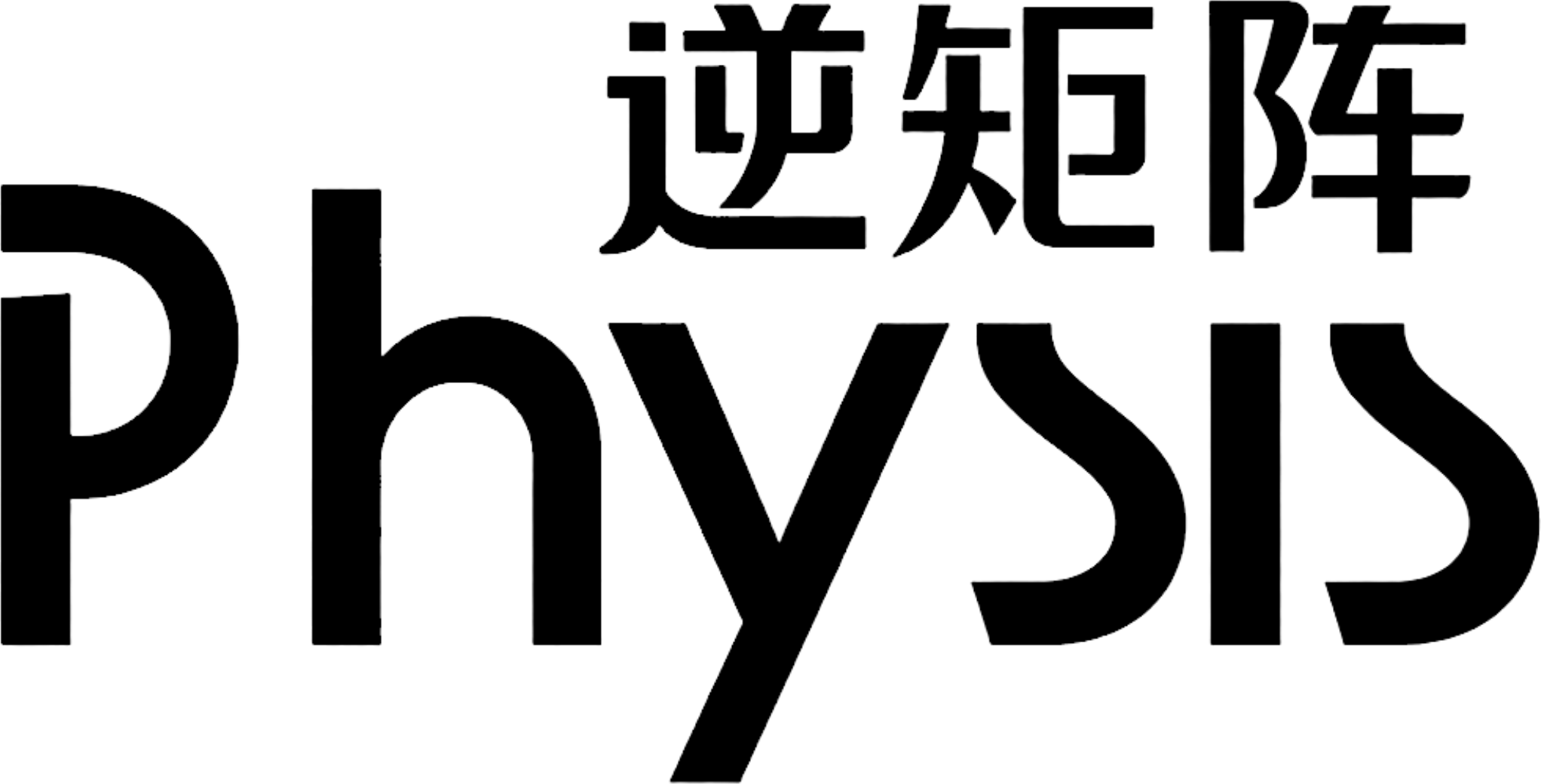}%
}
\newcommand{\coverinstitutionlogos}{%
    \raisebox{\dimexpr-0.5\height+0pt\relax}[0.5\height][0.5\height]{\baaicoverlogo}%
    \hspace{10pt}%
    \raisebox{-0.5\height}{\includegraphics[height=19pt]{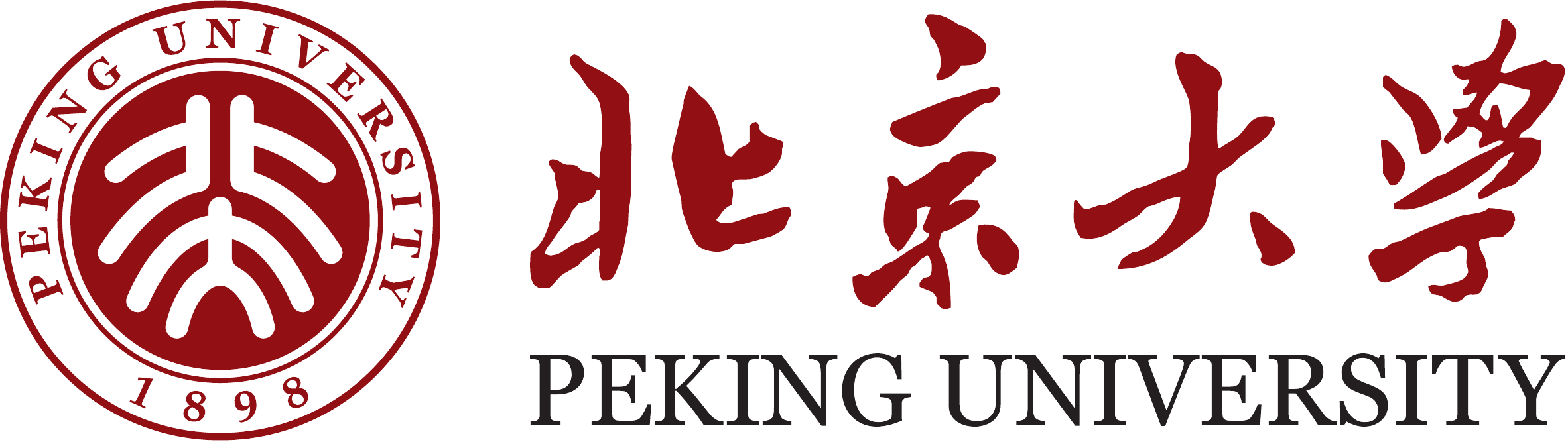}}%
    \hspace{10pt}%
    \raisebox{-0.5\height}{\physiscoverlogo}%
}

\definecolor{appendixgray}{HTML}{858D99}
\newcommand{\reportnavlink}[2]{\hyperref[#1]{\textcolor{black}{#2}}}
\newcommand{\reportappendixlink}[2]{\hyperref[#1]{\textcolor{appendixgray}{#2}}}
\DeclareRobustCommand{\physcapappendixtoc}{%
    \color{appendixgray}\hypersetup{linkcolor=appendixgray}%
}
\newcommand{\sectionheadertext}{%
    \reportnavlink{sec:introduction}{1: Intro}\enspace|\enspace
    \reportnavlink{sec:related}{2: Related Work}\enspace|\enspace
    \reportnavlink{sec:representation}{3: Design}\enspace|\enspace
    \reportnavlink{sec:data_annotation}{4: Data}\enspace|\enspace
    \reportnavlink{sec:evaluation}{5: Evaluation}\enspace|\enspace
    \reportnavlink{sec:limitations}{6: Discussion}\enspace|\enspace
    \reportnavlink{sec:conclusion}{7: Conclusion}\\
    \reportnavlink{sec:references}{References}\enspace|\enspace
    \reportappendixlink{app:schema}{Appendix A: Schema}\enspace|\enspace
    \reportappendixlink{app:selection}{B: Data}\enspace|\enspace
    \reportappendixlink{app:annotation}{C: Annotation}\enspace|\enspace
    \reportappendixlink{app:evaluation}{D: Scoring}\enspace|\enspace
    \reportappendixlink{app:generation}{E: Experiments}%
}
\fancypagestyle{firststyle}{%
    \fancyhf{}
    \fancyhead[L]{%
        \raisebox{\dimexpr-0.5\height+2pt\relax}[0.5\height][0.5\height]{\physcapbenchlogo}%
        \hspace{18pt}\coverinstitutionlogos%
    }
    \fancyhead[R]{%
        \raisebox{-13pt}[0pt][0pt]{\headerfont\textcolor{gray}{Technical Report}}%
    }
    \fancyfoot[C]{\footerfont\thepage}
    \fancyfoot[R]{\datesfont\shorttoday}
    
    \renewcommand{\headrule}{\color{black}\hrule width\headwidth height 1.2pt}
    
}
\newcommand{\physcapbenchrunningheader}{%
    \vbox{%
        \hbox to\headwidth{%
            \begin{minipage}[c]{\headwidth}%
                \centering\normalfont\fontsize{8.3}{10}\selectfont
                \setlength{\parskip}{0pt}%
                \sectionheadertext\par
            \end{minipage}%
        }%
        \vspace{3pt}%
        {\color{black}\hrule height 0.5pt}%
    }%
}
\fancypagestyle{mainstyle}{%
    \fancyhf{}
    \fancyhead[C]{\physcapbenchrunningheader}
    \fancyfoot[C]{\footerfont\thepage}

}
\newcommand{\abstractboxwidth}{\linewidth}
\makeatletter
\renewcommand{\abscontent}{%
    {\centering\color{black}\fontsize{12}{14}\selectfont\textbf{Abstract}\par}%
    \vspace{5pt}%
    \noindent\parbox{\abstractboxwidth}{%
        \linespread{1}\fontsize{10.25}{12.3}\selectfont\theabstract}%
    \par
    \@ifundefined{@keywords}{}{%
        \vskip6pt
        \noindent\parbox{\abstractboxwidth}{\keywordsfont Keywords: \@keywords}\par
    }%
}
\renewcommand{\maketitle}{%
    \begingroup
    \setlength{\parindent}{0pt}%
    \setlength{\parskip}{0pt}%
    \renewcommand\Authfont{\normalfont\fontsize{9}{10.8}\selectfont\centering}%
    \renewcommand\Affilfont{\normalfont\fontsize{8.3}{10}\selectfont\centering}%
    {\centering\titlefont\@title\par}%
    \vspace{6pt}%
    {\centering\@author\par}%
    \vspace{6pt}%
    \linktable
    \vspace{8pt}%
    \abscontent
    \endgroup
    \thispagestyle{firststyle}%
}
\makeatother

\begin{abstract}

Physical AI aims to learn transferable physical knowledge from real-world experience to understand, predict, and act in the physical world.
Internet videos and embodied demonstrations provide complementary observations of motion, interaction, and state evolution.
A foundational data challenge is to measure and preserve this physical information as training supervision.
In video--language pretraining, captions express observed processes in language, yet a semantically correct description can omit the interactions and transitions that define those processes.
\textbf{Rich physical experience in video does not automatically translate into sufficient language supervision.}
This gap calls for a yardstick to assess annotation quality, compare captioning strategies, and guide training-data refinement.
Motivated by our data construction for world-model pretraining, we introduce \textbf{PhysCapBench}, to our knowledge the \textbf{first unified physical-caption benchmark spanning all three domains}: third-person scenes, egocentric manipulation, and handheld-gripper demonstrations.
It provides human-verified annotations for 1,500 video sequences and evaluates accuracy and coverage through \textbf{seven physical dimensions}.
We examine downstream utility through video reconstruction and manipulation-policy fine-tuning.
Physical captions reduce reconstruction LPIPS by \textbf{8.6--9.5\%} on third-person videos and \textbf{26.9--30.9\%} on egocentric videos relative to metadata captions.
Training-instruction enrichment improves both PhysCapBench scores and policy success, including a \textbf{7.5-percentage-point} gain for $\pi_{0.5}$ on a RoboTwin 2.0 subtask.
PhysCapBench makes the physical content of language supervision inspectable and comparable, providing a cross-domain basis for its evaluation and improvement.

\end{abstract}

\begin{document}
\pagestyle{mainstyle}
\maketitle
\raggedbottom

\begingroup
\setlength{\parskip}{0pt}
\vspace{5pt}
\noindent\begin{minipage}{\linewidth}
    \centering
    \includegraphics[width=0.76\linewidth]{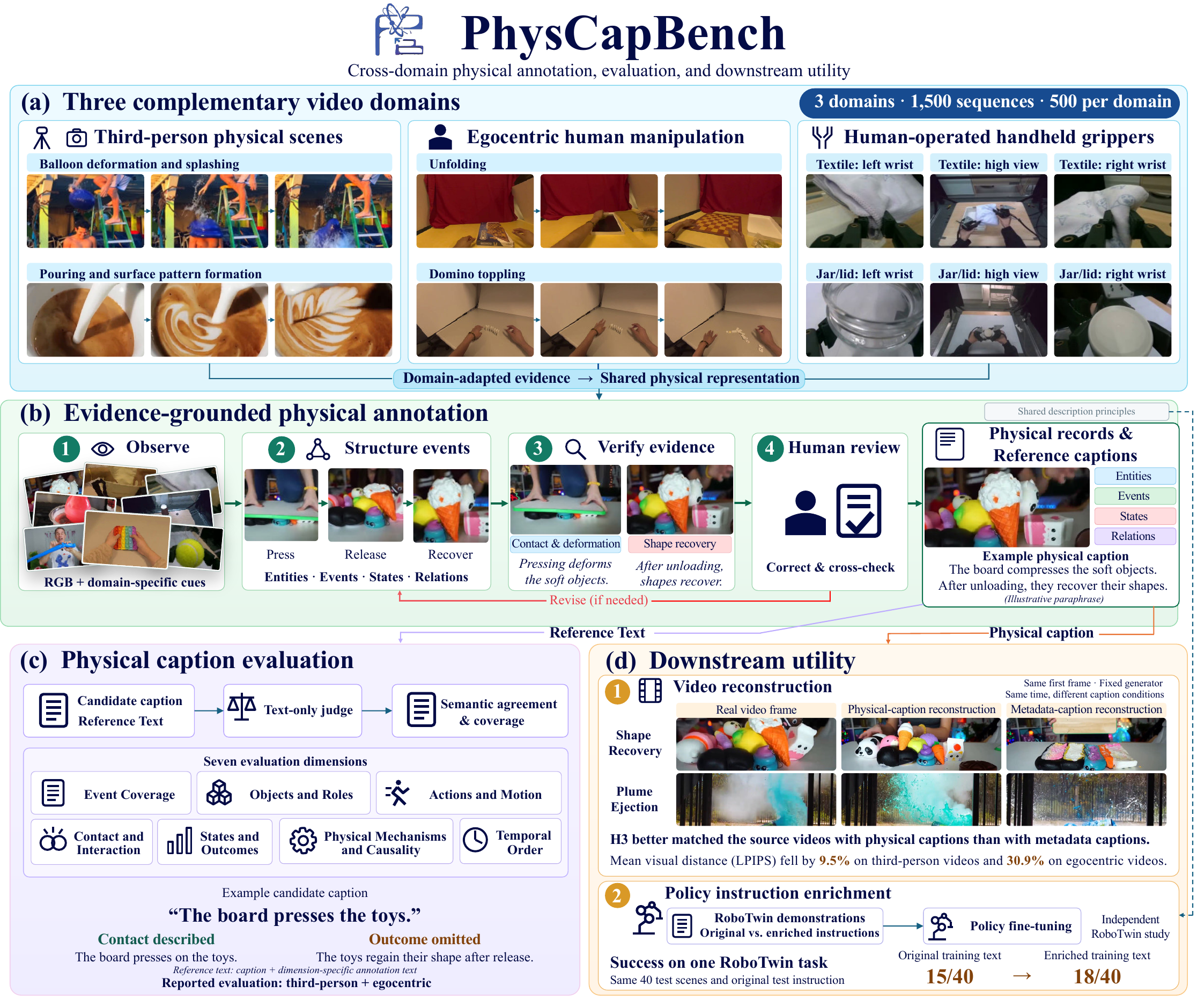}
    \captionsetup{font=small,skip=4pt}
    \captionof{figure}{\textbf{Overview of PhysCapBench.} Across three video domains, PhysCapBench provides evidence-grounded annotations and evaluates captions along seven physical dimensions. The example illustrates how a caption can describe contact while omitting shape recovery.}
    \label{fig:overview}
\end{minipage}
\par
\endgroup

\clearpage
\begingroup
\hypersetup{linkcolor=black}
\setlength{\parskip}{0pt}
\linespread{1}\fontsize{10}{11}\selectfont
\tableofcontents
\endgroup
\clearpage

\setlength{\LTpre}{10pt}
\setlength{\LTpost}{10pt}
\setlength{\LTcapwidth}{\textwidth}
\setlength{\textfloatsep}{12pt plus 2pt minus 2pt}
\setlength{\floatsep}{9pt plus 2pt minus 2pt}
\setlength{\intextsep}{10pt plus 2pt minus 2pt}
\renewcommand{\topfraction}{0.92}
\renewcommand{\bottomfraction}{0.85}
\renewcommand{\textfraction}{0.08}
\renewcommand{\floatpagefraction}{0.65}
\setcounter{topnumber}{4}
\setcounter{bottomnumber}{3}
\setcounter{totalnumber}{6}
\makeatletter
\setlength{\@fptop}{0pt}
\setlength{\@fpsep}{12pt plus 2pt minus 2pt}
\makeatother

\section{Introduction}
\label{sec:introduction}

\begin{center}
\small
\textit{``The limits of my language mean the limits of my world.''}\\
--- Ludwig Wittgenstein
\end{center}

Physical AI aims to learn from real-world experience to understand, predict, and act on the physical world~\citep{ha2018worldmodels,bruce2024genie,nvidia2025cosmos,zhou2026tau0wm}.
Recent foundation-model training pipelines support this goal by integrating videos from multiple domains~\citep{nvidia2025cosmos,zhou2026tau0wm,zhou2026omniworld,gao2026infiniteworlds} and using \textbf{captions as a central component of this cross-domain data infrastructure} to express physical information as language supervision and generation conditions~\citep{nvidia2025cosmos,brooks2024sora,chen2024panda70m,yang2025cogvideox}.
However, these captions can correctly summarize a scene or task while omitting visible physical information about interactions and state changes, leaving that physical information absent from language supervision.
These omissions can hinder physically consistent video generation and limit the ability of manipulation policies to follow detailed execution instructions~\citep{lu2026physislang,hu2026finevla}.
These omissions highlight the need to evaluate whether captions accurately and fully describe the physical information in videos.
We address this need by tackling a key open problem in Physical AI pretraining: \textbf{how can the physical information in captions be systematically evaluated across domains?}

Existing caption benchmarks address parts of this question, but physical evaluation dimensions, multi-domain coverage, and downstream validation have yet to be brought together in a unified benchmark.
First, general caption benchmarks evaluate events and broad visual attributes~\citep{wang2024tarsier,chen2025vidcapbench,zhang2026vcapsbench}.
Physical-caption benchmarks emphasize overall correctness and coverage or broad physical categories~\citep{lu2026physislang,qiu2026omnifysicscaptioner}.
Their limited breakdown across physical dimensions restricts detailed comparisons and targeted caption refinement.
Second, foundation-model training uses video data from multiple domains~\citep{zhou2026omniworld,zhou2026tau0wm,gao2026infiniteworlds}, but existing caption benchmarks do not sufficiently cover these data needs~\citep{zhang2026vcapsbench,hu2026finevla}.
Third, existing studies examine caption utility in video generation~\citep{brooks2024sora,chen2024panda70m,chen2025vidcapbench,lu2026physislang} or manipulation-policy learning~\citep{hu2026finevla}, but no single benchmark has validated physical-caption evaluation in both applications.

To address these gaps, we introduce PhysCapBench, a unified physical-caption benchmark with seven physical evaluation dimensions.
To our knowledge, it is \textbf{the first physical-caption benchmark} spanning all three domains: third-person scenes, egocentric manipulation, and handheld-gripper demonstrations.
PhysCapBench comprises 1,500 video sequences with balanced coverage across the three domains and diverse scenes, tasks, and physical processes within each domain.
In addition, we develop an evidence-grounded annotation pipeline with domain-specific observations, structured annotation, and human cross-checking.
This pipeline produces human-verified physical reference annotations for each video sequence.
We assess captions against these references to identify inconsistencies and omissions along each physical dimension.
Together, these components provide a practical yardstick for comparing captioning strategies and refining captions used in both training and inference.

Our experiments demonstrate clear differences in physical-caption quality across strategies and downstream gains from physical-caption enrichment.
Dimension-wise scores identify the physical details retained or omitted by each captioning strategy.
We then evaluate physical captions as generation conditions through zero-shot, first-frame-conditioned video reconstruction with pretrained video generation models~\citep{minimax2026h3,gao2026infiniteworlds}.
Compared with metadata captions, physical captions reduce LPIPS by 8.6--9.5\% on third-person videos and 26.9--30.9\% on egocentric videos.

To further examine the value of physical captions during training, we use robotic manipulation as a testbed and evaluate enriched physical captions as training instructions.
On RoboTwin 2.0~\citep{chen2026robotwin} and RoboDojo Sim, instruction enrichment raises PhysCapBench scores by 42.17 and 38.58 points, respectively, alongside higher mean task success across four policy families.
For example, on a RoboTwin 2.0 subtask, instruction enrichment improves $\pi_{0.5}$ success from 37.5\% to 45.0\%, a gain of 7.5 percentage points~\citep{intelligence2025pi05}.
The accompanying improvements in PhysCapBench scores and policy success, together with the reconstruction gains, support using PhysCapBench to assess caption quality and guide the refinement of physical information for downstream applications.

\newpage
Our main contributions are as follows:
\begin{enumerate}
    \item \textbf{We introduce PhysCapBench, the first unified physical-caption benchmark spanning third-person scenes, egocentric manipulation, and handheld-gripper demonstrations.}
    PhysCapBench comprises 1,500 video sequences and seven physical evaluation dimensions to assess descriptions of physical processes and identify errors and omissions across captioning strategies.

    \item \textbf{We develop an evidence-grounded physical annotation pipeline.}
    The pipeline adapts to the observations available in each domain and produces structured annotations of entities, interactions, and state evolution. Human verification and cross-checking ensure the quality of the resulting physical reference annotations.

    \item \textbf{We demonstrate the downstream utility of PhysCapBench.}
    {\emergencystretch=2em Video reconstruction and robot manip\-u\-la\-tion-policy fine-tuning support using PhysCapBench to guide caption refinement for downstream applications. Physical-caption enrichment improves reconstruction fidelity, and higher PhysCapBench scores accompany improved policy performance across different benchmarks and policy models.\par}
\end{enumerate}

\section{Related Work}
\label{sec:related}

\subsection{Video Caption Evaluation and Physical Grounding}

Physical understanding is evaluated through tasks with different outputs. PhysBench evaluates answers to physical questions, while VideoPhy assesses semantic adherence and physical commonsense in generated videos~\citep{chow2025physbench,bansal2025videophy}. Captioning raises a complementary question: how accurately and fully does a description convey the physical process in an observed video? PhysCapBench makes this descriptive information the object of evaluation, assessing consistency with structured physical references and coverage of the details they record.

Fine-grained caption benchmarks address this information-level question by checking individual events and facts. DREAM-1K's AutoDQ decomposes captions into events and measures bidirectional entailment between candidate and reference descriptions~\citep{wang2024tarsier}. VidCapBench and VCapsBench use question--answer evaluation to assess detailed caption content, including motion and physical laws alongside visual and aesthetic properties~\citep{chen2025vidcapbench,zhang2026vcapsbench}. These approaches already identify omissions and inconsistencies; VCapsBench also includes egocentric activities. Our annotation and evaluation design concentrates on physical content: it defines seven physical dimensions and uses structured annotations of interactions and state evolution to support caption assessment.

Physical-caption benchmarks pursue this focus through dedicated reference representations and evaluation protocols. Physis-Lang introduces a distinct benchmark, also named PhysCapBench, with human-verified atomic physical assertions; it checks caption precision against video evidence and recall against reference assertions~\citep{lu2026physislang}. OmniFysics-Captioner introduces OmniPhysCap, which probes the physical interactions and outcomes expressed in captions alongside other observable content~\citep{qiu2026omnifysicscaptioner}. Our emphasis is the joint coverage of third-person scenes, egocentric human manipulation, and multi-view demonstrations with human-operated handheld grippers, with annotation adapted to the evidence available in each domain. This design connects physical-caption assessment to the varied observations and data needs of world modeling and manipulation learning.

\subsection{Video Data for World Modeling and Manipulation}

The observations available for physical captioning depend on how video data are collected. Ego-Exo4D synchronizes first- and third-person views of skilled human activities and provides language annotations, including atomic action descriptions~\citep{grauman2024egoexo4d}. EgoDex pairs egocentric manipulation video with tracked hand and finger poses and natural-language metadata~\citep{hoque2025egodex}. The Universal Manipulation Interface (UMI) records human demonstrations with handheld grippers and supports transferring the demonstrated skills to robot policies~\citep{chi2024umi}. In PhysCapBench, EgoDex supplies egocentric video and hand-pose cues, while a separate multi-view handheld-gripper collection supplies synchronized views and trajectories. We adapt physical annotation to these observations while preserving participant identities and roles.

World-model research uses such complementary observations to learn physical dynamics. OmniWorld combines human, robot, simulated, and Internet video, using domain-specific prompts to produce captions~\citep{zhou2026omniworld}. $\tau_0$-WM trains on robot trajectories, UMI-style demonstrations, and egocentric human videos, assigning supervision according to the signals available in each source~\citep{zhou2026tau0wm}. These efforts establish heterogeneous video as a resource for world modeling. When heterogeneous video data are accompanied by captions used for supervision or conditioning, we focus on constructing and assessing their physical content across our three domains.

For manipulation, detailed language can specify how an interaction unfolds beyond its task goal. WGO-Bench evaluates temporal subtask boundaries and descriptions across human egocentric and robot recordings, with annotation centered on completed manipulation events and visible state changes~\citep{macrodata2026wgo}. FineVLA unifies trajectories from ten robot datasets and defines RoboFine-Bench, whose caption track evaluates step-level descriptions against atomic facts across ten execution-related dimensions~\citep{hu2026finevla}. We retain these interaction and state details in our manipulation annotations while also covering general physical scenes. The practical value of such detail can then be tested through the models that use the resulting descriptions.

\subsection{Caption Quality and Downstream Utility}

Video generation offers a setting for examining this practical value. VidCapBench relates caption-evaluation scores to generation quality when captions condition fixed text-to-video models~\citep{chen2025vidcapbench}. Physis-Lang studies physical captions as representations for video world modeling~\citep{lu2026physislang}. Our reconstruction study compares physical and source metadata captions on matched videos, using first-frame conditioning and perceptual distances to reference videos. By evaluating both caption conditions against the same reference, we examine whether enrichment helps preserve the observed visual content.

Manipulation policies provide a complementary test in which language conditions task execution. FineVLA examines how fine-grained instructions affect task success and instruction-sensitive robot behavior~\citep{hu2026finevla}. Our policy studies apply the same description principles to training-instruction enrichment across selected manipulation tasks and multiple policy families. Together with reconstruction, it examines description utility in both generation and control.

\paragraph{Positioning of PhysCapBench.}
Prior work establishes the value of physical descriptions and the use of diverse video data in world modeling. Evidence about caption quality or utility in one data setting leaves other settings to be examined. PhysCapBench addresses this coverage question by bringing third-person scenes, egocentric human manipulation, and multi-view demonstrations with human-operated handheld grippers into one physical-caption benchmark. It defines seven physical evaluation dimensions and adapts annotation to the evidence available in each domain. This scope connects caption assessment to data needs ranging from general scene modeling to embodied manipulation. Reconstruction and policy studies further examine how enriched physical descriptions can serve downstream models.

\section{Task Definition and Benchmark Design}
\label{sec:representation}

PhysCapBench focuses on the accurate and comprehensive description of observable physical processes in videos. Its design comprises a captioning task, a structured physical representation, and seven physical evaluation dimensions. The representation organizes annotated content into entities, events, states, and relations, while the evaluation dimensions guide reference-based assessment of candidate captions. These components establish the basis for the annotation procedures in Section~\ref{sec:data_annotation} and the evaluation protocol in Section~\ref{sec:evaluation}.

\subsection{Task Definition}
\label{sec:task_definition}

Given a video observation $v$, the task is to generate a caption $c$ describing its observable physical process. Depending on the setting, $v$ contains a single view or synchronized views of the same recording. The descriptive target comprises participating entities and their roles, interaction sequences, and visible state changes. The goal is an accurate and comprehensive account of this observable content.

Evaluation compares $c$ with a textual reference $r$, comprising the reference caption and any dimension-specific annotation text used for scoring, assessing \emph{semantic agreement} and \emph{coverage} of the physical information in $r$. Let $D(r)$ contain the physical dimensions assessable from $r$, and let $s_d(c,r)\in[0,100]$ jointly reflect agreement and coverage in dimension $d$. For nonempty $D(r)$, the caption score is
\begin{equation}
    S(c,r)=\frac{1}{|D(r)|}\sum_{d\in D(r)}s_d(c,r).
    \label{eq:caption_score}
\end{equation}
The physical representation below organizes annotation content, while Section~\ref{sec:physical_dimensions} defines the dimensions and Section~\ref{sec:evaluation} details the scoring protocol.

\subsection{Physical Representation}
\label{sec:physical_representation}

The representation organizes physical content around identifiable entities and their participation in events. Entity records associate identities with observed states, including the initial and final states when available. Event records distinguish the acting participant, any instrument, the affected object, and the entity whose motion or state changes. Keeping these roles explicit preserves who acts on what, even when several participants contribute to the same interaction.

Events connect these participants through time. Each event records its interval and the relevant motion, contact, or state change, together with spatial and temporal relations. Timeline segments link the events to the progression of the clip, retaining persistent states and overlapping actions alongside transitions. The representation also accommodates physical mechanism statements and causal relations when supported by observations. Natural-language captions express this organized content for supervision or conditioning, while the structured records retain details that can be inspected individually. These details provide the basis for defining the physical aspects of caption evaluation. Appendix~\ref{app:schema} provides the field definitions, including participant roles, event attributes, and domain-specific extensions.

\subsection{Physical Evaluation Dimensions}
\label{sec:physical_dimensions}

PhysCapBench defines seven physical evaluation dimensions that connect caption assessment to the information organized by the representation. The dimensions distinguish broad event coverage from the details of participants, interactions, and state evolution. A caption may preserve the main activity while omitting the roles of its participants or the resulting state; assessing these aspects separately makes such omissions visible. Table~\ref{tab:physical_dimensions} summarizes the evaluation focus of each dimension and the corresponding annotation content.

\begin{table}[!htbp]
\centering
\caption{\textbf{Physical evaluation dimensions and annotation content.} The mapping links dimensions to annotation content. Text-based scoring measures candidate agreement with and coverage of applicable reference content.}
\label{tab:physical_dimensions}
\begingroup
\small
\setlength{\tabcolsep}{5pt}
\renewcommand{\arraystretch}{1.22}
\setlength{\extrarowheight}{1.5pt}
\begin{tabularx}{\linewidth}{>{\raggedright\arraybackslash}p{0.225\linewidth}YY}
\toprule
\textbf{Physical dimension} & \textbf{Evaluation focus} & \textbf{Annotation content} \\
\midrule
\textbf{Event Coverage} & Coverage of the main physical activities and events. & Event chain and linked timeline segments. \\
\addlinespace[2pt]
\textbf{Objects and Roles} & Agreement on manipulated objects and specified actor or hand roles. & Entity identities and event participants. \\
\addlinespace[2pt]
\textbf{Actions and Motion} & Agreement on actions, movement types, and directions. & Action labels, moving entities, motion types and directions. \\
\addlinespace[2pt]
\textbf{Contact and Interaction} & Agreement on contact and interaction relations. & Contact participants, contact relations, and interaction events. \\
\addlinespace[2pt]
\textbf{States and Outcomes} & Agreement on states, spatial relations, and resulting changes. & Initial and final states, state changes, and spatial relations. \\
\addlinespace[2pt]
\textbf{Physical Mechanisms and Causality} & Agreement on explicitly stated mechanisms or causal explanations. & Mechanism statements, supporting events, and causal relations. \\
\addlinespace[2pt]
\textbf{Temporal Order} & Agreement on explicitly stated order or simultaneity. & Event intervals, temporal relations, and linked timeline segments. \\
\bottomrule
\end{tabularx}
\endgroup
\end{table}

The dimensions connect the physical representation to caption assessment by organizing the reference information into distinct aspects of the observed process. Section~\ref{sec:data_annotation} describes how the corresponding annotations are constructed across the three domains, and Section~\ref{sec:evaluation} presents the scoring protocol, dimension-level results, and downstream experiments.

\section{Data and Physical Annotation}
\label{sec:data_annotation}

\begin{figure}[!t]
\centering
\includegraphics[width=\linewidth]{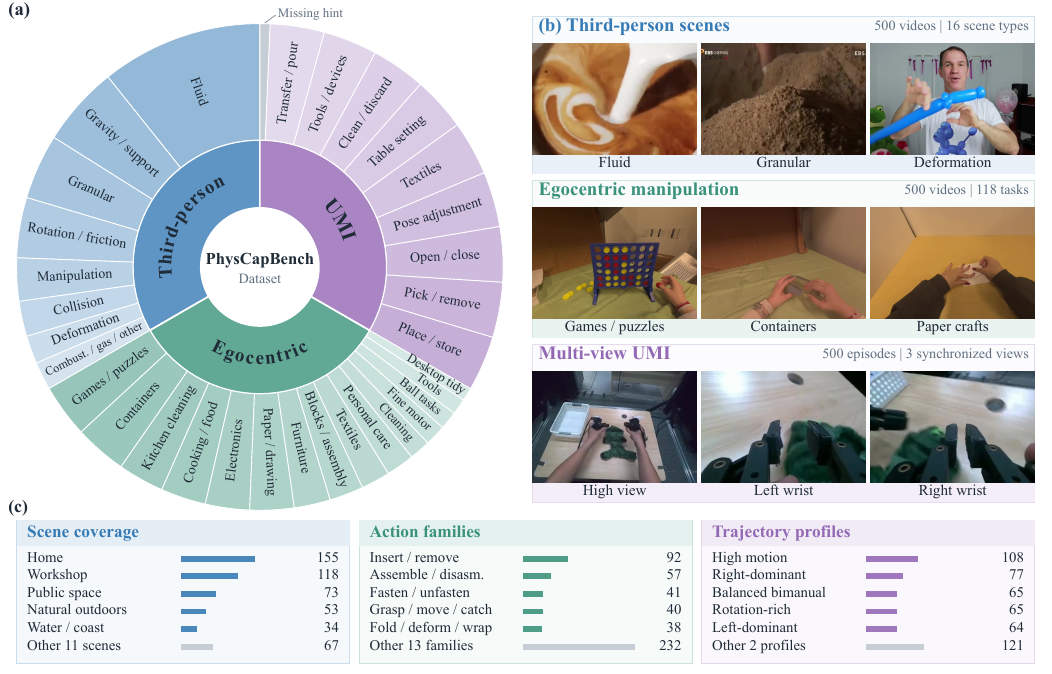}
\caption{\textbf{Dataset overview across three domains.} (a) Category composition. (b) Representative video frames and synchronized handheld-gripper views. (c) Third-person scene types, egocentric action families, and handheld-gripper trajectory profiles, with sequence counts. Each domain contains 500 videos or demonstration episodes.}
\label{fig:dataset_overview}
\end{figure}

The physical representation is applied to three data domains with different viewpoints, participants, and auxiliary observations. Annotation preserves the information available in each domain while organizing it around entities, events, states, and relations. Figure~\ref{fig:dataset_overview} summarizes the category composition, representative observations, and coverage statistics across the three domains. We first summarize the domain-specific choices, data coverage, and annotation quality controls, then describe the data and annotation procedures for each domain.

\subsection{Data and Annotation Overview}
\label{sec:annotation_overview}
\label{sec:annotation_quality}

The three domains provide complementary observations of physical processes. Third-person scenes expose scene-level interactions, egocentric videos show human manipulation from the wearer's viewpoint, and handheld-gripper demonstrations record human-operated interactions across synchronized views. Table~\ref{tab:domain_annotation} summarizes the observations and annotation emphasis in each setting. This coverage responds to the varied data needs of world modeling and manipulation learning, with annotation adapted to each source.

\begin{table}[!htbp]
\centering
\caption{\textbf{Three-domain data and annotation overview.} Auxiliary observations support annotation; their availability to a captioning model is specified separately by the evaluation setting.}
\label{tab:domain_annotation}
\begingroup
\small
\setlength{\tabcolsep}{5pt}
\renewcommand{\arraystretch}{1.22}
\setlength{\extrarowheight}{1.5pt}
\begin{tabularx}{\linewidth}{>{\raggedright\arraybackslash}p{0.225\linewidth}YY}
\toprule
\textbf{Data domain} & \textbf{Available observations} & \textbf{Annotation emphasis} \\
\midrule
\textbf{Third-person physical scenes} & Scene video showing participants and their evolving relations. & Scene entities, physical interactions, motion, and state evolution. \\
\addlinespace[2pt]
\textbf{Egocentric human manipulation} & Egocentric video with hand-pose information available as annotation cues. & Hand identity and roles, hand--object interactions, and visible object changes. \\
\addlinespace[2pt]
\textbf{Multi-view handheld-gripper demonstrations} & Synchronized views and trajectory cues for locating frames to inspect. & Human-operated gripper identities and roles, and corresponding observations across views. \\
\bottomrule
\end{tabularx}
\endgroup
\end{table}

Table~\ref{tab:data_inventory} summarizes the data coverage across the three domains, each containing 500 videos or demonstration episodes. Appendix~\ref{app:selection} provides detailed category distributions and source-video examples for all three domains.

\begin{table}[!htbp]
\centering
\caption{\textbf{Dataset coverage across the three domains.} Multiple views belong to the same handheld-gripper episode.}
\label{tab:data_inventory}
\begingroup\small
\setlength{\tabcolsep}{5pt}\renewcommand{\arraystretch}{1.2}
\begin{tabularx}{\linewidth}{>{\raggedright\arraybackslash}p{0.21\linewidth}>{\raggedright\arraybackslash}p{0.24\linewidth}Y}
\toprule
\textbf{Domain} & \textbf{Data} & \textbf{Coverage and observations} \\
\midrule
Third-person & 500 videos & Eight physical coverage categories and 16 scene categories; object motion, interactions, and state changes. \\
\addlinespace[2pt]
Egocentric & 500 videos & 118 tasks; 15 scenario groups; RGB and hand tracking; approximately 2.21 hours. \\
\addlinespace[2pt]
Handheld gripper & 500 episodes & Nine task groups plus missing hints; three synchronized views and paired gripper trajectories. \\
\bottomrule
\end{tabularx}
\endgroup
\end{table}

\begin{figure}[!htbp]
\centering
\includegraphics[width=\linewidth]{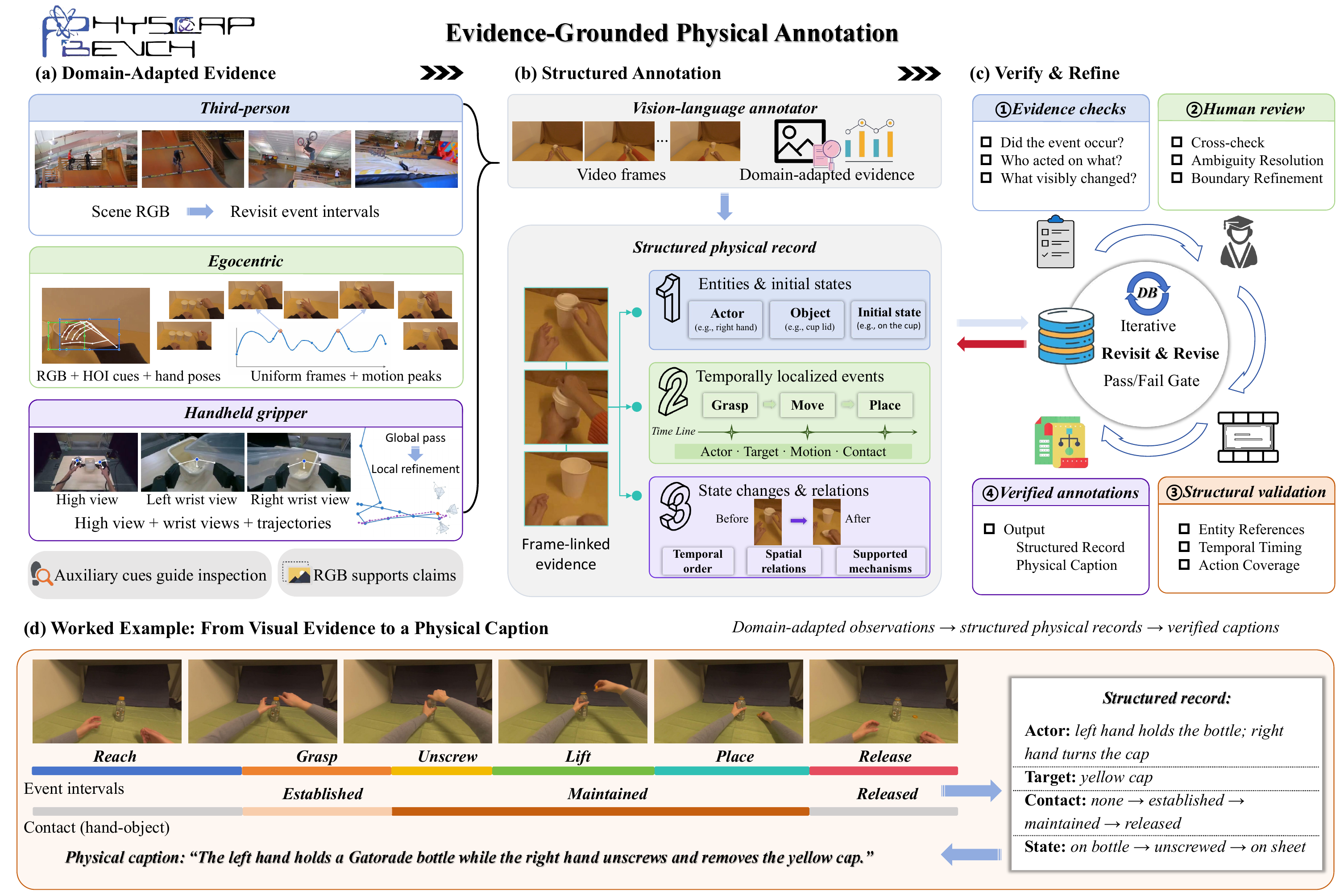}
\caption{\textbf{Evidence-grounded physical annotation.} (a) Domain-adapted observations guide frame selection and visual inspection across third-person, egocentric, and handheld-gripper data. (b) A vision--language annotator produces physical records linked to source frames, organizing entities, events, state changes, and relations. (c) Evidence checks, human cross-review, and structural validation refine the records and captions. (d) A bottle-opening example connects hand roles, event intervals, contact transitions, and object states to a physical caption.}
\label{fig:evidence_annotation}
\end{figure}

Figure~\ref{fig:evidence_annotation} summarizes the annotation workflow, from domain-adapted observations to verified records and captions. Scene RGB, hand--object cues and hand poses, or synchronized views and trajectories guide frame selection and inspection in the respective domains. A vision--language annotator organizes these observations into entities and initial states, temporally localized events, and state changes and relations. Event records retain actor and target identities, motion, and contact, with evidence references linking descriptions to source frames. Domain-specific fields preserve the observation details needed to interpret each interaction.

Across all three domains, annotation combines automated draft generation with human verification. A team of seven professional data annotators performed the annotation and cross-checked one another's work for quality control. All final benchmark annotations have been manually verified against the source videos. Quality control combines structural checks of entity and event references, frame ranges, time conversion, and timeline consistency with human review of the proposed actors, interactions, and state transitions. Evidence references link these checks to the relevant observations, while the annotation records retain review decisions and field-level uncertainty.

Quality control follows the observations available in each domain. Egocentric review checks handedness, hand--object interactions, and the phases of manipulation. Multi-view review checks actor attribution across cameras, object identity across views, and the before-and-after evidence for state transitions. Reviewers record the supporting observations and resolve ambiguous descriptions through source-video inspection. Appendix~\ref{app:annotation} details the review rules.

Figure~\ref{fig:evidence_annotation}(d) illustrates this process with a bottle-opening sequence. The record distinguishes left-hand stabilization from right-hand cap manipulation and follows the cap through grasping, unscrewing, lifting, placement, and release. Event intervals and before-and-after observations connect the action sequence to changes in contact and object state. The resulting caption preserves the two hands' roles and the cap-removal event. The following sections describe how this workflow is implemented in each domain.

\subsection{Third-Person Physical Scenes}
\label{sec:third_data}

Third-person videos extend the benchmark beyond intentional manipulation to physical processes observed at the scene level. The collection includes object motion, tool--material interaction, deformation, and other visible changes in the environment. The dataset contains 500 videos spanning eight primary physical coverage categories and 16 scene categories, and retains the source metadata descriptions. Appendix~\ref{app:third_coverage} reports these distributions, and Figure~\ref{fig:third_examples} presents source-video examples. Annotation uses the recorded visual sequence to determine the participating entities, their interactions, and the resulting states; source metadata is retained as the comparison caption.

The annotation procedure first organizes the scene into entities and temporally localized events, then records motion, contact, and state changes within each event. Review revisits the corresponding video intervals to correct object references, movement directions, and event boundaries. For the reconstruction study, the physical descriptions are expanded from the human-corrected fields, retaining the information needed to describe the observed process. This produces a direct comparison between the source description and a description grounded in the annotated video. Section~\ref{sec:reconstruction} reports the reconstruction results.

\subsection{Egocentric Human Manipulation}
\label{sec:ego_data}

The egocentric dataset contains 500 videos from EgoDex~\citep{hoque2025egodex}, covering 118 tasks with four or five videos per task. Each video is paired with its hand-tracking record, and the total duration is approximately 2.21 hours. Appendix~\ref{app:ego_coverage} details the selection and the scenario, action, object, and physical-phenomenon distributions. Figure~\ref{fig:ego_examples} illustrates the observed manipulation processes.

\begin{figure}[!t]
\centering
\includegraphics[width=\linewidth]{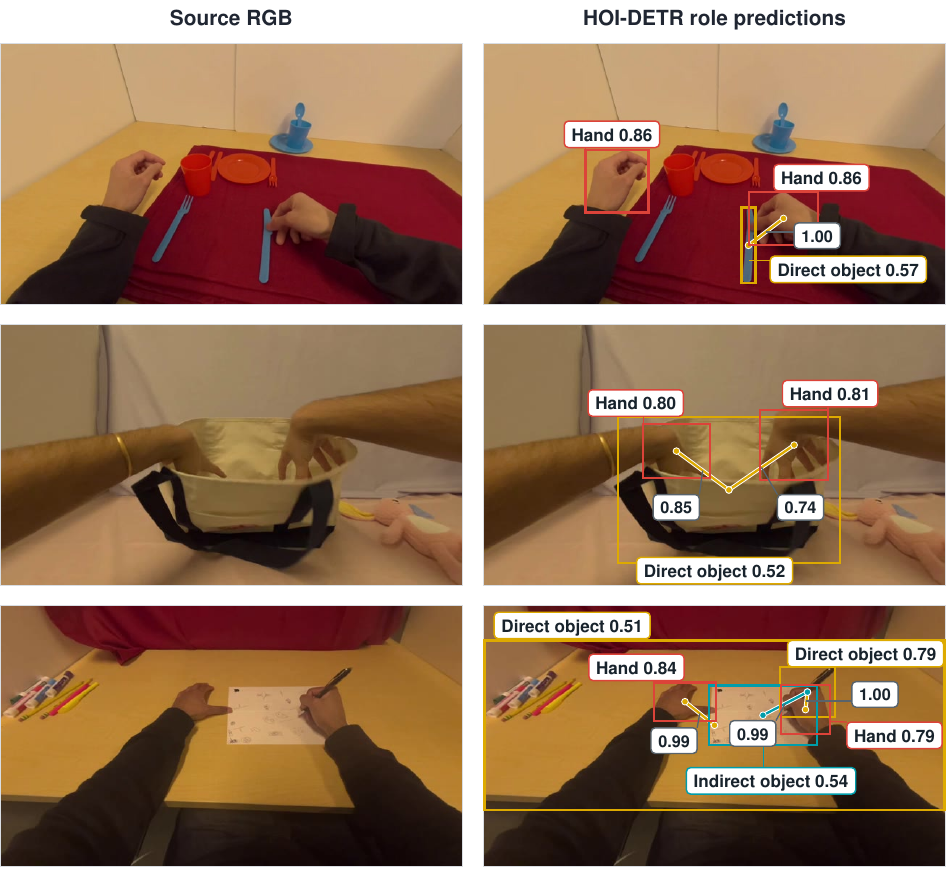}
\caption{\textbf{Interaction-role cues for egocentric physical annotation.} Source RGB frames (left) and HOI-DETR predictions (right). Red, yellow, and cyan boxes denote hands, directly manipulated objects, and objects affected through them; links show predicted interaction relations. Numbers beside boxes and links indicate detection and relation confidence, respectively. The role structure guides annotation of the acting hand, the manipulated object, and tool-mediated interactions, supporting the description of who acts on what.}
\label{fig:hoi}
\end{figure}

Egocentric annotation distinguishes the acting hand, the directly manipulated object, and any object affected through that interaction. HOI-DETR~\citep{darkhalil2026hoi} provides hand--object and object--object role predictions, while EgoDex hand poses supply anatomical hand identity and motion cues. The vision--language annotator combines these signals with RGB observations to describe the participants and their interactions. Figure~\ref{fig:hoi} illustrates how the detector's role structure localizes direct manipulation and tool-mediated interactions.

Hand poses also guide temporal sampling. We transform the palm and five fingertips into camera coordinates, compute hand translation speed and palm angular speed, and combine them into a motion score for each hand. Temporally separated peaks summarize each hand's active moments. For visual inspection, we select peaks from the maximum of the two hands' translation speeds and merge them with uniformly sampled frames. This combines coverage of the full sequence with closer inspection of rapid hand movements. Appendix~\ref{app:annotation} gives the equations and selection parameters.

Using these frames and auxiliary cues, the annotator assigns entity identities, separates action phases, and records changes in contact and object state. Camera-relative motion helps distinguish the two hands' contributions, and RGB observations connect these movements to the manipulated objects. Event intervals preserve concurrent actions, while a continuous timeline retains sustained holding and intervening movements alongside salient transitions.

\subsection{Multi-View Handheld-Gripper Demonstrations}
\label{sec:umi_data}

The third domain contains 500 human-operated handheld-gripper demonstrations from Tencent's \href{https://huggingface.co/datasets/tencent/Hy-Embodied-0.5-VLA-Data}{Hy-Embodied-0.5-VLA-Data}. It follows the handheld demonstration setting associated with UMI~\citep{chi2024umi}: the observed manipulation is performed by a person controlling the grippers. Each episode provides a fixed high view and two wrist views, synchronized at 30 frames per second, together with left and right gripper trajectories. Each view has resolution $424\times240$. Appendix~\ref{app:umi_coverage} reports the selection, task distribution, trajectory profiles, and durations; Figure~\ref{fig:umi_examples} shows examples of the synchronized observations.

We organize annotation into a global pass and local multi-view refinement~\citep{hu2026finevla}. The global pass examines high-view frames to identify scene entities and the broad event sequence. Local refinement then inspects synchronized high and wrist views around candidate transitions, located using gripper position, rotation, and opening changes. It assigns actor identity, matches manipulated objects across views, and resolves contact and state changes before producing the structured record and caption.

Cross-view consistency connects the global event sequence to local interaction details. Each observation records its camera view and source frame, while actor identity is assigned from the visible gripper and its interaction with the object. Motion is described relative to an observable reference to account for wrist-camera movement. High-view observations establish the scene context, and wrist views provide complementary detail about contact and state changes within the same demonstration.

\section{Evaluation and Downstream Utility}
\label{sec:evaluation}

Our evaluation connects physical-caption quality with its use in video generation and manipulation learning. We first compare captioning strategies on third-person and egocentric videos using the dimensions in Section~\ref{sec:physical_dimensions}. We then measure reconstruction fidelity under physical-caption conditioning and policy success under instruction enrichment. Scoring the original and enriched training descriptions links the benchmark evaluation to the policy comparison.

\subsection{Physical Caption Evaluation Protocol}
\label{sec:caption_protocol}

Caption evaluation measures semantic agreement and coverage along the seven physical dimensions in Table~\ref{tab:physical_dimensions}. Doubao Seed 2.1 Pro serves as the text judge, with temperature zero and thinking disabled, assigning each dimension an integer score from 0 to 100 and a brief rationale. For egocentric videos, physical mechanisms and causality are scored against the mechanism statements (\texttt{criteria.physical\_mechanism}) in the fixed Gold annotation; the other six dimensions use its short caption (\texttt{goldCaptionEn}).

Equation~\eqref{eq:caption_score} averages each caption's scores over the dimensions applicable to its Gold reference. These dimensions are fixed across strategies for the same video. Overall scores weight videos equally; dimension-level scores average the applicable entries. Appendix~\ref{app:evaluation} provides the judge instructions and scoring rubric.

\subsection{Caption Evaluation Results}
\label{sec:caption_results}

\begin{figure}[!htbp]
\centering
\includegraphics[width=0.78\linewidth]{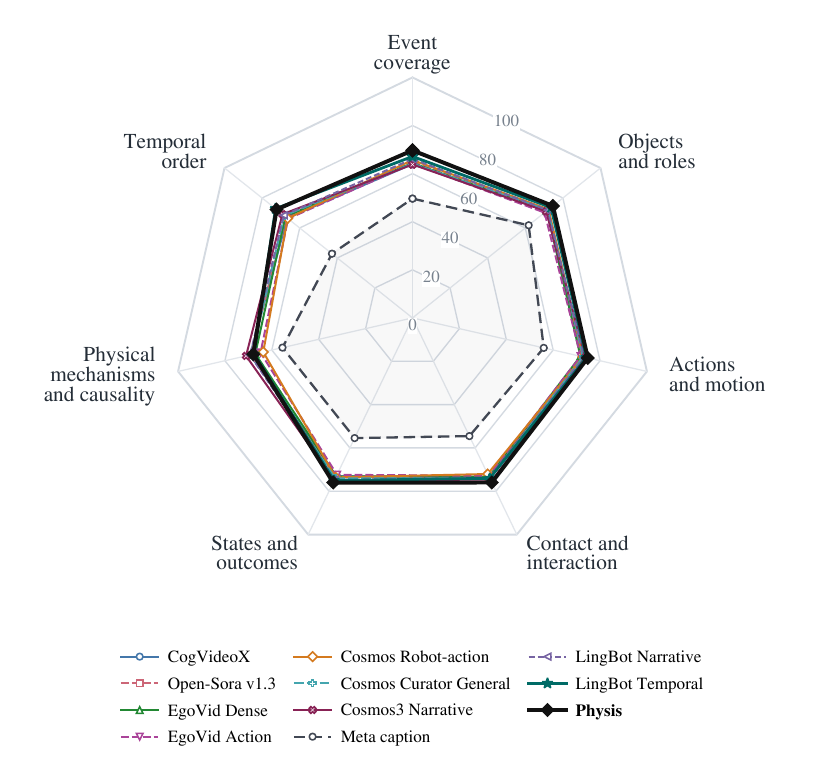}
\caption{\textbf{Physical caption evaluation on the third-person dataset.} Physis is highlighted in black. The seven axes correspond to physical evaluation dimensions, and each curve represents a captioning strategy. All strategies are evaluated against the same dimension-specific references. Scores range from 0 to 100, with higher scores farther from the center. The LingBot strategies are reconstructed from the paper description. \textbf{Physis captioning is currently in internal testing.}}
\label{fig:third_person_caption_results}
\end{figure}

The comparisons examine how captioning strategies capture physical content within each domain. Figure~\ref{fig:third_person_caption_results} reports the third-person results, and Table~\ref{tab:caption_results} reports the egocentric results, with references and scoring settings held fixed within each comparison.

\paragraph{Third-person caption evaluation.}
Physis achieves the highest overall score of 74.07 among the 11 strategies, compared with 72.91 for LingBot Temporal and 54.32 for Meta, and leads in five of the seven physical dimensions (Figure~\ref{fig:third_person_caption_results}). LingBot Temporal leads in temporal order (72.84), while Cosmos3 Narrative leads in physical mechanisms and causality (71.00). The dimension scores distinguish strategies that retain event progression from those that capture richer interaction and mechanism detail.

\paragraph{Egocentric caption evaluation.}
For egocentric videos, we implement five public captioning strategies on Qwen3-VL-8B-Instruct using their respective prompts and frame-sampling recipes. Physis uses captions from version 0921-2301, and Meta uses the original source captions. Table~\ref{tab:caption_results} compares all seven strategies under the same judge and references.

\begin{table}[!htbp]
\centering
\caption{\textbf{Physical caption evaluation on the egocentric dataset.} All strategies are evaluated on the same egocentric clips using the same dimension-specific references. Scores range from 0 to 100; higher is better. Bold indicates the highest score in each row. The overall average is computed from per-video scores over applicable dimensions, as defined in Section~\ref{sec:caption_protocol}.}
\label{tab:caption_results}
\label{tab:caption_dimensions}
\begingroup\footnotesize
\setlength{\tabcolsep}{2pt}\renewcommand{\arraystretch}{1.32}
\begin{tabular*}{\linewidth}{@{\extracolsep{\fill}}l*{7}{c}@{}}
\toprule
\textbf{Physical dimension} & \textbf{CogVideoX} & \shortstack{\textbf{Open-Sora}\\\textbf{v1.3}} & \shortstack{\textbf{EgoVid}\\\textbf{Dense}} & \shortstack{\textbf{EgoVid}\\\textbf{Action}} & \shortstack{\textbf{Cosmos}\\\textbf{Ego}} & \textbf{Physis} & \shortstack{\textbf{Meta}\\\textbf{caption}} \\
\midrule
Event coverage & 60.17 & 60.25 & 61.50 & 57.33 & 81.08 & \textbf{84.42} & 49.17 \\
Objects and roles & 61.17 & 62.92 & 63.08 & 57.25 & 78.50 & \textbf{84.83} & 46.67 \\
Actions and motion & 63.18 & 59.91 & 62.64 & 59.55 & 84.64 & \textbf{87.73} & 54.09 \\
Contact and interaction & 61.43 & 54.29 & 59.29 & 54.29 & \textbf{83.57} & 77.14 & 38.57 \\
States and outcomes & 71.88 & 68.25 & 70.88 & 60.62 & 80.12 & \textbf{92.75} & 55.00 \\
\shortstack[l]{Physical mechanisms\\and causality} & 25.83 & 31.42 & 30.00 & 35.83 & 46.50 & \textbf{49.58} & 7.08 \\
Temporal order & 56.25 & 66.25 & 63.75 & 67.50 & \textbf{77.50} & 75.75 & 42.50 \\
\addlinespace[3pt]
\textbf{Overall average} & 57.16 & 58.01 & 58.42 & 55.58 & 75.41 & \textbf{79.36} & 42.41 \\
\bottomrule
\end{tabular*}
\endgroup
\end{table}

Physis achieves the highest overall score of 79.36, compared with 42.41 for Meta, and scores 92.75 for states and outcomes. Meta scores 49.17 for event coverage, 38.57 for contact and interaction, and 42.50 for temporal order, identifying physical content that its task-level descriptions leave under-specified. Cosmos Ego leads in contact and interaction and temporal order, scoring 83.57 and 77.50, compared with 77.14 and 75.75 for Physis. Mechanism scores further separate the strategies: Physis scores 49.58, Cosmos Ego 46.50, and Meta 7.08.

The EgoVid comparison shows how compression shifts descriptive emphasis. EgoVid Action compresses the dense description and scores 55.58 overall, compared with 58.42 for EgoVid Dense. States and outcomes decrease from 70.88 to 60.62, while temporal order increases from 63.75 to 67.50 and mechanisms and causality from 30.00 to 35.83. These opposing changes show why the dimension scores complement the overall average.

\subsection{Video Reconstruction Experiments}
\label{sec:reconstruction}

Video reconstruction tests the value of physical descriptions as generation conditions. Each generator receives the first source frame and either the original metadata caption or a physical description, with the frame and target duration fixed across conditions. We evaluate MiniMax H3 with a 33B backbone~\citep{minimax2026h3} and LingBot-World 2.0 (14B)~\citep{gao2026infiniteworlds} in both domains. Third-person descriptions expand the human-corrected physical annotations; egocentric descriptions specify the manipulation sequence, hand roles, and visible changes.

We measure fidelity to the source video using LPIPS~\citep{zhang2018lpips} for frame-level perceptual differences and perceptual video clip similarity (PVCS) for local-clip differences in pretrained video features.\footnote{PVCS implementation: \href{https://github.com/MichiganCOG/video-inpainting-evaluation}{video-inpainting evaluation toolkit}.} Lower distances indicate better reconstruction. Qualitative comparisons examine the preservation of interaction roles, contact transitions, and object states.

\begin{figure}[!htbp]
\centering
\includegraphics[width=\linewidth]{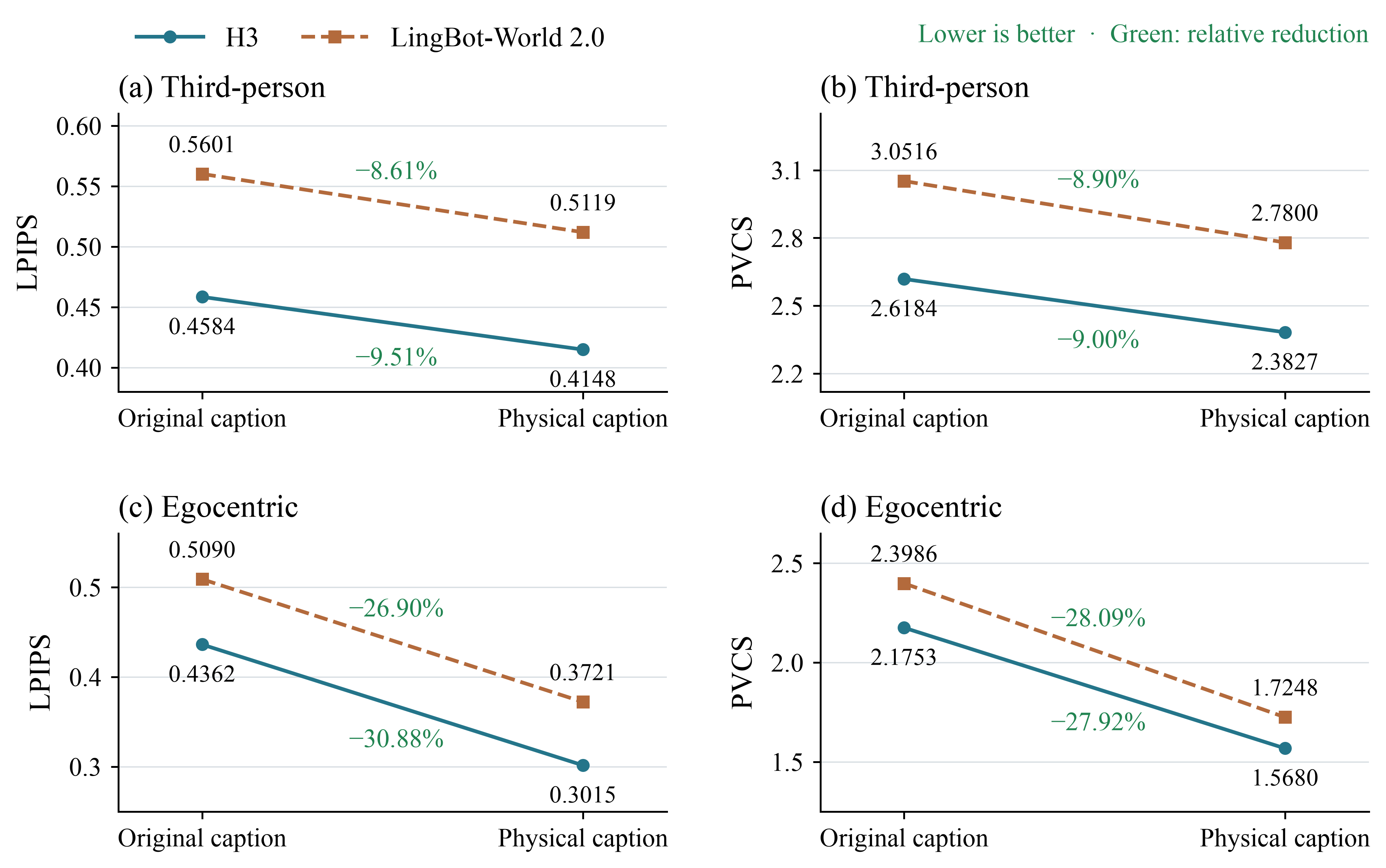}
\caption{\textbf{Physical captions improve video reconstruction across domains.} LPIPS and PVCS are shown for third-person (top) and egocentric (bottom) videos using H3 and LingBot-World 2.0. Original captions are source metadata captions. Point labels report mean distances; green annotations show relative reductions under physical-caption conditioning. Lower is better.}
\label{fig:reconstruction_results}
\end{figure}

Physical captions improve reconstruction in all four model--domain settings (Figure~\ref{fig:reconstruction_results}). LPIPS decreases by 8.6--9.5\% on third-person videos and 26.9--30.9\% on egocentric videos; PVCS decreases by 8.9--9.0\% and 27.9--28.1\%, respectively. The larger relative reductions on egocentric videos hold for both generators and both metrics. Appendix~\ref{app:generation} details the reconstruction setup and third-person confidence intervals.

\begin{figure}[!htbp]
\centering
\includegraphics[width=\linewidth]{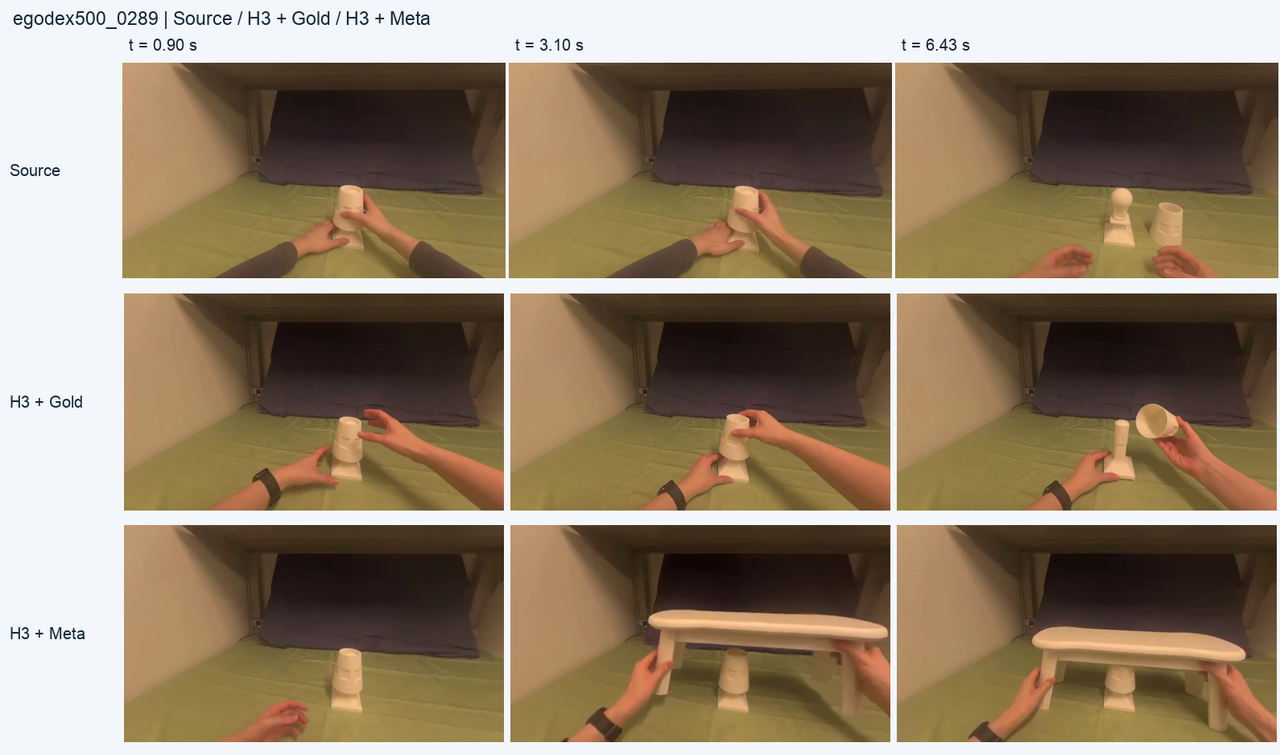}
\caption{\textbf{Physical captions preserve interaction roles and state changes in reconstruction.} Rows show source frames (top), H3 with the physical caption (middle; labeled Gold in the source experiment), and H3 with the metadata caption (bottom). Columns correspond to the same times. Physical conditioning better retains left-hand stabilization, right-hand shell removal, and the transition from a covered to an exposed interior. Metadata conditioning instead generates the assembly of a larger stool.}
\label{fig:reconstruction}
\end{figure}

Physical conditioning better preserves the hand roles and state transition in Figure~\ref{fig:reconstruction}. The source video shows the left hand stabilizing a small object's base while the right hand rotates and lifts its shell to expose the interior. The physical-caption reconstruction retains this division of roles and the shell--base separation. The metadata caption describes assembling a stool, and its reconstruction follows that object and activity. The comparison highlights the contribution of describing the acting hand, affected part, and direction of change.

\begin{figure}[!htbp]
\centering
\begingroup
\setlength{\tabcolsep}{2pt}
\renewcommand{\arraystretch}{1.0}
\begin{tabular}{@{}p{0.065\linewidth}p{0.299\linewidth}p{0.299\linewidth}p{0.299\linewidth}@{}}
 & \centering\small\textbf{Source} & \centering\small\textbf{Gold (physical)} & \centering\arraybackslash\small\textbf{Meta} \\
\multicolumn{4}{@{}l@{}}{\rule{0pt}{14pt}\small\textbf{(a) Shape after unloading}\enspace\texttt{physcap9k\_08653}}\\[2pt]
{\scriptsize $2.00$\,s} & \includegraphics[width=\linewidth]{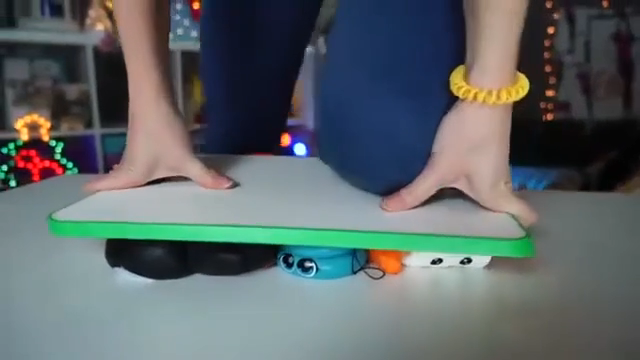} & \includegraphics[width=\linewidth]{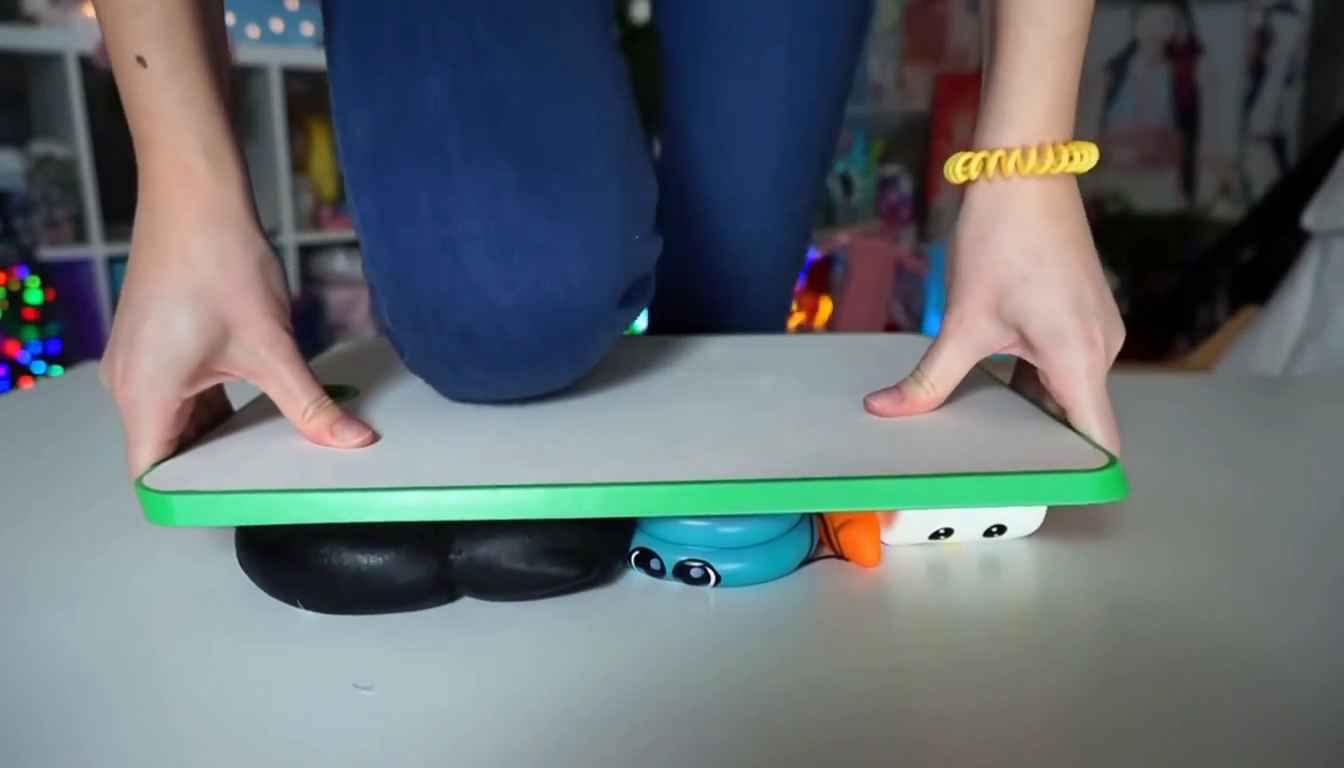} & \includegraphics[width=\linewidth]{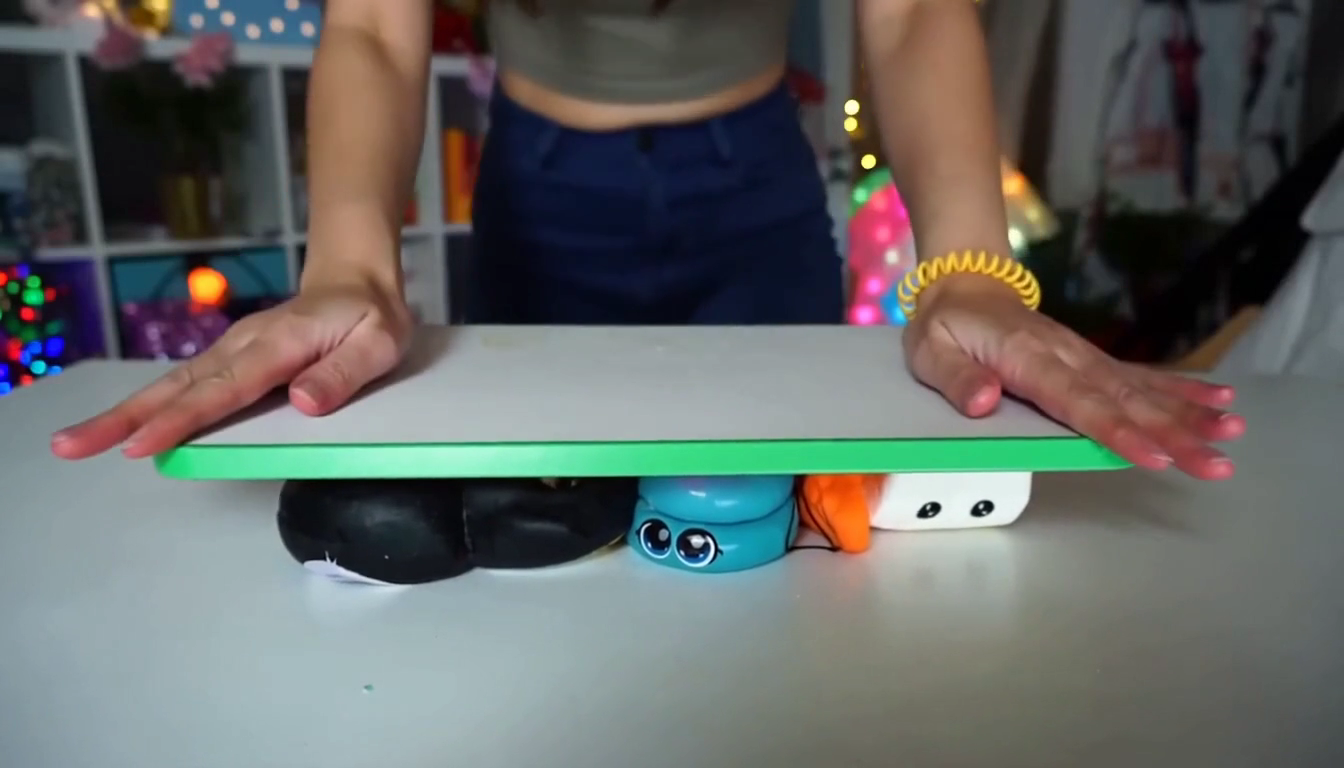}\\[2pt]
{\scriptsize $9.50$\,s} & \includegraphics[width=\linewidth]{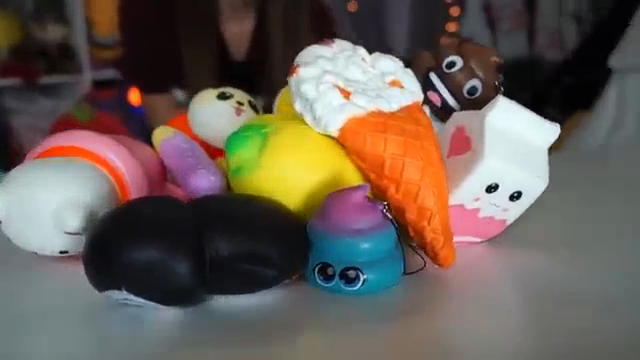} & \includegraphics[width=\linewidth]{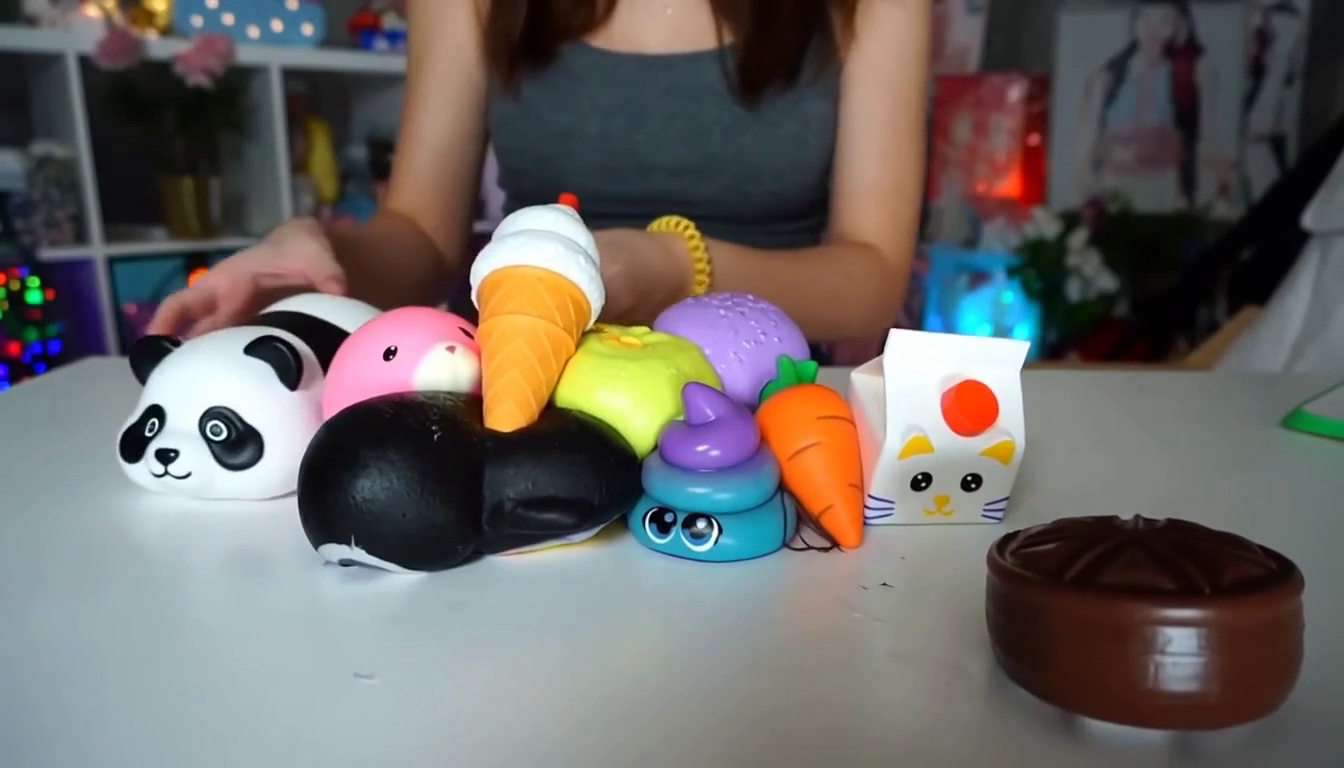} & \includegraphics[width=\linewidth]{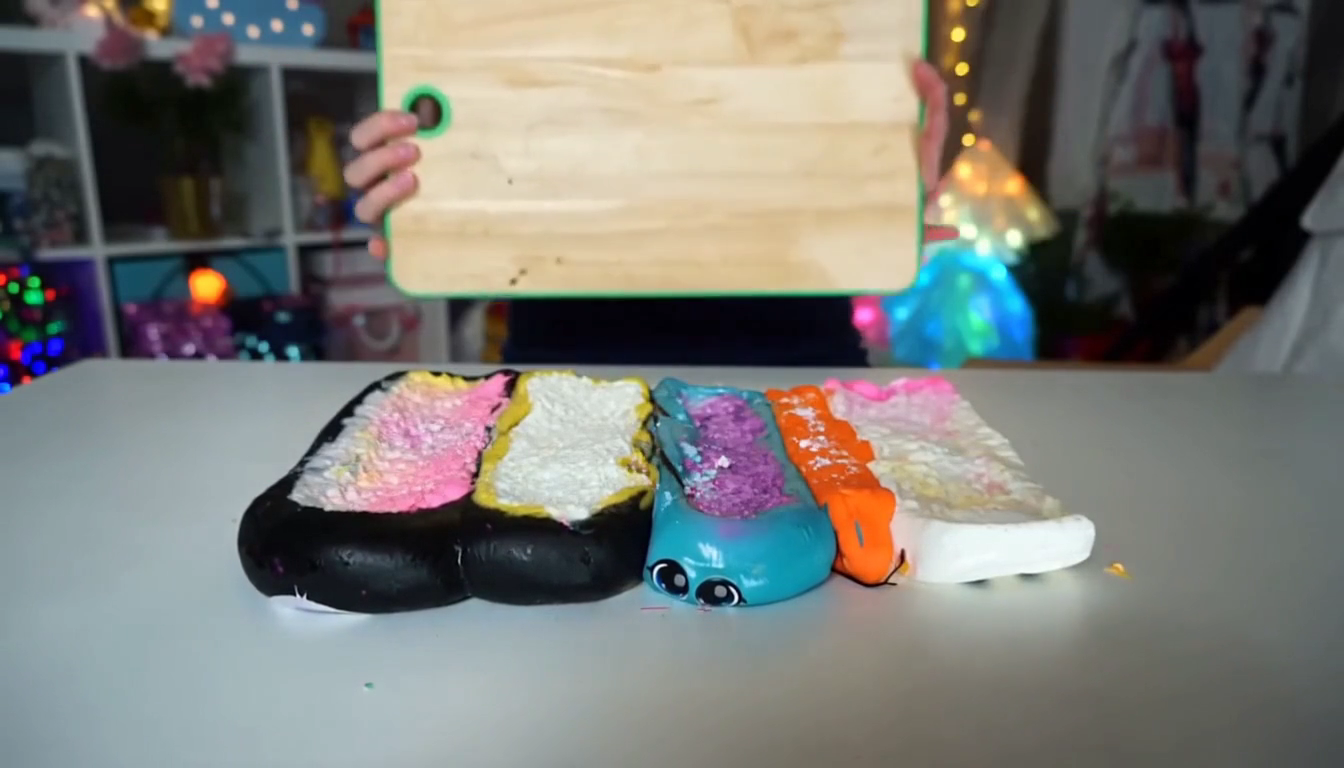}\\[2pt]
\multicolumn{4}{@{}l@{}}{\rule{0pt}{14pt}\small\textbf{(b) Drilling and contact termination}\enspace\texttt{physcap9k\_02537}}\\[2pt]
{\scriptsize $6.00$\,s} & \includegraphics[width=\linewidth]{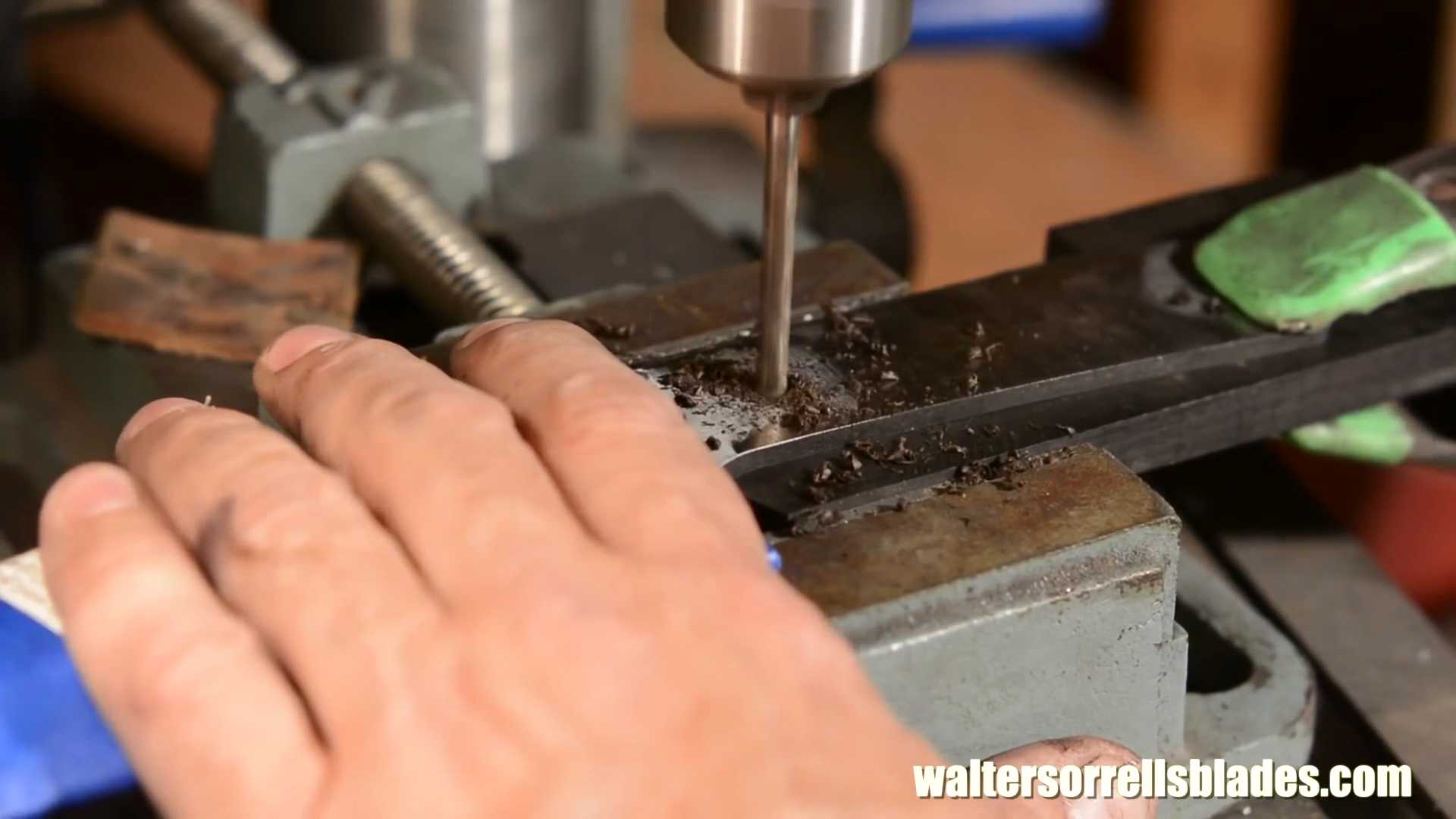} & \includegraphics[width=\linewidth]{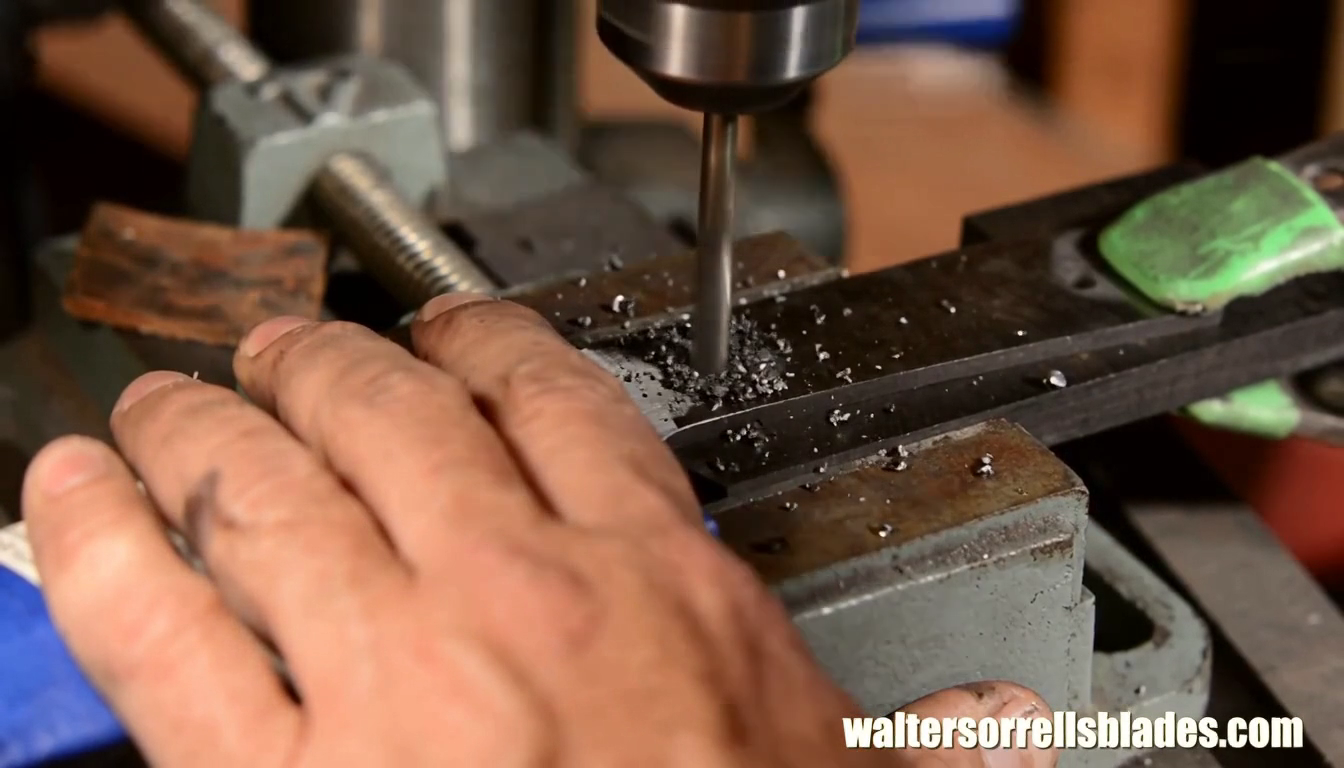} & \includegraphics[width=\linewidth]{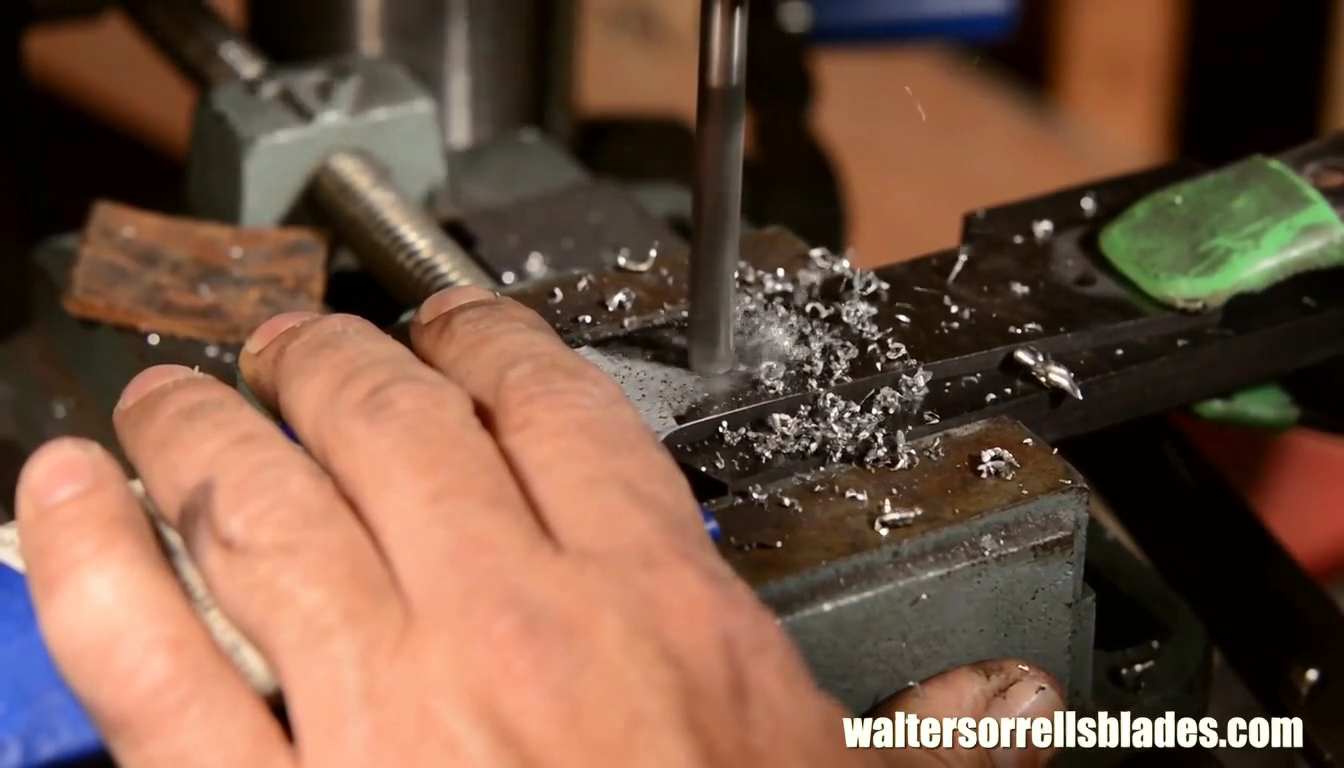}\\[2pt]
{\scriptsize $9.13$\,s} & \includegraphics[width=\linewidth]{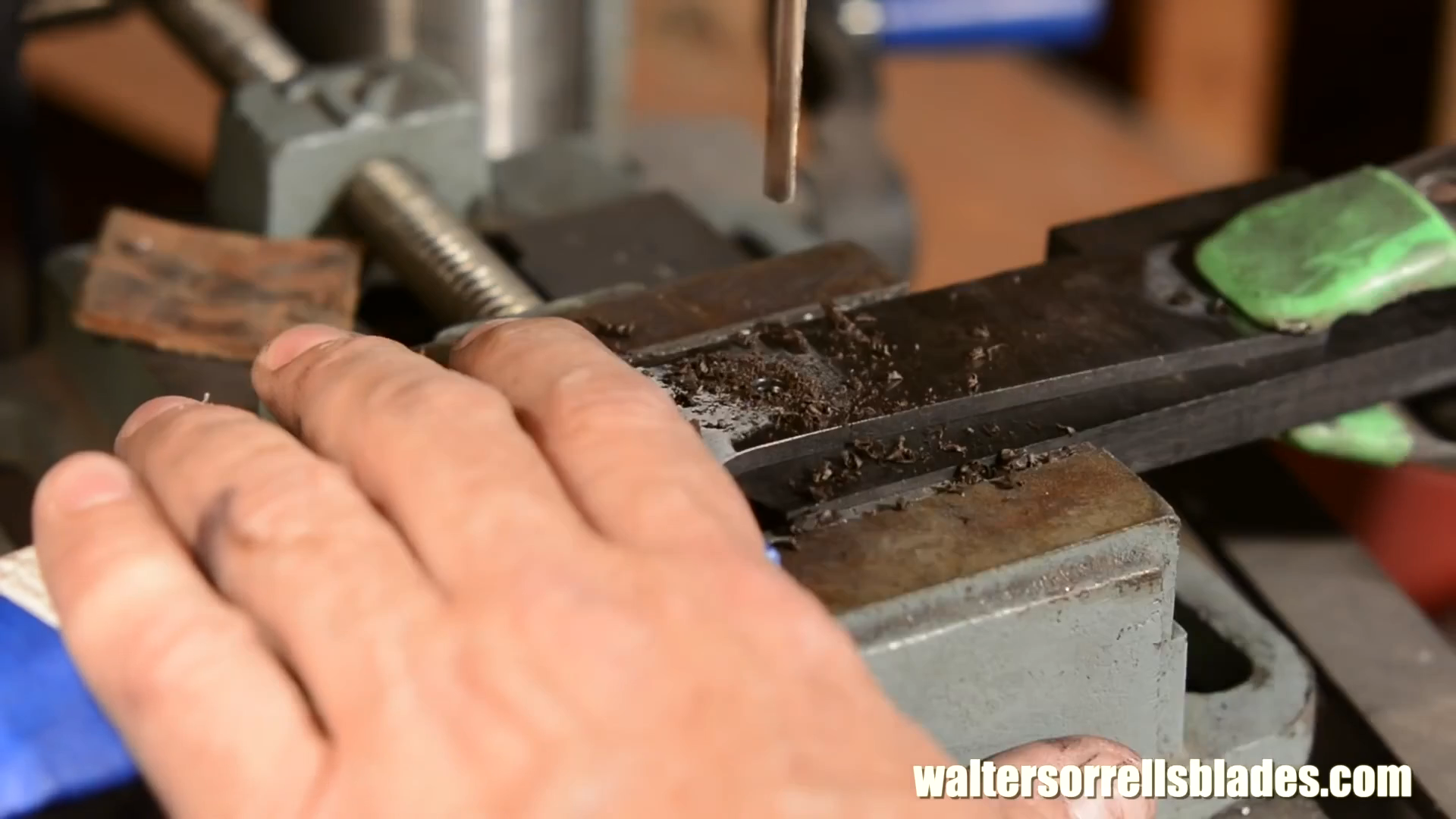} & \includegraphics[width=\linewidth]{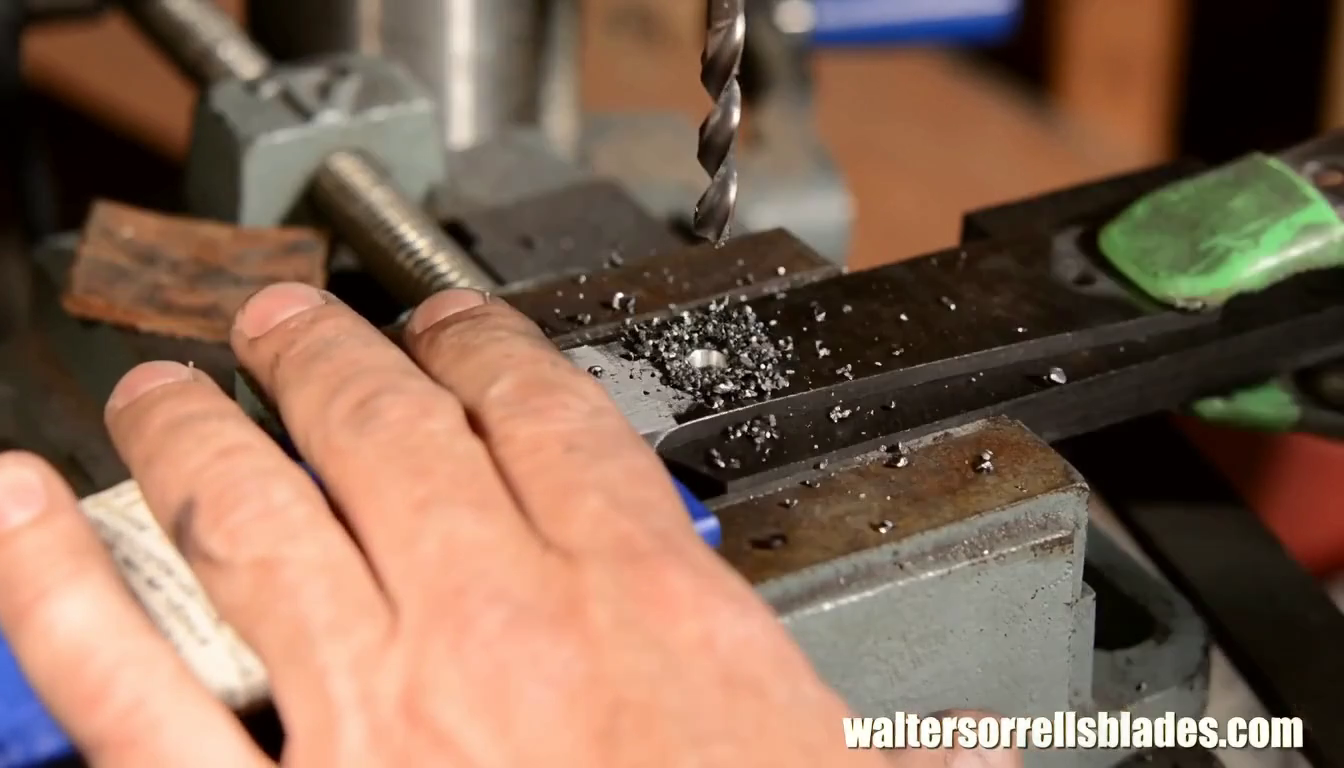} & \includegraphics[width=\linewidth]{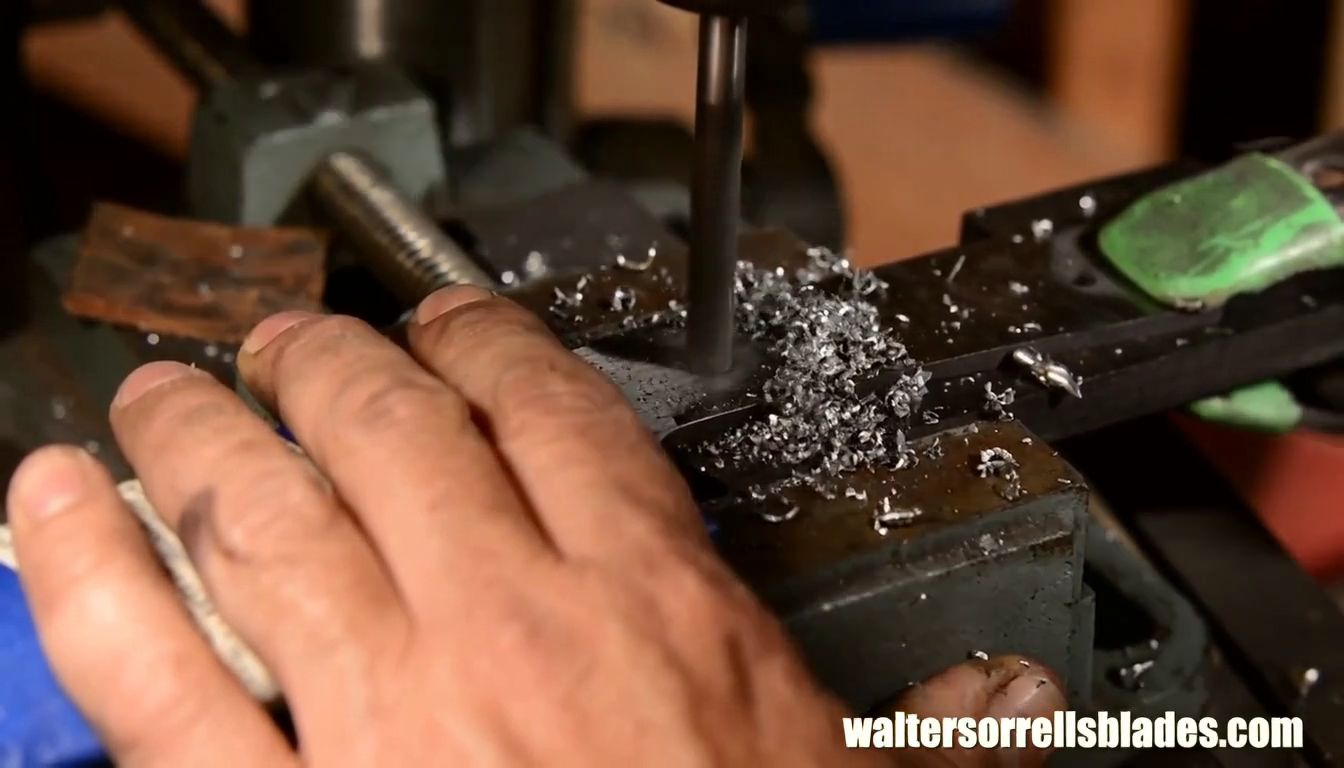}\\[2pt]
\multicolumn{4}{@{}l@{}}{\rule{0pt}{14pt}\small\textbf{(c) Plume structure and barrel orientation}\enspace\texttt{physcap9k\_09822}}\\[2pt]
{\scriptsize $6.47$\,s} & \includegraphics[width=\linewidth]{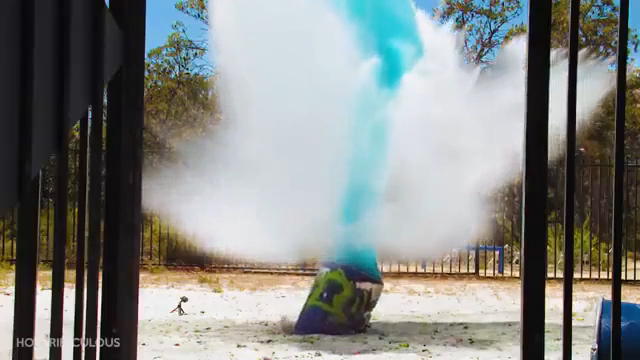} & \includegraphics[width=\linewidth]{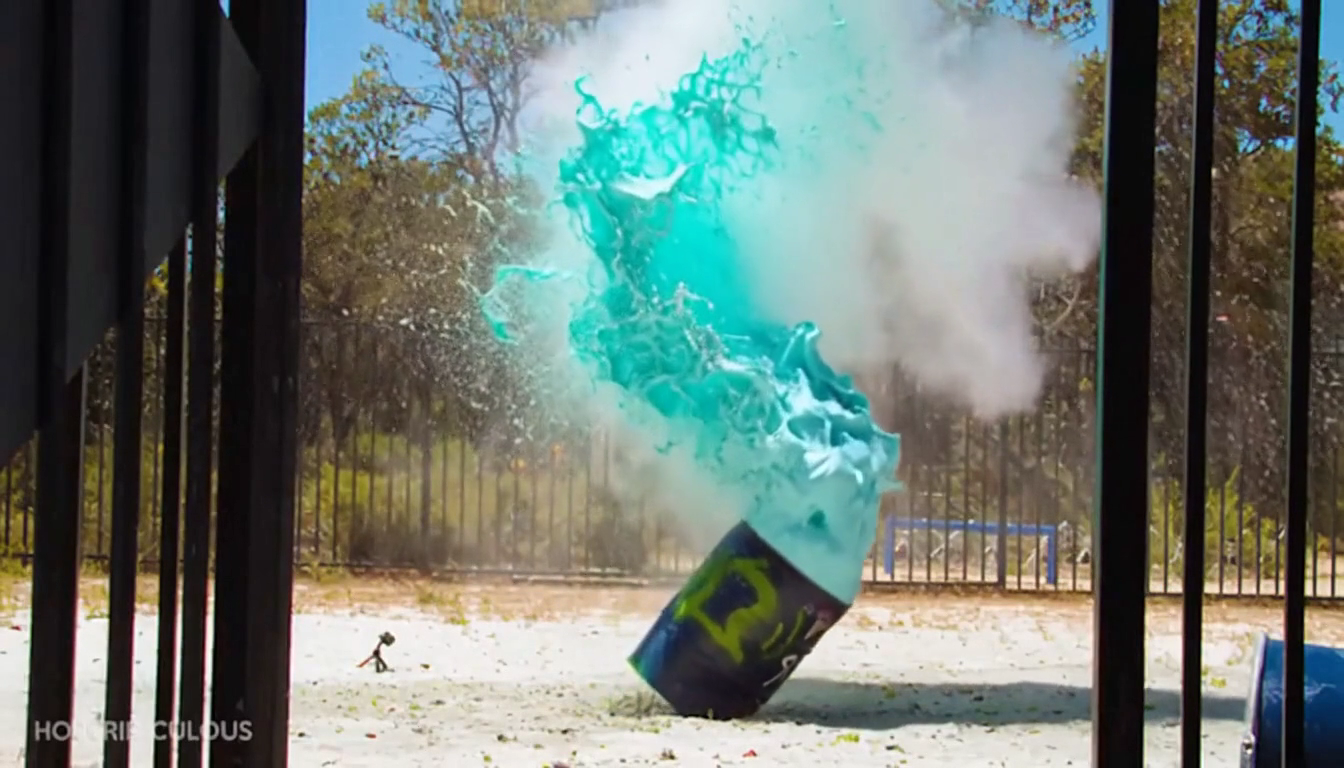} & \includegraphics[width=\linewidth]{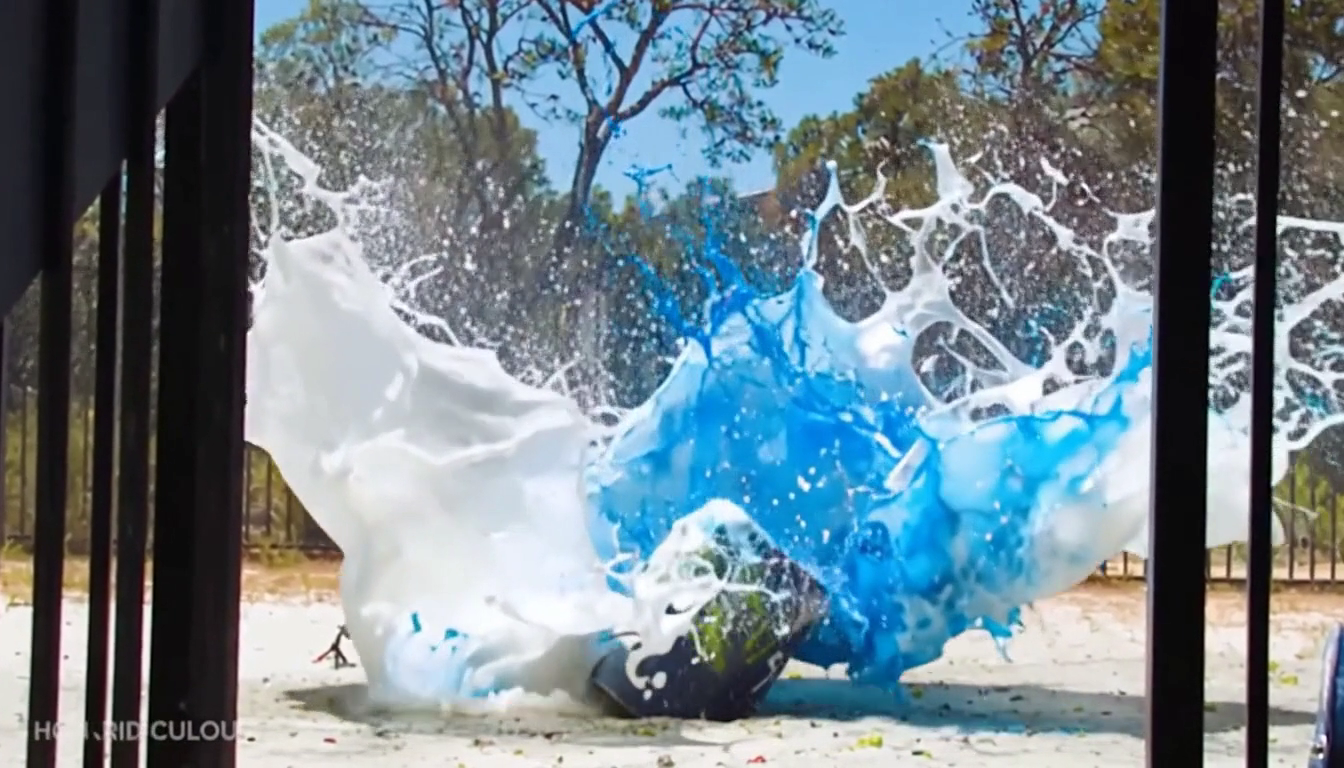}\\[2pt]
{\scriptsize $9.07$\,s} & \includegraphics[width=\linewidth]{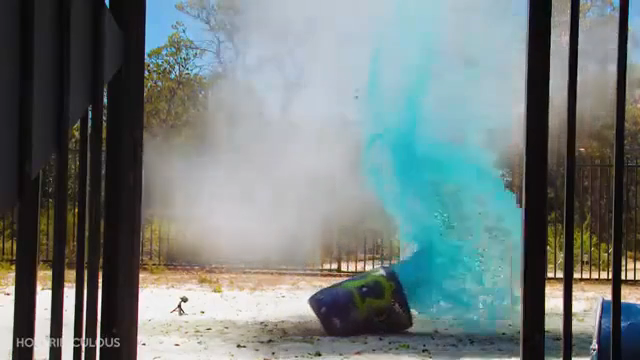} & \includegraphics[width=\linewidth]{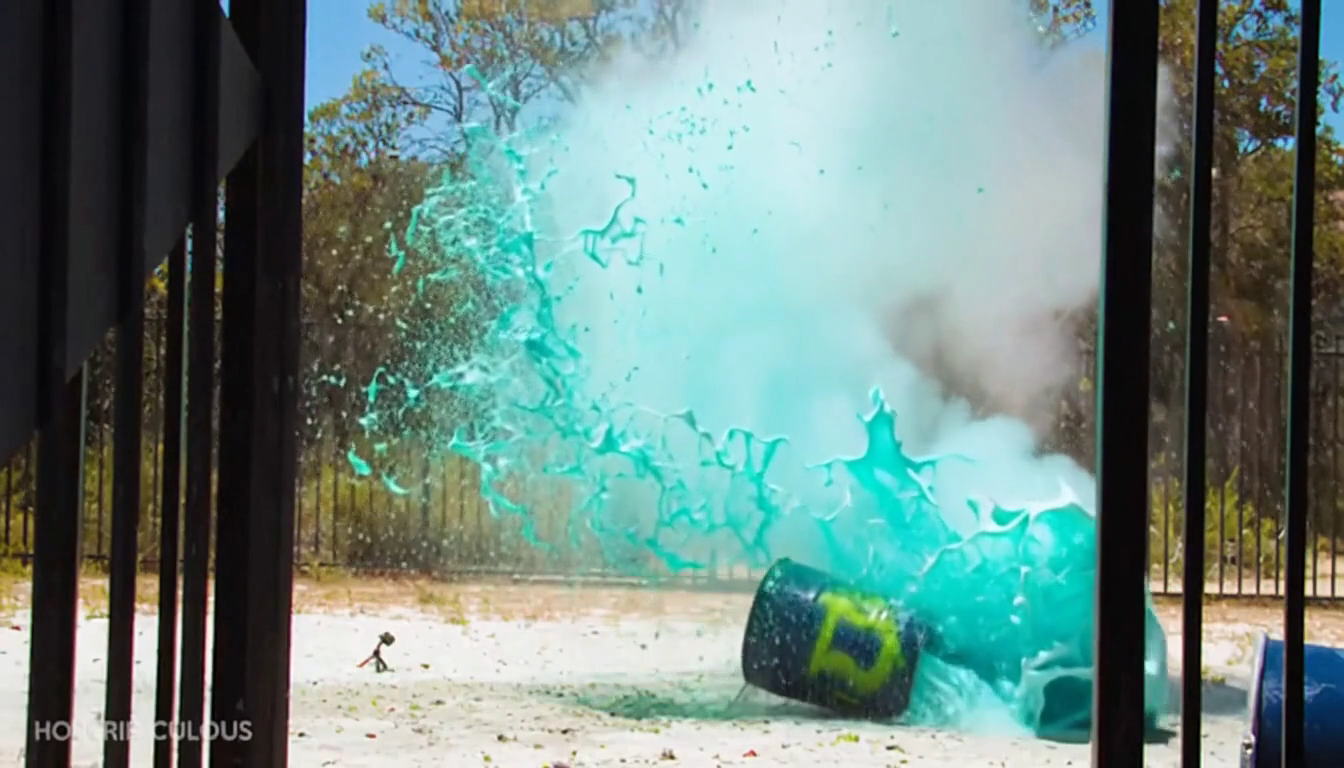} & \includegraphics[width=\linewidth]{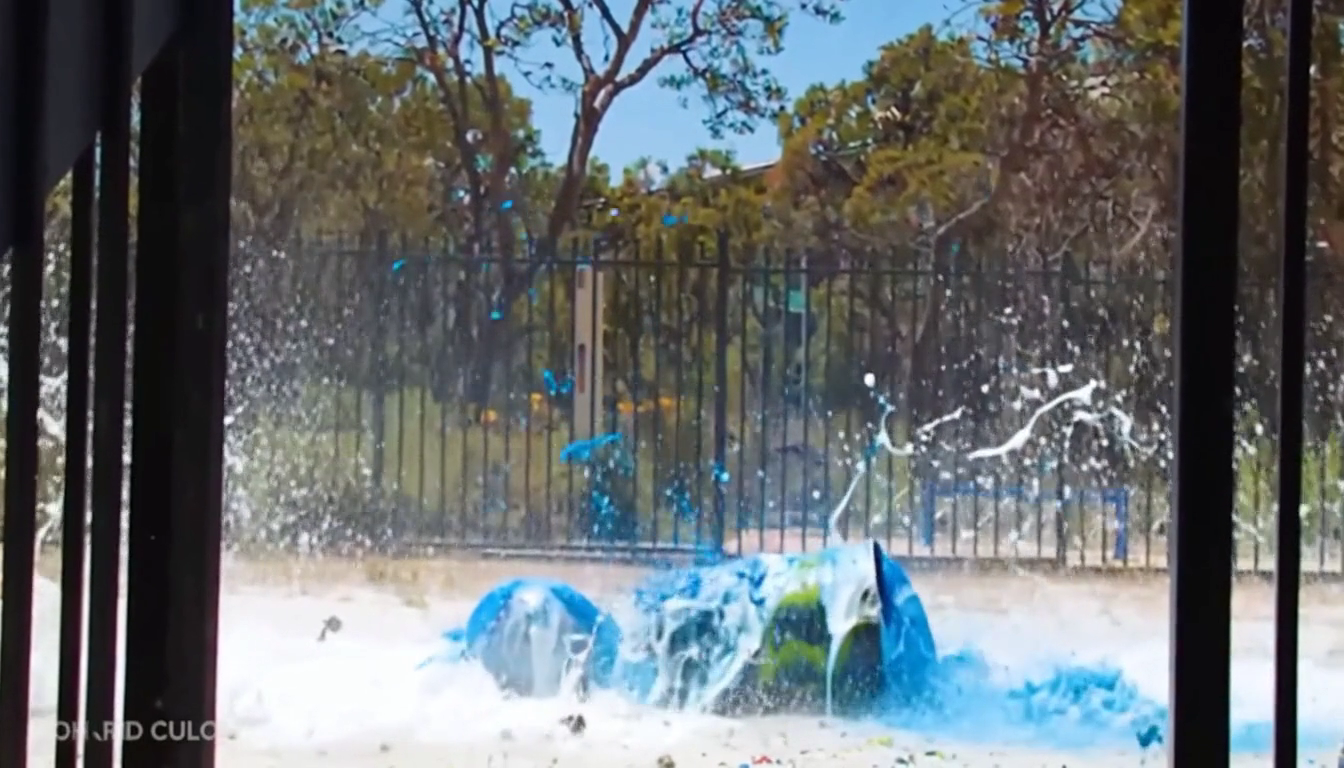}\\[2pt]
\end{tabular}
\endgroup
\caption{\textbf{Physical captions preserve state transitions and interaction details in third-person reconstruction.} Columns compare the source video, H3 conditioned on the physical caption (Gold), and H3 conditioned on the metadata caption (Meta). Each row uses the same timestamp. (a) After the board is lifted, Gold retains volumetric toy shapes, while Meta exposes flattened, sheet-like shapes; the 9.50\,s row shows both generated conditions after unloading. (b) Gold better retains localized drilling debris and the final separation of the drill bit from the workpiece. (c) Gold more closely preserves the persistent white--turquoise plume and barrel orientation, while Meta emphasizes thin liquid sheets and spray.}
\label{fig:third_reconstruction}
\end{figure}

Figure~\ref{fig:third_reconstruction} extends the comparison to deformation, tool--material interaction, and evolving material distributions. Each example juxtaposes the source video with H3 reconstructions conditioned on the physical caption (Gold) and the original metadata caption (Meta).

\paragraph{Shape after unloading (\texttt{physcap9k\_08653}).}
Gold preserves the recovery of volumetric toy shapes after the board is raised, matching the source video's transition from compression to recovery. The physical caption explicitly describes this return toward the original shape. The metadata description ends with revealing deformed toys, and Meta exposes flattened, sheet-like forms.

\paragraph{Drilling and contact termination (\texttt{physcap9k\_02537}).}
Gold better preserves localized drilling debris and the final separation of the drill bit from the workpiece. Both details appear in the physical caption, which describes debris around the hole followed by upward withdrawal. Meta describes continuous drilling, produces more extensive flying chips, and keeps the bit at the drilling location in the final displayed frame.

\paragraph{Plume structure and barrel orientation (\texttt{physcap9k\_09822}).}
Gold more closely preserves the persistent white--turquoise plume, its occlusion of the background, and the changing barrel orientation. The physical caption tracks the plume's upward and outward expansion as the barrel tips onto its side. Meta emphasizes thin blue liquid sheets and spray that recede earlier.

Together, these examples illustrate the physical content preserved by caption conditioning: participant roles, contact transitions, object-state changes, and material evolution.

\subsection{Manipulation-Policy Fine-Tuning}
\label{sec:policy}

We evaluate physical descriptions as training instructions through a controlled single-task study and a comparison across four policy families on two benchmarks. PhysCapBench scores quantify the change in description quality, and task success measures its downstream utility (Table~\ref{tab:policy_utility}).

\begin{table}[!htbp]
\centering
\caption{\textbf{Physical-description quality and policy performance under instruction enrichment.} (a) PhysCapBench scores of training descriptions. (b) Single-task success on 20 standard and 20 randomized shared scenes (40 overall), with demonstrations and training settings held fixed. (c,d) Official baseline and enriched-instruction success rates, averaged over six selected tasks in each benchmark. Gains are differences between displayed values, in score points (a) or percentage points (b--d). Table~\ref{tab:policy_task_results} reports per-task results.}
\label{tab:policy_utility}
\begingroup\small
\setlength{\tabcolsep}{6pt}\renewcommand{\arraystretch}{1.12}
\begin{tabularx}{\linewidth}{@{}Y*{3}{>{\raggedleft\arraybackslash}p{0.17\linewidth}}@{}}
\toprule
\textbf{Evaluation} & \textbf{Original} & \textbf{Enriched} & \textbf{Gain} \\
\midrule
\multicolumn{4}{@{}l}{\textbf{(a) Physical-caption score}}\\
\addlinespace[2pt]
RoboTwin 2.0 & 27.27 & \textbf{69.44} & +42.17 \\
RoboDojo Sim & 34.68 & \textbf{73.26} & +38.58 \\
\midrule
\multicolumn{4}{@{}l}{\textbf{(b) Single-task success (\%)}}\\
\addlinespace[2pt]
Standard scenes & 50.0 & \textbf{55.0} & +5.0 \\
Randomized scenes & 25.0 & \textbf{35.0} & +10.0 \\
Overall & 37.5 & \textbf{45.0} & +7.5 \\
\midrule
\multicolumn{4}{@{}l}{\textbf{(c) RoboTwin 2.0: mean success (\%)}}\\
\addlinespace[2pt]
$\pi_{0.5}$ & 50.4 & \textbf{51.9} & +1.5 \\
X-VLA & 44.4 & \textbf{45.4} & +1.0 \\
GalaxeaVLA & 29.1 & \textbf{30.1} & +1.0 \\
starVLA & 23.5 & \textbf{24.0} & +0.5 \\
\midrule
\multicolumn{4}{@{}l}{\textbf{(d) RoboDojo Sim: mean success (\%)}}\\
\addlinespace[2pt]
$\pi_{0.5}$ & 30.6 & \textbf{32.6} & +2.0 \\
X-VLA & 30.9 & \textbf{32.2} & +1.3 \\
GalaxeaVLA & 13.6 & \textbf{14.9} & +1.3 \\
starVLA & 13.7 & \textbf{14.3} & +0.6 \\
\bottomrule
\end{tabularx}
\endgroup
\end{table}

\paragraph{Physical-caption scores of training instructions.}
Instruction enrichment raises PhysCapBench scores from 27.27 to 69.44 on RoboTwin 2.0 and from 34.68 to 73.26 on RoboDojo Sim, gains of 42.17 and 38.58 points, respectively (Table~\ref{tab:policy_utility}(a)). These original and enriched descriptions provide the training instructions for the multi-task policy comparison.

\paragraph{Controlled single-task comparison.}
On an independent RoboTwin 2.0 subtask~\citep{chen2026robotwin}, we fine-tune two policies from the released $\pi_{0.5}$ base model~\citep{intelligence2025pi05} using the same 50 demonstrations and 23,098 frames. Images, robot states, action targets, and repeated sampling indices are fixed across conditions. The original condition uses the task instruction; the enriched condition replaces it for half of the training indices with a description of the demonstrated sequence, acting gripper, target placement, and post-release state.

Both policies receive 5,000 LoRA updates with batch size 32 and training seed 0. Evaluation uses 40 shared held-out initial scenes: 20 standard and 20 randomized. Each policy performs one rollout per scene using the original short instruction, current images, and robot state, with success determined by the native task criterion. Appendix~\ref{app:policy_config} details instruction construction, training, and closed-loop execution.

Instruction enrichment raises success from 37.5\% to 45.0\%, a gain of 7.5 percentage points (Table~\ref{tab:policy_utility}(b)). Gains reach 5.0 percentage points in standard scenes and 10.0 percentage points in randomized scenes. Enriched training descriptions thus improve execution with the original short task instruction.

\paragraph{Multi-task policy evaluation.}
We extend the comparison to $\pi_{0.5}$, X-VLA, GalaxeaVLA, and starVLA on six selected tasks from each of RoboTwin 2.0 and RoboDojo Sim. Table~\ref{tab:policy_utility}(c,d) compares the mean success rates of official baselines and enriched-instruction policies; Table~\ref{tab:policy_task_results} provides the per-task results.

\begin{table}[!htbp]\centering\caption{\textbf{Per-task policy success under instruction enrichment.} Entries show official baseline $\rightarrow$ enriched-instruction success rate (\%). Column headers give additional successful cases per task. Each mean summarizes the six selected tasks in its benchmark.}\label{tab:policy_task_results}\label{tab:robotwin_multitask}\label{tab:robodojo_multitask}\begingroup\footnotesize\setlength{\tabcolsep}{3pt}\renewcommand{\arraystretch}{1.18}\begin{tabularx}{\linewidth}{Y*{4}{c}}\toprule\textbf{Task} & \shortstack{\boldmath$\pi_{0.5}$\\(+3 cases)} & \shortstack{\textbf{X-VLA}\\(+2 cases)} & \shortstack{\textbf{GalaxeaVLA}\\(+2 cases)} & \shortstack{\textbf{starVLA}\\(+1 case)} \\\midrule\multicolumn{5}{l}{\textbf{(a) RoboTwin 2.0}}\\\addlinespace[2pt]Sort blocks by size & $33.0 \rightarrow 34.5$ & $40.5 \rightarrow 41.5$ & $9.5 \rightarrow 10.5$ & $7.0 \rightarrow 7.5$ \\Grasp two bottles & $64.5 \rightarrow 66.0$ & $46.0 \rightarrow 47.0$ & $49.0 \rightarrow 50.0$ & $40.0 \rightarrow 40.5$ \\Place an object in a cabinet & $30.5 \rightarrow 32.0$ & $31.0 \rightarrow 32.0$ & $21.0 \rightarrow 22.0$ & $14.5 \rightarrow 15.0$ \\Stack three bowls & $66.5 \rightarrow 68.0$ & $43.0 \rightarrow 44.0$ & $32.0 \rightarrow 33.0$ & $23.0 \rightarrow 23.5$ \\Stamp & $37.0 \rightarrow 38.5$ & $26.0 \rightarrow 27.0$ & $17.5 \rightarrow 18.5$ & $6.0 \rightarrow 6.5$ \\Press a stapler & $71.0 \rightarrow 72.5$ & $80.0 \rightarrow 81.0$ & $45.5 \rightarrow 46.5$ & $50.5 \rightarrow 51.0$ \\\addlinespace[3pt]\textbf{Mean (selected six tasks)} & $\mathbf{50.4} \rightarrow \mathbf{51.9}$ & $\mathbf{44.4} \rightarrow \mathbf{45.4}$ & $\mathbf{29.1} \rightarrow \mathbf{30.1}$ & $\mathbf{23.5} \rightarrow \mathbf{24.0}$ \\\midrule\multicolumn{5}{l}{\textbf{(b) RoboDojo Sim}}\\\addlinespace[2pt]Stack Bowls & $38.0 \rightarrow 40.0$ & $26.0 \rightarrow 27.3$ & $16.7 \rightarrow 18.0$ & $5.3 \rightarrow 6.0$ \\Fold Clothes & $24.0 \rightarrow 26.0$ & $22.0 \rightarrow 23.3$ & $8.7 \rightarrow 10.0$ & $8.0 \rightarrow 8.7$ \\Pour Liquid Into Cup & $14.7 \rightarrow 16.7$ & $13.3 \rightarrow 14.7$ & $6.0 \rightarrow 7.3$ & $9.3 \rightarrow 10.0$ \\Build Tower & $24.0 \rightarrow 26.0$ & $53.3 \rightarrow 54.7$ & $8.0 \rightarrow 9.3$ & $26.0 \rightarrow 26.7$ \\Put Bottles Into Dustbin & $69.3 \rightarrow 71.3$ & $62.0 \rightarrow 63.3$ & $33.3 \rightarrow 34.7$ & $28.0 \rightarrow 28.7$ \\Pour Balls Into Vase & $13.3 \rightarrow 15.3$ & $8.7 \rightarrow 10.0$ & $8.7 \rightarrow 10.0$ & $5.3 \rightarrow 6.0$ \\\addlinespace[3pt]\textbf{Mean (selected six tasks)} & $\mathbf{30.6} \rightarrow \mathbf{32.6}$ & $\mathbf{30.9} \rightarrow \mathbf{32.2}$ & $\mathbf{13.6} \rightarrow \mathbf{14.9}$ & $\mathbf{13.7} \rightarrow \mathbf{14.3}$ \\\bottomrule\end{tabularx}\endgroup\end{table}

All four policies improve on every selected task in both benchmarks. For $\pi_{0.5}$, mean success increases from 50.4\% to 51.9\% on RoboTwin 2.0 and from 30.6\% to 32.6\% on RoboDojo Sim. The same instruction-enrichment scheme therefore improves both physical-caption scores and policy success across the evaluated policy families, supporting PhysCapBench's use in assessing training-description quality.

\begin{figure}[!htbp]
\centering
\includegraphics[width=\linewidth]{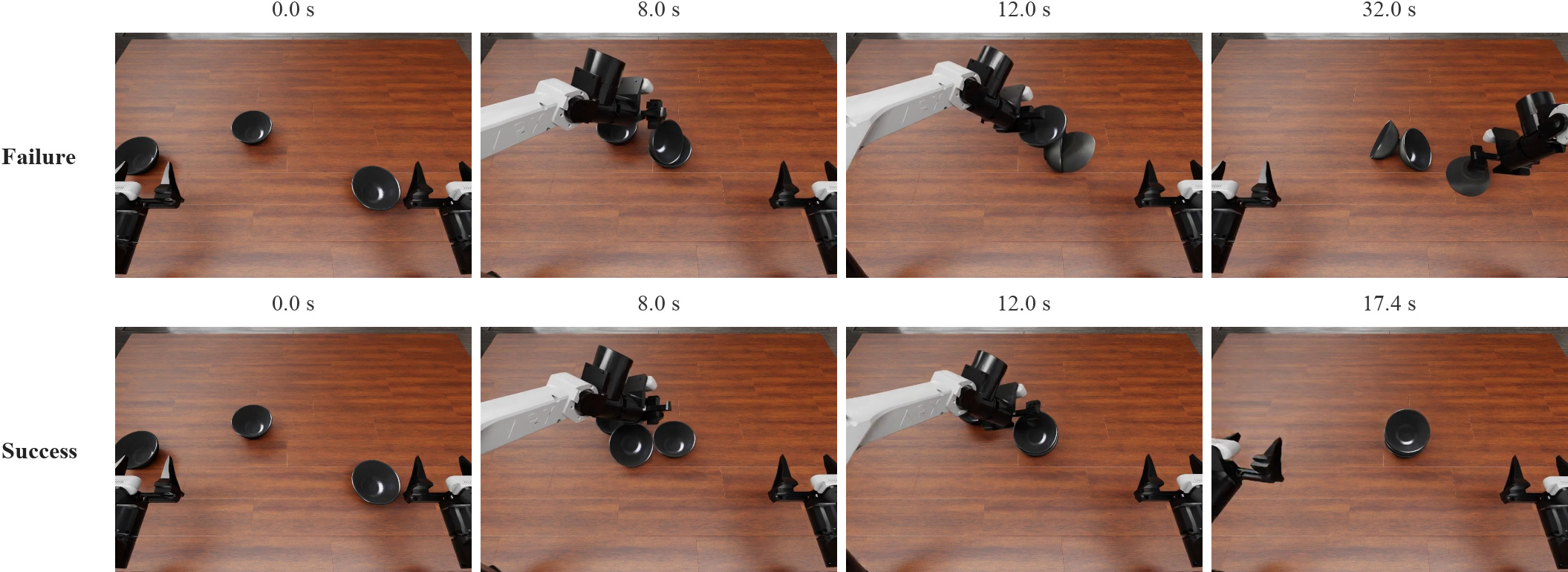}
\caption{\textbf{Robot manipulation with Training Instruction Upsampler.} Frames compare a failed rollout (top) with a successful rollout from a policy fine-tuned using Training Instruction Upsampler (bottom) on RoboDojo. The failed rollout leaves bowls tilted and overturned; the fine-tuned policy completes a stable three-bowl stack.}
\label{fig:robot_example}
\end{figure}

Figure~\ref{fig:robot_example} shows a RoboDojo bowl-stacking example in which the policy fine-tuned with Training Instruction Upsampler completes a stable three-bowl stack.

\section{Discussion and Limitations}
\label{sec:limitations}

PhysCapBench connects physical caption assessment with the data and applications used in world modeling. Its current scope suggests three directions for further development.

\subsection{Data and Annotation Coverage}
The three domains provide complementary observations of physical scenes and manipulation, supported by domain-adapted annotation and human cross-checking. Coverage remains tied to the selected video sources and their observable interactions. Extending scene diversity and temporal coverage would broaden the range of physical processes represented while retaining the current annotation structure.

\subsection{Evaluation Consistency}
The evaluation uses a fixed text judge to assess agreement with physical references. Scores depend on the reference content and the judge's interpretation of the rubric. Comparisons with human ratings and additional judges would help characterize scoring consistency across domains and captioning styles.

\subsection{Downstream Applications}
Video reconstruction and manipulation-policy fine-tuning demonstrate the utility of physical descriptions in the evaluated settings. The contribution of each physical dimension to these gains remains to be examined. Controlled changes to individual description components and broader task coverage would help identify which physical information is most useful for different downstream models.

\section{Conclusion}
\label{sec:conclusion}

We present PhysCapBench, a benchmark for physically grounded video captioning across third-person scenes, egocentric human manipulation, and multi-view human-operated handheld-gripper demonstrations. Human-verified annotations and seven physical evaluation dimensions support systematic assessment of interactions and state evolution. Experiments show improved video reconstruction with physical captions and higher manipulation success with enriched training instructions, alongside gains in PhysCapBench scores. PhysCapBench provides a basis for developing and evaluating physical descriptions as language supervision for world models and embodied learning.

\phantomsection\label{sec:references}

\appendix
\setlength{\parskip}{0.35\baselineskip}
\titlespacing*{\section}{0pt}{2.5ex plus 2pt minus 2pt}{6pt}
\titlespacing*{\subsection}{0pt}{2ex plus 2pt minus 1pt}{3pt}
\addtocontents{toc}{\protect\physcapappendixtoc}

\section{Annotation Schema and Domain Profiles}
\label{app:schema}

The annotation schema connects entity identities and states to events, sequence-level relations, and visual evidence. Tables~\ref{tab:schema}--\ref{tab:domain_fields} expand the representation in Section~\ref{sec:physical_representation}, using the field names of the annotation schemas and interfaces. Captions express the recorded physical process in connected text and provide references for the evaluation in Section~\ref{sec:caption_protocol}.

\subsection{Entities and Initial States}

Entity records establish the identities and starting configuration needed to describe subsequent changes. The \texttt{objects\_and\_states} record links each entity to its observed states and supporting frames, while initial spatial relations specify the scene configuration. Table~\ref{tab:schema} lists these fields and their domain-specific evidence references.

\begingroup\small
\setlength{\tabcolsep}{5pt}\renewcommand{\arraystretch}{1.15}
\begin{xltabular}{\linewidth}{>{\raggedright\arraybackslash}p{0.45\linewidth}Y}
\caption{\textbf{Entity and initial-scene fields.} The first seven rows describe entity records; the remaining rows describe initial spatial relations and their evidence.}
\label{tab:schema}\\
\toprule
\textbf{Field / record} & \textbf{Meaning} \\
\midrule
\endfirsthead
\toprule
\textbf{Field / record} & \textbf{Meaning} \\
\midrule
\endhead

\texttt{object\_\allowbreak id}, \texttt{object\_\allowbreak name} & Entity identifier and descriptive name. \\
\texttt{initial\_\allowbreak state}, \texttt{final\_\allowbreak state} & Descriptions of the observed initial and final states. \\
\texttt{local\_\allowbreak observation\_\allowbreak ids} & Links to local observations associated with an entity. \\
\texttt{instance\_\allowbreak group}, \texttt{instance\_\allowbreak index} & Group and instance identifiers for distinguishing similar entities. \\
\texttt{identity\_\allowbreak basis} & Recorded basis for associating observations with the same entity. \\
\texttt{identity\_\allowbreak reference\_\allowbreak frame\_\allowbreak indices} & Source frames supporting the entity identification. \\
\texttt{identity\_\allowbreak visible\_\allowbreak frame\_\allowbreak ranges} & Frame intervals in which the entity is visible. \\
\texttt{initial\_\allowbreak spatial\_\allowbreak relations} & Scene-level relations among entities at the start of the recording. \\
\texttt{subject}, \texttt{reference\_\allowbreak object}, \texttt{relation} & Within an initial relation: the related entity, its reference entity, and their spatial relationship. \\
\texttt{evidence\_\allowbreak frame\_\allowbreak indices} (Ego), \texttt{witness\_\allowbreak ids} (UMI) & Evidence attached to an initial relation: source frames in Ego and witness references in UMI. \\
\bottomrule
\end{xltabular}
\endgroup
\subsection{Event Fields}

Event records describe how entities participate in a physical process: as actors, instruments, targets, moving entities, or entities undergoing state changes. An entity may occupy multiple roles. Table~\ref{tab:event_fields} groups the fields by timing, participants, motion, interaction, and event relations; Table~\ref{tab:domain_fields} specifies domain-specific evidence and state-transition links.

\begingroup\small
\setlength{\tabcolsep}{5pt}\renewcommand{\arraystretch}{1.15}
\begin{xltabular}{\linewidth}{>{\raggedright\arraybackslash}p{0.45\linewidth}Y}
\caption{\textbf{Detailed event fields.} Fields appear within each \texttt{event\_chain} entry. The table specifies their semantic roles; domain profiles retain their own value vocabularies and evidence structures.}
\label{tab:event_fields}\\
\toprule
\textbf{Field / record} & \textbf{Meaning} \\
\midrule
\endfirsthead
\toprule
\textbf{Field / record} & \textbf{Meaning} \\
\midrule
\endhead

\texttt{event\_\allowbreak id} & Identifier used to link the event to other records. \\
\texttt{frame\_\allowbreak span}, \texttt{time\_\allowbreak span} & Event interval in source frames and corresponding seconds. \\
\texttt{boundary\_\allowbreak tolerance\_\allowbreak frames}, \texttt{boundary\_\allowbreak source} & Boundary tolerance and provenance of the proposed boundary. \\
\texttt{actors} & Entities that perform the action. \\
\texttt{instrument\_\allowbreak objects} & Objects used as instruments by the acting entities. \\
\texttt{action\_\allowbreak type} & Action category assigned to the event. \\
\texttt{target\_\allowbreak object} & Object directly targeted by the action, when applicable. \\
\texttt{motion\_\allowbreak subjects} & Entities whose motion is described. \\
\texttt{motion\_\allowbreak type}, \texttt{motion\_\allowbreak direction} & Type and direction of the recorded motion. \\
\texttt{motion\_\allowbreak reference\_\allowbreak object} & Reference entity used for an object-relative motion description. \\
\texttt{contact\_\allowbreak subjects}, \texttt{contact\_\allowbreak object} & Entities participating in contact and the contacted counterpart. \\
\texttt{contact\_\allowbreak relation} & Contact or interaction label; its vocabulary depends on the domain profile. \\
\texttt{state\_\allowbreak change\_\allowbreak subjects}, \texttt{state\_\allowbreak change} & Entities whose states change and the category of that change. \\
\texttt{spatial\_\allowbreak relation\_\allowbreak changes} & Changes in spatial relationships during the event. \\
\texttt{temporal\_\allowbreak relation}, \texttt{causal\_\allowbreak relation} & Temporal or supported causal relation to another event. \\
\texttt{evidence} & Description of the observations supporting the event. \\
\bottomrule
\end{xltabular}
\endgroup
\subsection{Sequence Records and Domain Extensions}

Sequence records combine events into a temporal account that includes persistent states and overlapping actions. Mechanism records link explanations to supporting events, and domain extensions add explicit state transitions and camera-specific evidence (Table~\ref{tab:domain_fields}). Mechanism participants use \texttt{actor} in the UMI profile and \texttt{actor\_ids} in the egocentric profile; event-level \texttt{actors} lists the acting entities.

\begingroup\small
\setlength{\tabcolsep}{5pt}\renewcommand{\arraystretch}{1.15}
\begin{xltabular}{\linewidth}{>{\raggedright\arraybackslash}p{0.15\linewidth}>{\raggedright\arraybackslash}p{0.39\linewidth}Y}
\caption{\textbf{Sequence-level records and domain-specific fields.} Ego and UMI rows identify the profiles in which the listed field forms occur. The caption fields store the natural-language descriptions associated with the structured records.}
\label{tab:domain_fields}\\
\toprule
\textbf{Scope} & \textbf{Fields / record} & \textbf{Meaning} \\
\midrule
\endfirsthead
\toprule
\textbf{Scope} & \textbf{Fields / record} & \textbf{Meaning} \\
\midrule
\endhead

Timeline & \texttt{timeline\_\allowbreak segments}, \texttt{linked\_\allowbreak event\_\allowbreak ids} & Segments organize the sequence and identify active events; source-frame and time intervals locate each segment. \\
Mechanisms & \texttt{physical\_\allowbreak mechanism}, \texttt{mechanism\_\allowbreak type}, \texttt{supporting\_\allowbreak event\_\allowbreak ids} & Mechanism statements are connected to their supporting events. \\
Ego & \texttt{mechanism\_\allowbreak id}, \texttt{actor\_\allowbreak ids}, \texttt{target\_\allowbreak object\_\allowbreak ids}, \texttt{statement}, \texttt{statement\_\allowbreak zh} & Fields within each mechanism record identify participants and retain English and Chinese descriptions. \\
UMI & \texttt{actor}, \texttt{description}, \texttt{witness\_\allowbreak ids} & Fields within the mechanism object identify an actor, describe the mechanism, and link visual evidence. \\
UMI & \texttt{state\_\allowbreak delta\_\allowbreak ids}, \texttt{state\_\allowbreak deltas} & Event links to explicit state-transition records. \\
UMI & \texttt{removed}, \texttt{added}, \texttt{predicate}, \texttt{subject}, \texttt{object}, \texttt{polarity} & State-transition records contain removed and added state atoms, each encoding a relation and its polarity. \\
Ego & \texttt{evidence\_\allowbreak frame\_\allowbreak indices}, \texttt{manual\_\allowbreak review\_\allowbreak status} & Frame evidence and review status associated with an event. \\
UMI evidence & \texttt{witness\_\allowbreak ids}, \texttt{rgb\_\allowbreak witnesses} & Annotation records reference witness IDs; witness details reside in a separate evidence sidecar with view, frames, observability, and review status. \\
UMI & \texttt{entity\_\allowbreak type} & Entity category in the handheld-gripper profile. \\
Taxonomy & \texttt{taxonomy\_\allowbreak labels} & Domain-specific labels summarizing the annotated content. \\
Ego caption & \texttt{goldCaptionEn}, \texttt{goldCaptionZh} & English and Chinese reference-caption text. \\
UMI caption & \texttt{caption.short}, \texttt{caption.dense}, \texttt{caption.language} & Short and dense descriptions, with an explicit language field. \\
\bottomrule
\end{xltabular}
\endgroup
\subsection{Temporal and Observation Conventions}

Timing is expressed in source-frame coordinates. Event intervals use half-open ranges $[a,b)$, and dividing their boundaries by the source frame rate gives the corresponding times. Evidence frames must lie within the recording. Events may overlap to represent simultaneous actions; timeline segments partition the sequence and identify the events active in each segment. Entity references distinguish actors, moving entities, manipulated objects, and affected targets, with hands acting directly and objects serving as instruments when they transmit the action.

Observation records preserve actor identity and the viewpoint supporting each annotation. Egocentric annotations use anatomical left and right hand identities and record ambiguity when handedness is not visible. Handheld-gripper annotations identify left and right grippers and associate observations with the high, left-wrist, or right-wrist view. State transitions record removed and added state atoms; linked witness records specify the view, source frames, observability, and review status for cross-view verification. Visibility and physical state occupy separate fields.

\section{Data Coverage and Video Examples}
\label{app:selection}

Dataset coverage is organized by physical phenomena, scenes, tasks, and trajectory profiles, followed by source-video examples. The distributions use manifest labels and count videos or demonstration episodes. Egocentric physical-phenomenon tags allow multiple labels per video.

\subsection{Third-Person Coverage}
\label{app:third_coverage}

The 500 third-person videos span eight primary physical categories and 16 scene categories, covering fluid and granular processes, motion and contact, and deformation across home, workshop, public, and outdoor settings. Each video contributes once to each distribution in Tables~\ref{tab:third_physical} and~\ref{tab:third_scenes}; its annotations record additional phenomena present in the sequence.

\begingroup\small\renewcommand{\arraystretch}{1.13}
\begin{xltabular}{\linewidth}{Yrr}
\caption{\textbf{Third-person primary physical coverage.} Each video has one label in this grouping; counts sum to 500.}
\label{tab:third_physical}\\
\toprule
\textbf{Primary coverage category} & \textbf{Count} & \textbf{Share} \\
\midrule
\endfirsthead
\toprule
\textbf{Primary coverage category} & \textbf{Count} & \textbf{Share} \\
\midrule
\endhead

Fluid phenomena & 161 & 32.2\% \\
Gravity and support & 80 & 16.0\% \\
Granular phenomena & 65 & 13.0\% \\
Rotation and friction & 60 & 12.0\% \\
Manipulation and state change & 43 & 8.6\% \\
Collision and momentum & 35 & 7.0\% \\
Combustion, gas, and other phenomena & 28 & 5.6\% \\
Deformation and breakage & 28 & 5.6\% \\
\addlinespace[2pt]
\textbf{Total} & \textbf{500} & \textbf{100.0\%} \\
\bottomrule
\end{xltabular}
\endgroup

\begingroup\small\renewcommand{\arraystretch}{1.13}
\begin{xltabular}{\linewidth}{Yrr}
\caption{\textbf{Third-person scene coverage.} Each video has one label in this grouping; counts sum to 500.}
\label{tab:third_scenes}\\
\toprule
\textbf{Scene category} & \textbf{Count} & \textbf{Share} \\
\midrule
\endfirsthead
\toprule
\textbf{Scene category} & \textbf{Count} & \textbf{Share} \\
\midrule
\endhead

Home & 155 & 31.0\% \\
Industrial workshop & 118 & 23.6\% \\
Public space & 73 & 14.6\% \\
Natural outdoor setting & 53 & 10.6\% \\
Waterfront and coast & 34 & 6.8\% \\
Laboratory & 16 & 3.2\% \\
Urban street & 14 & 2.8\% \\
Commercial setting & 10 & 2.0\% \\
Fields and countryside & 9 & 1.8\% \\
Mountains and forest & 8 & 1.6\% \\
Snow-covered setting & 4 & 0.8\% \\
Other outdoor setting & 2 & 0.4\% \\
Indoor gymnasium & 1 & 0.2\% \\
Indoor ski venue & 1 & 0.2\% \\
Indoor sports venue & 1 & 0.2\% \\
Aerial setting & 1 & 0.2\% \\
\addlinespace[2pt]
\textbf{Total} & \textbf{500} & \textbf{100.0\%} \\
\bottomrule
\end{xltabular}
\endgroup

\subsection{Egocentric Coverage}
\label{app:ego_coverage}

Egocentric selection balances task coverage with diversity in scenarios, actions, and objects. From 1,413 individually captioned videos in an index of 338,234 videos, we select four examples per task across 118 tasks. The remaining 28 slots broaden coverage of descriptions, scenarios, actions, objects, and source archives. Every selected video includes a source caption and a paired hand-tracking file. Table~\ref{tab:ego_scenarios} summarizes the 15 scenario groups. The source-archive distribution is 111 videos from EgoDex test, 83 from part1, 76 from part5, 75 from part3, 64 from part4, 49 from extra, and 42 from part2.

\begingroup\small\renewcommand{\arraystretch}{1.13}
\begin{xltabular}{\linewidth}{Yrr}
\caption{\textbf{Egocentric scenario coverage.} Categories are derived from task names, with metadata captions used for disambiguation. Counts sum to 500.}
\label{tab:ego_scenarios}\\
\toprule
\textbf{Scenario group} & \textbf{Videos} & \textbf{Share} \\
\midrule
\endfirsthead
\toprule
\textbf{Scenario group} & \textbf{Videos} & \textbf{Share} \\
\midrule
\endhead

Board games and puzzles & 53 & 10.6\% \\
Container opening, closing, and storage & 52 & 10.4\% \\
Kitchen cleaning and tableware organization & 46 & 9.2\% \\
Cooking and food preparation & 45 & 9.0\% \\
Electronic-device interaction & 44 & 8.8\% \\
Paper crafts, writing, and drawing & 44 & 8.8\% \\
Furniture assembly and disassembly & 36 & 7.2\% \\
Blocks and structural assembly & 34 & 6.8\% \\
Textiles, cords, and soft objects & 29 & 5.8\% \\
Personal care and dressing & 28 & 5.6\% \\
Cleaning and surface maintenance & 21 & 4.2\% \\
Toys and fine hand manipulation & 21 & 4.2\% \\
Ball handling, throwing, and catching & 18 & 3.6\% \\
Tools, locks, and fasteners & 16 & 3.2\% \\
General tabletop transport and organization & 13 & 2.6\% \\
\bottomrule
\end{xltabular}
\endgroup

Physical-phenomenon tags describe processes that co-occur across scenarios (Table~\ref{tab:ego_phenomena}). Multiple tags per video yield a total tag count above 500. These video-level labels summarize coverage; event annotations resolve the processes over time.

\begingroup\small\renewcommand{\arraystretch}{1.13}
\begin{xltabular}{\linewidth}{Yrr}
\caption{\textbf{Egocentric physical-phenomenon coverage tags.} Videos can carry multiple coverage tags; percentages use 500 videos as the denominator.}
\label{tab:ego_phenomena}\\
\toprule
\textbf{Coverage tag} & \textbf{Videos} & \textbf{Share} \\
\midrule
\endfirsthead
\toprule
\textbf{Coverage tag} & \textbf{Videos} & \textbf{Share} \\
\midrule
\endhead

Rigid-object transport and placement & 267 & 53.4\% \\
Insertion and extraction constraints & 145 & 29.0\% \\
Connection and separation & 103 & 20.6\% \\
Stacking and stability & 102 & 20.4\% \\
Deformation and folding & 58 & 11.6\% \\
Friction and wiping & 53 & 10.6\% \\
Rolling and collision & 36 & 7.2\% \\
Contact and support & 34 & 6.8\% \\
Rotation and threaded fastening & 32 & 6.4\% \\
Liquid flow & 28 & 5.6\% \\
Gravity, falling, throwing, and catching & 26 & 5.2\% \\
Thermal processes and cooking & 12 & 2.4\% \\
Cutting, tearing, and breakage & 4 & 0.8\% \\
\bottomrule
\end{xltabular}
\endgroup

Action and object classifications further characterize manipulation diversity. Table~\ref{tab:ego_actions_objects} reports 18 action families and 14 object categories. Each video receives one label in each classification, alongside its multi-label physical-phenomenon tags.

\begingroup\small\setlength{\tabcolsep}{4pt}\renewcommand{\arraystretch}{1.17}
\begin{xltabular}{\linewidth}{YrYr}
\caption{\textbf{Egocentric action and object coverage.} The two halves present independently sorted distributions over the same 500 videos, with one action-family label and one object-category label per video.}
\label{tab:ego_actions_objects}\\
\toprule
\textbf{Action family} & \textbf{Videos} & \textbf{Object category} & \textbf{Videos} \\
\midrule
\endfirsthead
\toprule
\textbf{Action family} & \textbf{Videos} & \textbf{Object category} & \textbf{Videos} \\
\midrule
\endhead

Insertion / removal & 92 & Blocks, puzzles, and game pieces & 68 \\
Assembly / disassembly & 57 & Tableware and kitchen containers & 60 \\
Fastening / unfastening & 41 & Furniture and structural parts & 53 \\
Grasping / transport / throw--catch & 40 & Paper, cards, and stationery & 50 \\
Folding / deformation / wrapping & 38 & Food, ice, and liquids & 46 \\
Games / fine manipulation & 33 & Tools, locks, and fasteners & 40 \\
Stacking / unstacking & 29 & Electronic devices and accessories & 39 \\
Cleaning / wiping & 28 & Textiles, cords, and soft objects & 37 \\
Arrangement / organization & 26 & Storage containers and bags & 31 \\
Rolling / rotation / flipping & 22 & Small toys and parts & 21 \\
Writing / drawing / input & 20 & Balls and objects for throwing/catching & 18 \\
Washing & 20 & General tabletop objects & 13 \\
Cooking / heating & 16 & Personal-care and wearable items & 12 \\
Pouring / scooping / dispensing & 12 & Cleaning items and surfaces & 12 \\
Connection / disconnection & 10 &  &  \\
Dressing / personal care & 8 &  &  \\
Measurement / tool use & 4 &  &  \\
Food handling & 4 &  &  \\
\addlinespace[2pt]
\textbf{Total} & \textbf{500} & \textbf{Total} & \textbf{500} \\
\bottomrule
\end{xltabular}
\endgroup

\subsection{Handheld-Gripper Coverage}
\label{app:umi_coverage}

Handheld-gripper selection combines data-validity checks with task diversity. We check video availability and decoding, synchronization with metadata and trajectories, sufficient duration, and finite action/state values, retaining 58,318 eligible episodes from 66,806 recordings. We then select 500 episodes across the task categories in Table~\ref{tab:umi_tasks}. Episodes without task hints are annotated from their visual observations.

\begingroup\small\renewcommand{\arraystretch}{1.15}
\begin{xltabular}{\linewidth}{Yrr}
\caption{\textbf{Task categories of the handheld-gripper dataset.} Categories summarize source task intent for selection; annotation determines the observed process from video.}
\label{tab:umi_tasks}\\
\toprule
\textbf{Task category} & \textbf{Episodes} & \textbf{Share} \\
\midrule
\endfirsthead
\toprule
\textbf{Task category} & \textbf{Episodes} & \textbf{Share} \\
\midrule
\endhead

Placement, storage, and return & 55 & 11.0\% \\
Grasping, picking, and removal & 55 & 11.0\% \\
Opening, closing, pulling, and switch operation & 55 & 11.0\% \\
Position and orientation adjustment & 55 & 11.0\% \\
Clothing and flexible-object organization & 54 & 10.8\% \\
Table setting and tableware clearing & 54 & 10.8\% \\
Wiping, sweeping, and disposal & 54 & 10.8\% \\
Tool and device interaction & 54 & 10.8\% \\
Transfer, pouring, and replenishment & 54 & 10.8\% \\
Missing task hint & 10 & 2.0\% \\
\bottomrule
\end{xltabular}
\endgroup

Trajectory profiles complement task categories by characterizing overall motion, rotation, opening range, and left--right activity (Table~\ref{tab:umi_profiles}). Each episode receives one profile during selection. Episodes range from 4.0 to 89.7 seconds, with a median duration of 28.1 seconds, a 90th percentile of 64.6 seconds, and approximately 4.82 hours in total.

\begingroup\small\renewcommand{\arraystretch}{1.13}
\begin{xltabular}{\linewidth}{Yrr}
\caption{\textbf{Handheld-gripper trajectory profiles.} Each episode has one label in this grouping; counts sum to 500.}
\label{tab:umi_profiles}\\
\toprule
\textbf{Trajectory profile} & \textbf{Count} & \textbf{Share} \\
\midrule
\endfirsthead
\toprule
\textbf{Trajectory profile} & \textbf{Count} & \textbf{Share} \\
\midrule
\endhead

High motion & 108 & 21.6\% \\
Right-dominant motion & 77 & 15.4\% \\
Balanced bimanual motion & 65 & 13.0\% \\
Rotation-rich trajectories & 65 & 13.0\% \\
Left-dominant motion & 64 & 12.8\% \\
Large gripper-opening range & 61 & 12.2\% \\
Low motion & 60 & 12.0\% \\
\addlinespace[2pt]
\textbf{Total} & \textbf{500} & \textbf{100.0\%} \\
\bottomrule
\end{xltabular}
\endgroup

\subsection{Video Examples across Three Domains}
\label{app:video_examples}

\begin{figure}[H]
\centering\small
\begin{minipage}{0.88\linewidth}
\noindent\textbf{(a) Surface compaction}\hfill{\scriptsize\texttt{physcap9k\_02382}}\par\smallskip
\begin{minipage}[t]{0.241\linewidth}\centering
\includegraphics[width=\linewidth]{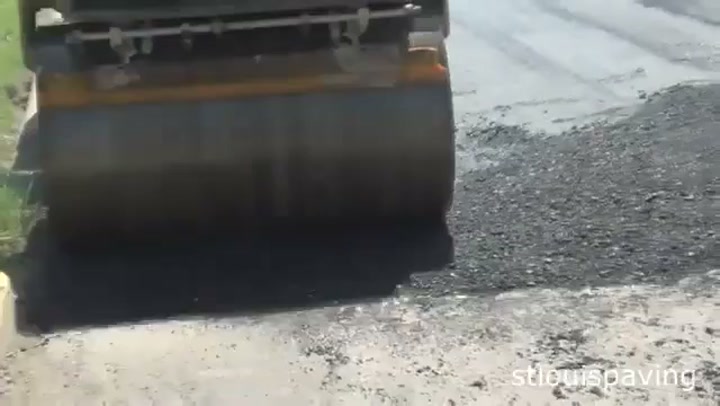}
{\scriptsize $t=0.00\,\mathrm{s}$}\end{minipage}
\hfill
\begin{minipage}[t]{0.241\linewidth}\centering
\includegraphics[width=\linewidth]{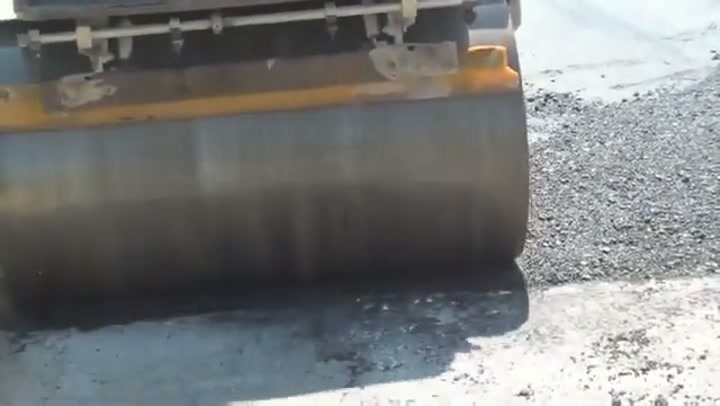}
{\scriptsize $t=2.73\,\mathrm{s}$}\end{minipage}
\hfill
\begin{minipage}[t]{0.241\linewidth}\centering
\includegraphics[width=\linewidth]{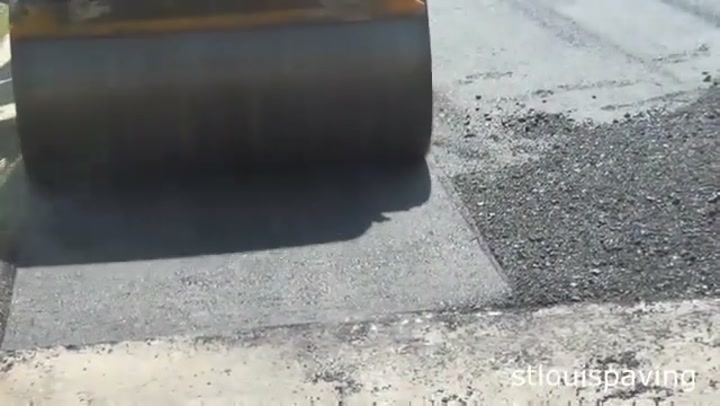}
{\scriptsize $t=5.50\,\mathrm{s}$}\end{minipage}
\hfill
\begin{minipage}[t]{0.241\linewidth}\centering
\includegraphics[width=\linewidth]{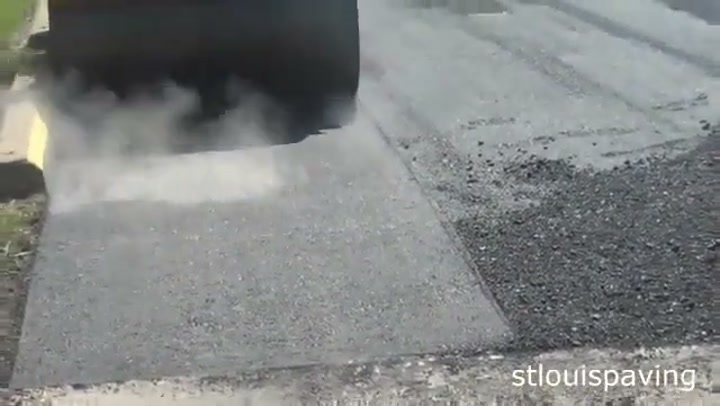}
{\scriptsize $t=8.23\,\mathrm{s}$}\end{minipage}
\par\medskip

\noindent\textbf{(b) Ejection from a barrel}\hfill{\scriptsize\texttt{physcap9k\_09822}}\par\smallskip
\begin{minipage}[t]{0.241\linewidth}\centering
\includegraphics[width=\linewidth]{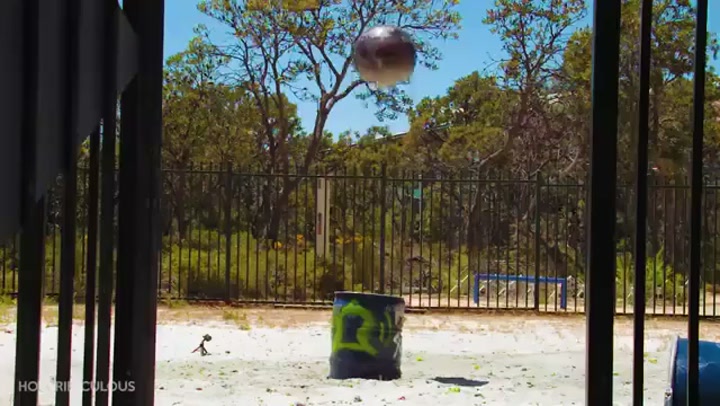}
{\scriptsize $t=0.00\,\mathrm{s}$}\end{minipage}
\hfill
\begin{minipage}[t]{0.241\linewidth}\centering
\includegraphics[width=\linewidth]{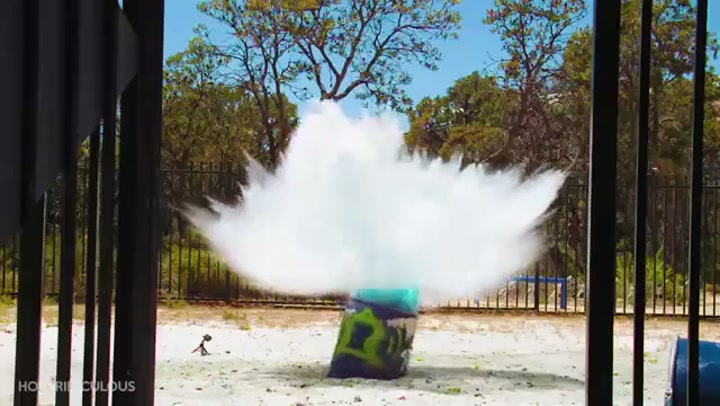}
{\scriptsize $t=3.43\,\mathrm{s}$}\end{minipage}
\hfill
\begin{minipage}[t]{0.241\linewidth}\centering
\includegraphics[width=\linewidth]{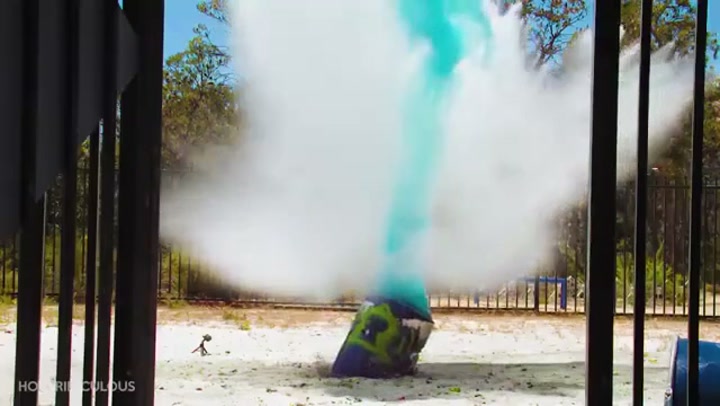}
{\scriptsize $t=6.93\,\mathrm{s}$}\end{minipage}
\hfill
\begin{minipage}[t]{0.241\linewidth}\centering
\includegraphics[width=\linewidth]{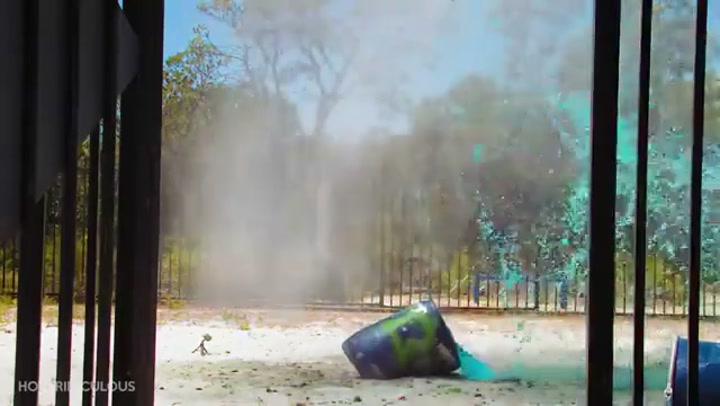}
{\scriptsize $t=10.37\,\mathrm{s}$}\end{minipage}
\par\medskip

\noindent\textbf{(c) Soft-object compression and recovery}\hfill{\scriptsize\texttt{physcap9k\_08653}}\par\smallskip
\begin{minipage}[t]{0.241\linewidth}\centering
\includegraphics[width=\linewidth]{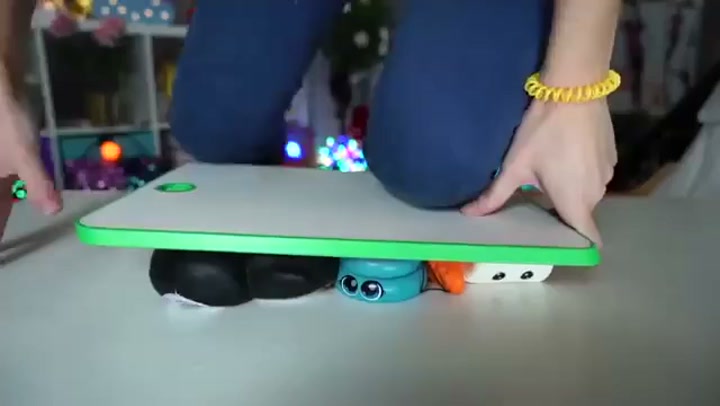}
{\scriptsize $t=0.00\,\mathrm{s}$}\end{minipage}
\hfill
\begin{minipage}[t]{0.241\linewidth}\centering
\includegraphics[width=\linewidth]{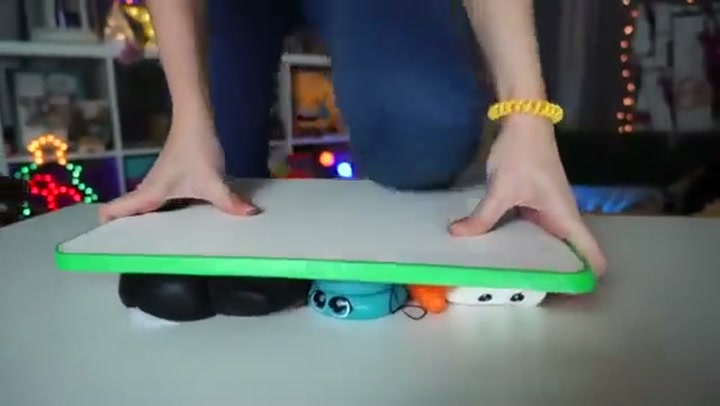}
{\scriptsize $t=2.90\,\mathrm{s}$}\end{minipage}
\hfill
\begin{minipage}[t]{0.241\linewidth}\centering
\includegraphics[width=\linewidth]{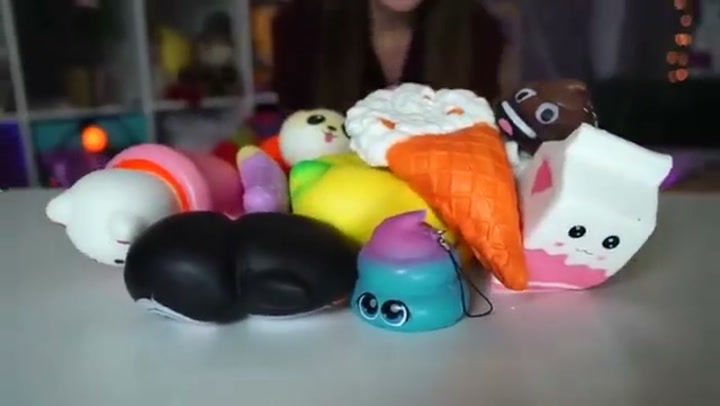}
{\scriptsize $t=7.70\,\mathrm{s}$}\end{minipage}
\hfill
\begin{minipage}[t]{0.241\linewidth}\centering
\includegraphics[width=\linewidth]{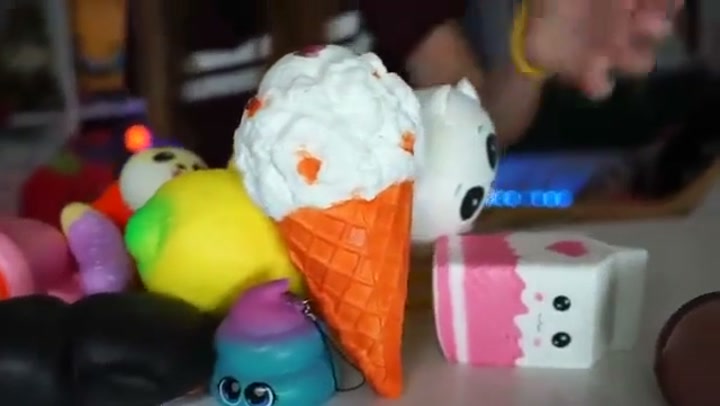}
{\scriptsize $t=18.27\,\mathrm{s}$}\end{minipage}
\par\medskip

\caption{\textbf{Third-person source-video examples.} Rows show surface compaction, material ejection from a barrel, and compression and recovery of soft objects. Frames within each row follow temporal order, with timestamps relative to the start of the clip. Sample IDs identify the videos in the dataset manifest.}
\label{fig:third_examples}
\end{minipage}
\end{figure}

\begin{figure}[H]
\centering\small
\begin{minipage}{0.88\linewidth}
\noindent\textbf{(a) Unpacking and unfolding a chessboard}\hfill{\scriptsize\texttt{egodex500\_0073}}\par\smallskip
\begin{minipage}[t]{0.241\linewidth}\centering
\includegraphics[width=\linewidth]{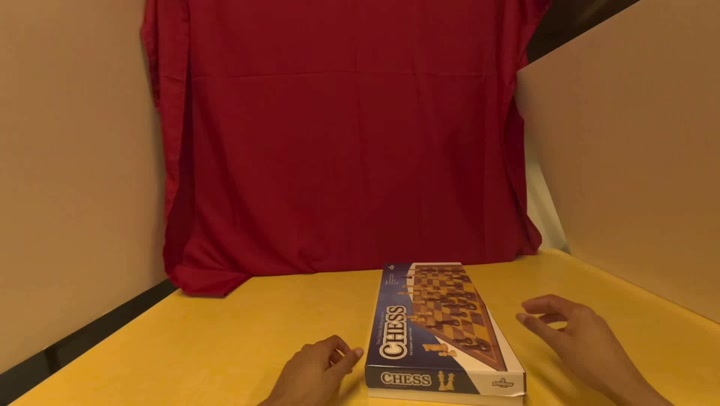}
{\scriptsize $t=0.00\,\mathrm{s}$}\end{minipage}
\hfill
\begin{minipage}[t]{0.241\linewidth}\centering
\includegraphics[width=\linewidth]{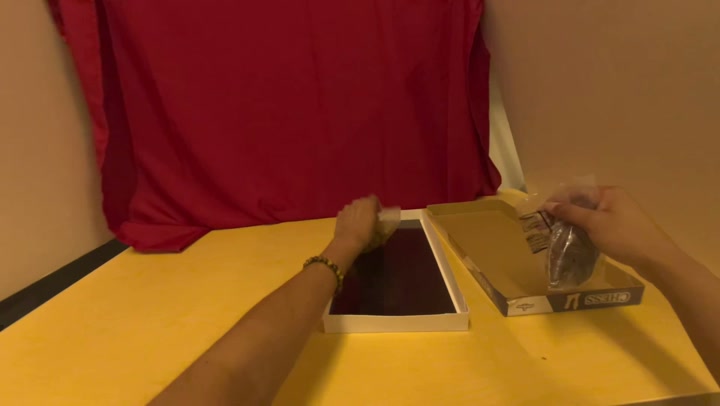}
{\scriptsize $t=3.70\,\mathrm{s}$}\end{minipage}
\hfill
\begin{minipage}[t]{0.241\linewidth}\centering
\includegraphics[width=\linewidth]{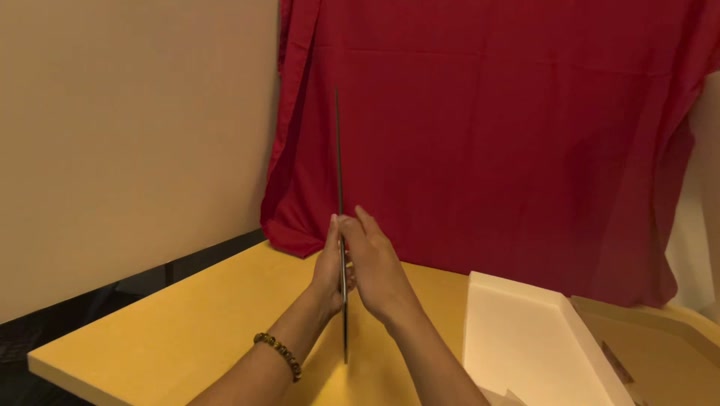}
{\scriptsize $t=7.50\,\mathrm{s}$}\end{minipage}
\hfill
\begin{minipage}[t]{0.241\linewidth}\centering
\includegraphics[width=\linewidth]{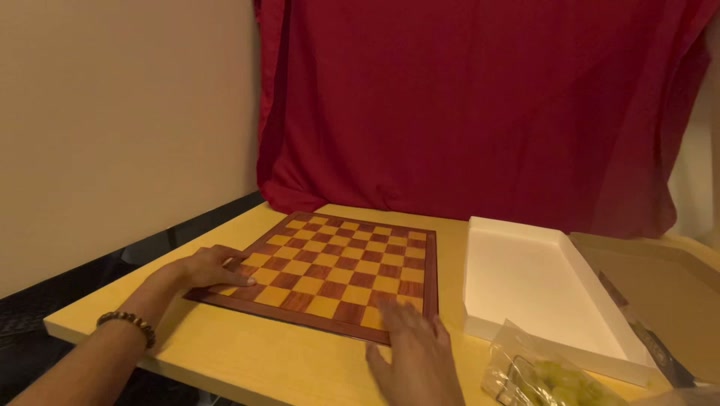}
{\scriptsize $t=11.20\,\mathrm{s}$}\end{minipage}
\par\medskip

\noindent\textbf{(b) Separating a shell from its base}\hfill{\scriptsize\texttt{egodex500\_0289}}\par\smallskip
\begin{minipage}[t]{0.241\linewidth}\centering
\includegraphics[width=\linewidth]{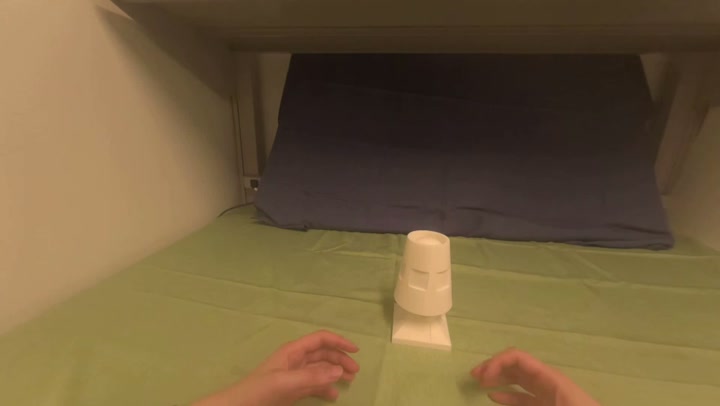}
{\scriptsize $t=0.00\,\mathrm{s}$}\end{minipage}
\hfill
\begin{minipage}[t]{0.241\linewidth}\centering
\includegraphics[width=\linewidth]{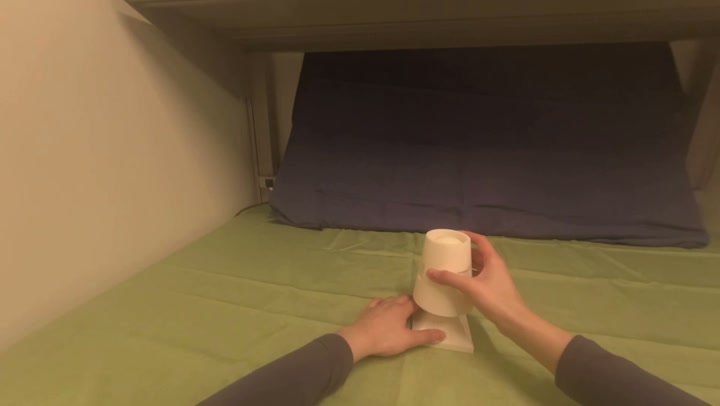}
{\scriptsize $t=2.23\,\mathrm{s}$}\end{minipage}
\hfill
\begin{minipage}[t]{0.241\linewidth}\centering
\includegraphics[width=\linewidth]{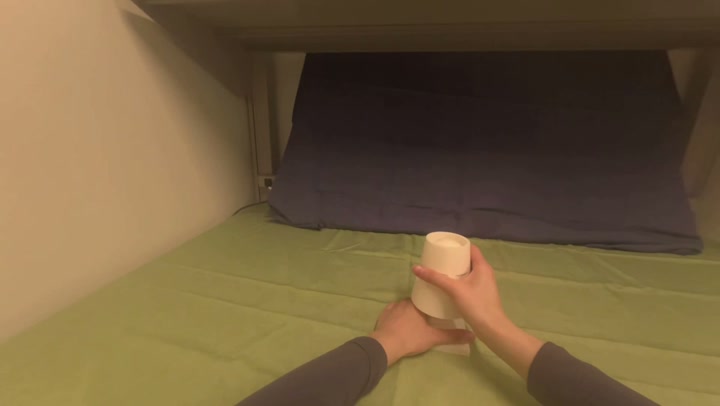}
{\scriptsize $t=4.50\,\mathrm{s}$}\end{minipage}
\hfill
\begin{minipage}[t]{0.241\linewidth}\centering
\includegraphics[width=\linewidth]{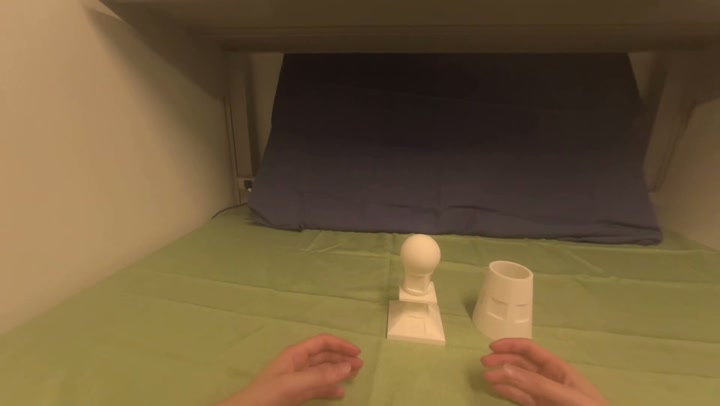}
{\scriptsize $t=6.73\,\mathrm{s}$}\end{minipage}
\par\medskip

\noindent\textbf{(c) Disassembling a small structure}\hfill{\scriptsize\texttt{egodex500\_0480}}\par\smallskip
\begin{minipage}[t]{0.241\linewidth}\centering
\includegraphics[width=\linewidth]{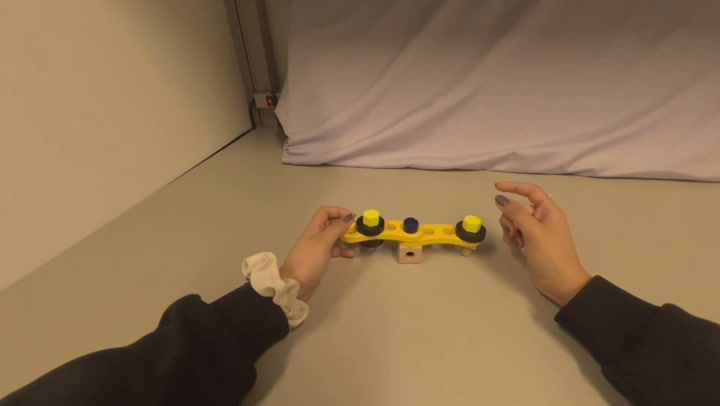}
{\scriptsize $t=0.00\,\mathrm{s}$}\end{minipage}
\hfill
\begin{minipage}[t]{0.241\linewidth}\centering
\includegraphics[width=\linewidth]{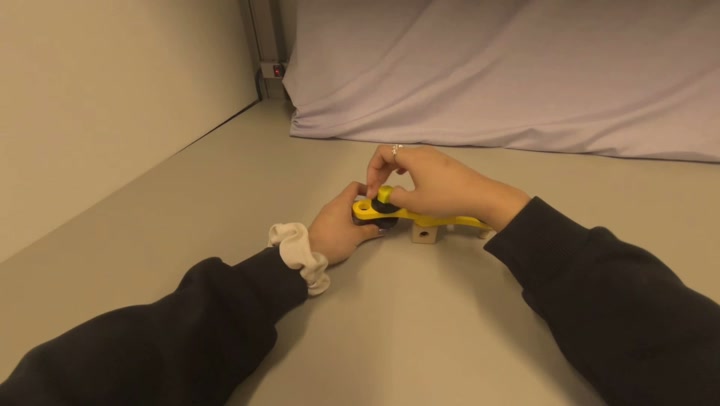}
{\scriptsize $t=2.77\,\mathrm{s}$}\end{minipage}
\hfill
\begin{minipage}[t]{0.241\linewidth}\centering
\includegraphics[width=\linewidth]{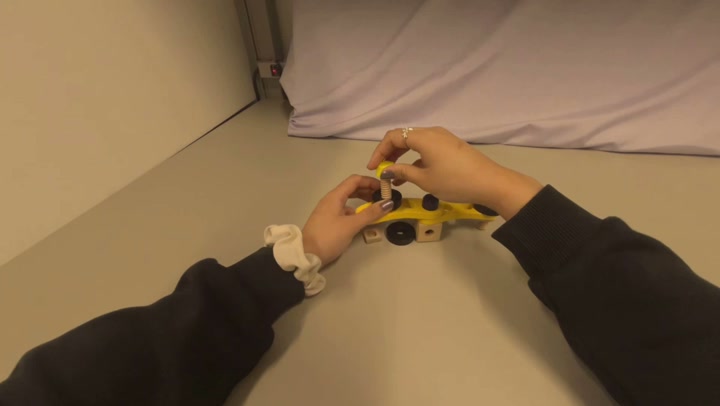}
{\scriptsize $t=5.60\,\mathrm{s}$}\end{minipage}
\hfill
\begin{minipage}[t]{0.241\linewidth}\centering
\includegraphics[width=\linewidth]{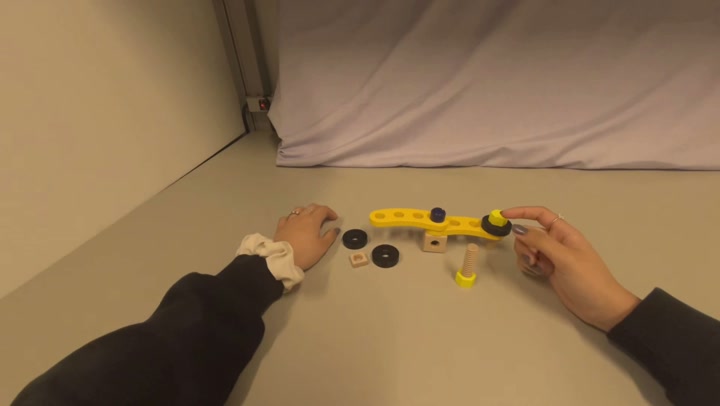}
{\scriptsize $t=8.37\,\mathrm{s}$}\end{minipage}
\par\medskip

\caption{\textbf{Egocentric source-video examples.} Rows show unpacking and unfolding a chessboard, separating an outer shell from its base, and disassembling a small structure. The observations retain both hands and the manipulated objects across time. Timestamps and sample IDs follow the same convention as Figure~\ref{fig:third_examples}.}
\label{fig:ego_examples}
\end{minipage}
\end{figure}

\begin{figure}[!p]
\centering\small
\setlength{\parskip}{0pt}
\captionsetup{skip=6pt}
\begin{minipage}{0.84\linewidth}\centering
{\scriptsize\makebox[0.333\linewidth]{Left wrist}\makebox[0.333\linewidth]{High view}\makebox[0.333\linewidth]{Right wrist}}\par\smallskip
\noindent\textbf{(a) Placing an object in a compartment box}\hfill{\scriptsize\texttt{umi500\_0141}}\par\smallskip
\includegraphics[width=\linewidth]{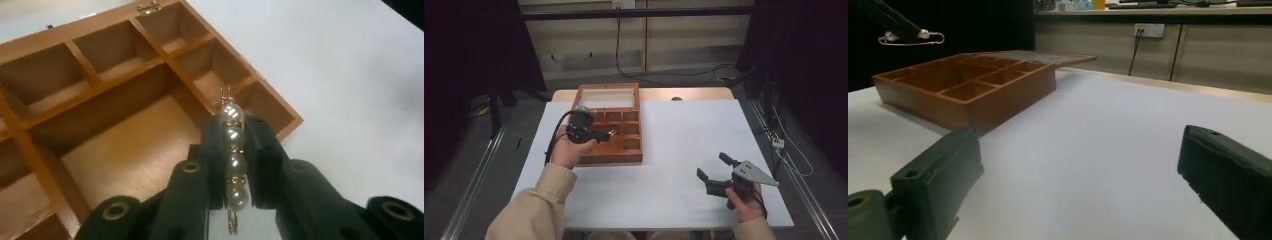}\par
{\scriptsize $t=6.03\,\mathrm{s}$}\par\smallskip
\includegraphics[width=\linewidth]{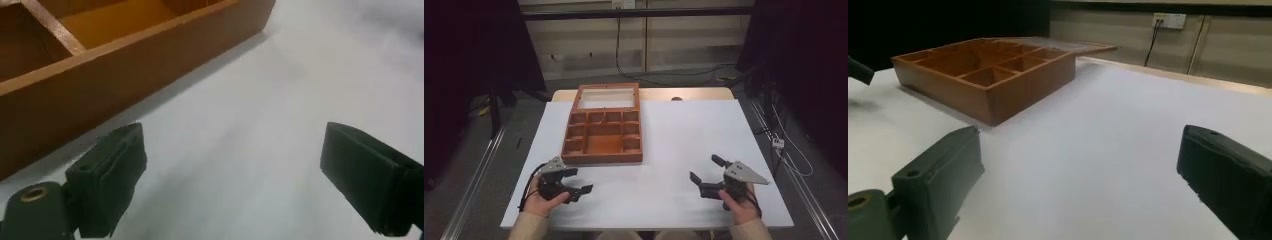}\par
{\scriptsize $t=18.23\,\mathrm{s}$}\par\smallskip
\medskip
\noindent\textbf{(b) Removing a lid with both grippers}\hfill{\scriptsize\texttt{umi500\_0082}}\par\smallskip
\includegraphics[width=\linewidth]{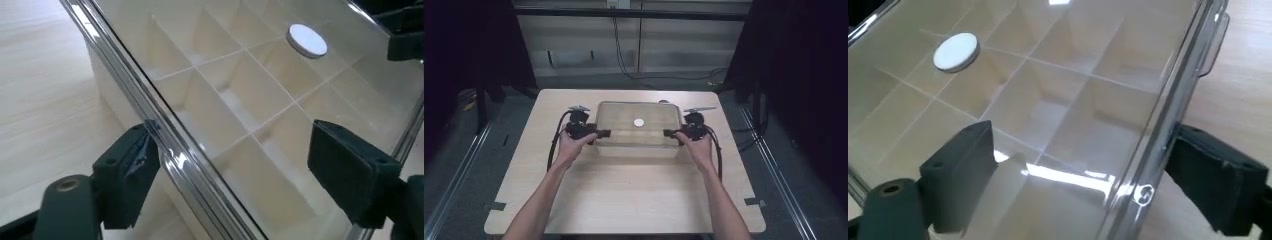}\par
{\scriptsize $t=5.73\,\mathrm{s}$}\par\smallskip
\includegraphics[width=\linewidth]{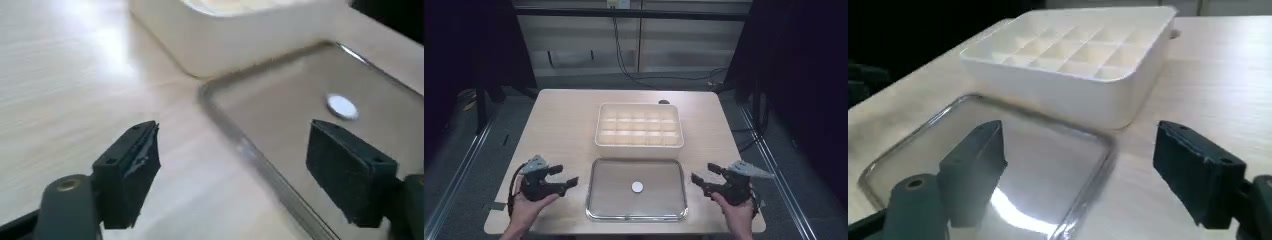}\par
{\scriptsize $t=17.33\,\mathrm{s}$}\par\smallskip
\medskip
\noindent\textbf{(c) Transferring a folded textile to a basket}\hfill{\scriptsize\texttt{umi500\_0297}}\par\smallskip
\includegraphics[width=\linewidth]{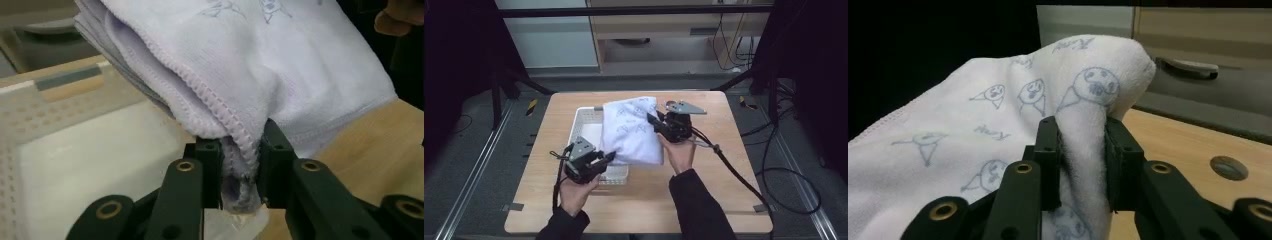}\par
{\scriptsize $t=8.87\,\mathrm{s}$}\par\smallskip
\includegraphics[width=\linewidth]{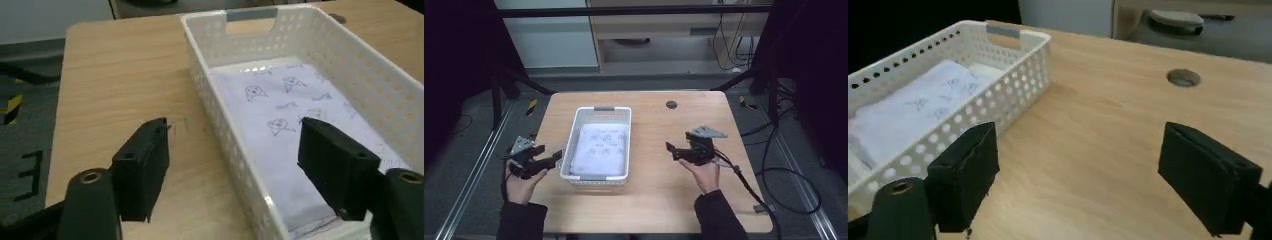}\par
{\scriptsize $t=13.23\,\mathrm{s}$}\par\smallskip
\medskip
\end{minipage}
\caption{\textbf{Handheld-gripper source-video examples.} Three episodes show object placement, lid removal, and textile transfer, each at two time points. Triptychs present synchronized left-wrist, high, and right-wrist views from left to right. Timestamps are measured from the start of each episode.}
\label{fig:umi_examples}
\end{figure}

Figures~\ref{fig:third_examples}--\ref{fig:umi_examples} illustrate the physical processes through three source-video examples per domain. Ordered frames show changes over time, and synchronized views reveal complementary aspects of handheld-gripper manipulation. Sample identifiers and timestamps link the displayed frames to their source recordings.

\let\physcapSavedFloatBarrier\FloatBarrier
\let\FloatBarrier\relax
\section{Annotation Prompts and Review Guidelines}
\let\FloatBarrier\physcapSavedFloatBarrier
\label{app:annotation}

The annotation workflow combines domain-specific cues, structured prompting, and human review. Auxiliary cues select informative observations; prompts organize the observed process into entities, events, and state changes; reviewers verify the resulting records against the video.

\subsection{Auxiliary Cues and Frame Selection}

\paragraph{Hand-pose motion cues.}
For egocentric videos, the camera pose $C_t$ and each hand-joint pose $J_t$ map local coordinates to world coordinates at frame $t$. We obtain the camera-relative joint pose as $C_t^{-1}J_t$. For each hand, let $p_t$ be the centroid of the palm and five fingertips in camera coordinates, $R_t$ the camera-relative palm rotation, and $f$ the video frame rate. Translation speed, angular speed, and the hand motion score are
\begin{align}
    v_t &= f\lVert p_t-p_{t-1}\rVert_2, \\
    \omega_t &= f\arccos\!\left(\operatorname{clip}\!\left(
    \frac{\operatorname{tr}(R_{t-1}^{\mathsf T}R_t)-1}{2},-1,1\right)\right), \\
    m_t &= v_t+0.05\,\omega_t.
    \label{eq:hand_motion}
\end{align}
We initialize speeds to zero at the first frame and retain up to six peaks of $m_t$ per hand as motion summaries. Candidates are selected in descending score order, excluding any candidate within eight frames of a previously selected peak.

\paragraph{Motion-guided visual sampling.}
For the visual frame set, we form the combined translation-speed curve $q_t=\max(v_t^{L},v_t^{R})$, where $L$ and $R$ denote the anatomical left and right hands. We select up to eight peaks from $q_t$ with the same temporal exclusion rule. Let $P$ contain these combined-speed peak frames and $U$ contain up to 12 uniformly spaced frames over the full video. The selected keyframe set is
\begin{equation}
    F=\operatorname{UniqueSort}(U\cup P).
    \label{eq:pose_keyframes}
\end{equation}
Uniform frames cover the sequence, and motion peaks focus inspection on active manipulation intervals. Per-hand motion summaries accompany the selected frames to distinguish the two hands' contributions.

\paragraph{Interaction-role cues.}
HOI-DETR supplies spatial interaction cues from source frames at a detection threshold of 0.5. Outputs include bounding boxes, confidence scores, and hand--object and object--object relations. Associating adjacent detections produces tracks and relation intervals, which the annotator combines with hand-motion summaries and RGB observations to describe the physical process.

\paragraph{Handheld-gripper trajectory cues.}
Handheld-gripper trajectories guide the inspection of interaction transitions. Each gripper provides a position, rotation quaternion, and opening value; derived translation speed, rotation speed, acceleration, and opening changes identify candidate transitions. The high view establishes the overall sequence, and synchronized wrist views resolve local interactions. Camera identifiers locate observations, actor identifiers specify the participating grippers, and ambiguous intervals receive further visual review.

\subsection{Prompt Content}

Annotation prompts request a structured record and a caption of the observed physical process. They establish entity identities and initial states, then order events in time and record actors, affected objects, motion, contact, and state changes. Each event includes source-frame evidence. The caption integrates these records into a coherent account of hand or gripper roles, interactions, and visible outcomes.

Visual evidence anchors event boundaries, grasp outcomes, and state transitions. Motion peaks and gripper-opening changes guide frame selection; RGB observations establish the events and before--after states. Occluded intervals retain the surrounding observations and visibility status for review.

\subsection{Review and Consistency Checks}

Seven professional data annotators perform annotation and cross-checking, with human verification of every final record. Review follows the structure of the physical process: event occurrence, actor and object attribution, and the resulting change. For manipulation sequences, reviewers distinguish reaching, contact establishment, object motion, and contact termination, while retaining sustained holding and repositioning in the timeline.

Multi-view review checks entity identity and before--after states across cameras, distinguishing object motion from wrist-camera motion. Structural validation then checks record references, frame bounds, timing, and timeline coverage. Review decisions accompany the initial generated records and final human-verified annotations.

\section{Evaluation Instructions and Scoring Rubric}
\label{app:evaluation}

\subsection{Dimension-Wise Caption Scoring}

The scoring prompt provides the candidate caption, dimension-specific references, and physical-dimension definitions to Doubao Seed 2.1 Pro (\texttt{doubao-seed-2-1-pro-260628}), with temperature zero and thinking disabled. Egocentric references use the short Gold caption for six dimensions and structured Gold mechanism statements for mechanisms and causality. The judge returns an integer score from 0 to 100 and a short rationale for each applicable dimension, following this instruction:
\begin{quote}
Compare the caption with the supplied reference for each applicable physical dimension. Assess semantic agreement and coverage using the bands in Table~\ref{tab:rubric}. Briefly identify matched content, omissions, and contradictions. Additional details absent from the reference receive neither credit nor a penalty unless they contradict it. Assess causal explanations explicitly expressed in the caption.
\end{quote}

The seven dimensions cover events, participant identities and roles, actions and motion, contact and interaction, states and outcomes, physical mechanisms and causality, and temporal order. Reference content determines applicability, which is held fixed across strategies for each video.

Mechanism scoring assesses explicit causal explanations, including those expressed in ordinary language. Action descriptions such as grasping, lifting, or moving are distinguished from explanations of the underlying mechanism. Within the lowest band of Table~\ref{tab:rubric}, scores of 1--24 indicate weak or fragmentary causal content; zero indicates no expressed reference mechanism or wholly incompatible content.

\begingroup\small\renewcommand{\arraystretch}{1.18}
\begin{xltabular}{\linewidth}{lY}
\caption{\textbf{Per-dimension scoring rubric.} Each applicable dimension receives an integer score. The judge considers both agreement and information coverage.}
\label{tab:rubric}\\
\toprule
\textbf{Score} & \textbf{Criterion} \\
\midrule
\endfirsthead
\toprule
\textbf{Score} & \textbf{Criterion} \\
\midrule
\endhead

90--100 & Nearly complete coverage with no material contradiction. \\
75--89 & Main content agrees, with minor omissions. \\
50--74 & Partial agreement, with important omissions or conflicts. \\
25--49 & Little agreement, with major omissions or conflicts. \\
0--24 & No meaningful agreement, or reversal of core facts. \\
\bottomrule
\end{xltabular}
\endgroup

\subsection{Score Aggregation}

For each video, Eq.~\eqref{eq:caption_score} averages scores over $D(r)$, the dimensions with assessable reference content. Omissions within applicable dimensions affect the score, including a score of zero when the candidate expresses none of the reference mechanisms. Dimensions without assessable reference content are excluded. Overall scores give each video equal weight; dimension-level scores average the applicable entries. All captioning strategies use this aggregation procedure.

\section{Generation and Policy Experiment Configurations}
\label{app:generation}

\subsection{Reconstruction Setup and Statistical Analysis}

Reconstruction conditions share the source first frame and target duration, with one output generated for each caption condition. Results are paired by source video and summarized by generator and domain in Figure~\ref{fig:reconstruction_results}. Relative reduction is $(m-p)/m\times100\%$, where $m$ and $p$ denote the metadata- and physical-caption means. Paired bootstrap resampling over source videos gives the third-person confidence intervals in Table~\ref{tab:reconstruction_uncertainty}.

\begingroup\small\setlength{\tabcolsep}{4pt}\renewcommand{\arraystretch}{1.18}
\begin{xltabular}{\linewidth}{Yll}
\caption{\textbf{Paired differences in third-person reconstruction distance.} The 95\% bootstrap intervals use physical minus metadata distance; negative values favor physical captions.}
\label{tab:reconstruction_uncertainty}\\
\toprule
\textbf{Model} & \textbf{Metric} & \textbf{95\% interval} \\
\midrule
\endfirsthead
\toprule
\textbf{Model} & \textbf{Metric} & \textbf{95\% interval} \\
\midrule
\endhead

H3 & LPIPS & $[-0.0588,-0.0295]$ \\
 & PVCS & $[-0.3402,-0.1311]$ \\
LingBot & LPIPS & $[-0.0670,-0.0294]$ \\
 & PVCS & $[-0.3720,-0.1648]$ \\
\bottomrule
\end{xltabular}
\endgroup

\subsection{Manipulation-Policy Configuration}
\label{app:policy_config}

The single-task experiment evaluates instruction enrichment on RoboTwin 2.0's block-size ordering subtask using an Aloha-AgileX bimanual embodiment. The policy arranges three blocks from largest to smallest across the center of the table. Both conditions use 50 demonstrations containing 23,098 frames, three camera views, a 14-dimensional robot state, and corresponding action targets. The original instruction states the ordering goal; enriched instructions add the demonstrated sequence, gripper roles, and resulting placement and support relations. We construct these instructions from demonstration events and scene information, verify them against keyframes, and keep them fixed within each demonstration.

Enriched instructions replace the original text for half of the repeated training indices and contain 151--167 tokens under the recorded tokenizer, within the 256-token language limit. The remaining training indices and all evaluation rollouts use the original short task instruction. Table~\ref{tab:policy_config} lists the shared training and execution settings.

\begingroup\small\renewcommand{\arraystretch}{1.16}
\begin{xltabular}{\linewidth}{p{0.33\linewidth}Y}
\caption{\textbf{Shared policy-training and execution settings.} The language substitution described above is the experimental difference.}
\label{tab:policy_config}\\
\toprule
\textbf{Setting} & \textbf{Value} \\
\midrule
\endfirsthead
\toprule
\textbf{Setting} & \textbf{Value} \\
\midrule
\endhead

Initialization & Released \texttt{pi05\_base} weights \\
Adaptation & OpenPI JAX LoRA \\
Training updates / batch size & 5,000 / 32 \\
Training seed & 0 for both conditions \\
Optimizer & AdamW \\
Learning-rate schedule & 200 warm-up steps; peak $5\times10^{-5}$; cosine decay to $5\times10^{-6}$ \\
Language limit & 256 tokens \\
Checkpoint selection & Final checkpoint after 5,000 updates \\
Prediction chunk & 50 steps of 14-dimensional absolute joint actions \\
Replanning & Execute the first 10 predicted steps, then observe again \\
Rollout limit & 1,200 executed action steps \\
Evaluation scenes & 20 standard and 20 randomized; identical across policies \\
Rollouts & One per policy per scene; 80 in total \\
Success & Native task success criterion \\
\bottomrule
\end{xltabular}
\endgroup

Each held-out scene is checked for solvability with the expert policy before evaluation. Randomized scenes vary backgrounds, clutter, lighting, and table height.

\end{document}